\documentclass[twoside,twocolumn,english]{article}
\usepackage[titletoc,toc]{appendix}
\usepackage{mathptmx}
\usepackage[T1]{fontenc}
\usepackage[latin9]{inputenc}
\usepackage{geometry}
\usepackage{fancyhdr}
\usepackage{color}
\usepackage{array}
\usepackage{float}
\usepackage{url}
\usepackage{multirow}
\usepackage{amsmath}
\usepackage{amssymb}
\usepackage{graphicx}
\usepackage[authoryear]{natbib}
\usepackage{babel}
\usepackage{subfig}
\usepackage{bm} 
\usepackage{balance} 

\newcommand{\inv}{^{-1}}
\newcommand{\tr}{^{\!\top}}
\newcommand{\invtr}{^{-\!\top}}
	
\newcommand{\invs}{^{-1\backslash2}}

\newcommand{\argmax}{\operatornamewithlimits{argmax}}
\newcommand{\argmin}{\operatornamewithlimits{argmin}}

\usepackage{savesym}
\savesymbol{AND}
\savesymbol{OR}
\savesymbol{NOT}
\savesymbol{TO}
\savesymbol{COMMENT}
\savesymbol{BODY}
\savesymbol{IF}
\savesymbol{ELSE}
\savesymbol{ELSIF}
\savesymbol{FOR}
\savesymbol{WHILE}
\usepackage{algorithm}
\usepackage{algorithmicx}
\usepackage{algpseudocode}
\usepackage{float}
\floatstyle{boxed}
\newfloat{myprogram}{tb}{alg}
\floatname{myprogram}{Algorithm}
\renewcommand*\Call[2]{\textproc{#1}(#2)} 

\usepackage{amssymb}

\let\emptyset\varnothing
\usepackage[colorlinks=true,allcolors=black,bookmarks=false]{hyperref}

\makeatletter

\floatstyle{ruled}
\newfloat{algorithm}{tbp}{loa}
\providecommand{\algorithmname}{Algorithm}
\floatname{algorithm}{\protect\algorithmname}

\@ifundefined{date}{}{\date{}}

\makeatother 

\usepackage{sectsty}
\sectionfont{\large}
\subsectionfont{\normalsize}
\subsubsectionfont{\normalsize\itshape}

\usepackage{mwe}
\edef\mttopnumber{\arabic{topnumber}} 

\begin{document}
\twocolumn[\begin{@twocolumnfalse}
\title{
\renewcommand{\thefootnote}{\arabic{footnote}} 
Highly Efficient Compact Pose SLAM with SLAM++\footnotemark[1]
\renewcommand{\thefootnote}{\fnsymbol{footnote}} 
}
\author{Viorela Ila, Lukas Polok, Marek Solony and Pavel Svoboda}
\maketitle

\begin{abstract}

The most common way to deal with the uncertainty present in noisy sensorial perception and action is to model the problem with a probabilistic framework. 
Maximum likelihood estimation (MLE) is a well-known estimation method used in many robotic and computer vision applications. Under Gaussian assumption, the MLE converts to a nonlinear least squares (NLS) problem.
Efficient solutions to NLS exist and they are based on iteratively solving sparse linear systems until convergence. In general, the existing solutions provide only an estimation of the mean state vector, the resulting covariance being computationally too expensive to recover. Nevertheless, in many simultaneous localisation and mapping (SLAM) applications, knowing only the mean vector is not enough. Data association, obtaining reduced state representations, active decisions and next best view are only a few of the applications that require fast state covariance recovery. Furthermore, computer vision and robotic applications are in general performed online. In this case, the state is updated and recomputed every step and its size is continuously growing, therefore, the estimation process may become highly computationally demanding.

This paper introduces a general framework for incremental MLE called SLAM++, which fully benefits from the incremental nature of the online applications, and provides efficient estimation of both the mean and the covariance of the estimate. Based on that, we propose a strategy for maintaining a sparse and scalable state representation for large scale mapping, which uses information theory measures to integrate only informative and non-redundant contributions to the state representation. SLAM++ differs from existing implementations by performing all the matrix operations by blocks. This led to extremely fast matrix manipulation and arithmetic operations. Even though this paper tests SLAM++ efficiency on SLAM problems, its applicability remains general.


\end{abstract}
\end{@twocolumnfalse}]

\footnotetext[1]{Not to be confused with ``\textit{SLAM++: Simultaneous Localisation and Mapping at the Level of Objects}'' proposed by \cite{Salas-Moreno13cvpr}. Both softwares were developed and named independently and simultaneously. Our incremental SLAM framework was first introduced at ICRA 2013 (May 6-10) during the interactive presentation of our seminal work in this direction~\cite{Polok13icra}.} 
\let\thefootnote\relax\footnotetext{Draft manuscript, \today. Submitted to IJRR.}
\let\thefootnote\relax\footnotetext{V. Ila is with the Australian National University, Canberra, Australia. {\tt\small viorela.ila@anu.edu.au}, L. Polok, M. Solony and P. Svoboda are with Brno University of Technology, Faculty of Information Technology. Bozetechova 2, 612 66 Brno, Czech Republic{\tt\small \{ipolok, isolony, isvoboda\}@fit.vutbr.cz}.}

\newcommand{\thefootnote}{\arabic{footnote}} 
\addtocounter{footnote}{1} 

\section{Introduction}

Probabilistic methods have been extensively applied in robotics and computer vision to handle noisy perception of the environment and the inherent uncertainty in the estimation.
There are a variety of solutions to the estimation problems in today's literature. Filtering and maximum likelihood estimation (MLE) are among the most used in robotics. Since filtering easily becomes inconsistent when applied to nonlinear processes, MLE gained a prime role among the estimation solutions. In simultaneous localisation and mapping (SLAM) \cite{Dellaert06ijrr,Kaess08tro,Kuemmerle11icra,Kaess11ijrr} or other mathematical equivalent problems such as bundle adjustment (BA) \cite{Agarwal09iccv,Konolige10bmvc} or structure from motion (SFM) \cite{Beall10iros}, the estimation problem is solved by finding the MLE of a set of variables (e.g. camera/robot poses and 3D points in the environment) given a set of observations. Assuming Gaussian noises and processes, the MLE has an elegant nonlinear least squares (NLS) solution.

A major challenge appears in online robotic applications, where the state changes at every step.
For very large problems, updating the system and solving the NLS at every step can become very expensive. Efficient incremental NLS solutions have been developed, either by working directly on the matrix factorization of the linearised system \cite{Kaess08tro}, by using graphical model-based data structures such as the Bayes tree, \cite{Kaess10wafr,Kaess11ijrr} or by exploiting the inherent sparse block structure 
 \cite{Polok13rss}. 

The existing incremental NLS solutions provide fast and accurate estimations of the mean state vector, for example the mean position of the robot or the features in the environment. However, in real applications, the uncertainty of the estimation plays an important role. It is given by the covariance matrix, which generalizes the notion of variance to multiple dimensions. In particular, the marginal covariances encoding the uncertainties between a subset of variables are required in many applications.

In online SLAM and SFM applications, the marginal covariances can be used to perform \emph{data association},~\cite{Neira01tro},~\cite{Kaess09ras}, to obtain a \emph{reduced state representation} which will further allow efficiently handling large scale applications,~\cite{Ila10tro,Kretzschmar11iros,Johannsson13icra,Huang13ecmr,Carlevaris13icra}, to perform \emph{active sensing},~\cite{Davison02pami,Haner12eccv}, to decide on the \emph{next best actions} to take,~\cite{Vidal06icra} or the \emph{best reliable path} to follow,~\cite{Valencia13tro} to reduce the uncertainty of the estimate or simply to \emph{provide feedback} about the error of the estimation.

A novel technique to obtain exact \emph{marginal covariances} in an online NLS framework, where the system changes every step was proposed in \cite{Ila15icra}. It is based on incremental updates of marginal covariances every time new variables and observations are integrated into the system, and on the fact that, in practice,when, the changes in the linearisation point are often very small and can be ignored. The formulation and the implementation depart from the existing approaches by exploiting the sparsity and the block structure of the robotic problems. A new data-structure was introduced in our previous work, which enabled efficient block-matrix manipulation, block-wise arithmetic operations \cite{Polok13icra}, and block matrix factorisation. Furthermore, a novel algorithm for incrementally solving NLS problems, with a new efficient incremental ordering scheme was proposed in \cite{Polok13rss} and extensively tested on well known datasets. The current work integrates the strategy for fast covariance recovery proposed in \cite{Ila15icra} into the NLS solution introduced in \cite{Polok13rss}.

Furthermore, based on the incremental solver with fast covariance recovery, this paper proposes an incremental algorithm which performs state-based data association and, at the same time, maintains a scalable representation of the state. This is achieved by computing two measures, the proximity in terms of sensor range of the current pose to any other previous poses the robot traveled, and the mutual information of each candidate link, which require good estimations of the marginal covariances. The proposed algorithm is an extension of the 2D filtering approach proposed by~\cite{Ila10tro} to the 3D MLE SLAM framework. Comparing to 2D filtering approach, several new challenges are met in compact 3D MLE SLAM. Careful handling of the thresholds in the rotation space, complex incremental covariance calculations, changes in the linearization point are among them. 

This paper addresses all above mentioned challenges and proposes solutions which are implemented within the SLAM++ nonlinear least squares library. The name of the library was chosen to reflect the incremental nature of the solving. The early version of the library was introduced in~\cite{Polok13icra} and proved to supersede the existing similar implementations. The current version differentiates from other similar libraries used proposed in SLAM community (g2o (~\cite{Kuemmerle11icra}), iSAM~\cite{Kaess08tro} or iSAM2~\cite{Kaess11ijrr}) in several aspects:
\begin{itemize}
\item It is based on a sparse, block data-structure which has been proven to be extremely efficient for problems with higher than two variable size~\cite{Polok13icra,Polok13hpc}
\item It integrates an efficient incremental, sparse, block Cholesky factorisation~\cite{Polok13rss}  
\item It integrates an incremental ordering which maintains a sparse matrix factorization, and in consequence allows for efficient solving~\cite{Polok13rss}  
\item Allows for incremental covariance calculation which provides marginal covariances two orders of magnitude faster than previous implementations~\cite{Ila15icra}  
\item Integrates a compact pose SLAM algorithm where information theoretic measures are used to maintain a sparse, conservative representation of the MLE SLAM 
\item  It comprises not only the C++ code but also the datasets and the scripts that can be used to reproduce the benchmarks presented in this paper, as well as scripts for automatic calculation of all involved thresholds in compact pose SLAM
\end{itemize}
 
The above mentioned characteristics are validated in this paper through extensive tests of the time efficiency, accuracy and conservativeness of the estimation on several simulated and real datasets.

\section{Related Works}\label{Sec:Related}

When using MLE in real applications such as online SLAM, the recovery of the uncertainty of the estimate, \emph{the covariance}, can become a computational bottleneck. The covariance is needed, for example, to generate data association hypotheses, or to evaluate the mutual information required in active mapping or graph sparsification. The calculation of the covariance amounts to inverting the system matrix, $\Sigma = \Lambda\inv$, where the resulting matrix $\Sigma$ is no longer sparse.

Several approximations for marginal covariance recovery have been proposed in the literature. 
\cite{Thrun04ijrr} suggested using conditional covariances, which are inversions of sub-blocks of $\Lambda$ called the Markov blankets. The result is an overconfident approximation of the marginal covariances. Online, conservative approximations were proposed in \cite{Eustice06ijrr}, where at every step, the covariances corresponding to the new variables are computed by solving the augmented system with a set of basis vectors. The recovered covariance column is passed to a Kalman filter bank, which updates the rest of the covariance matrix. The filtering is reported to run in constant time, and the recovery speed is bounded by the linear solving. In the context of MLE, belief propagation over a spanning tree or loopy intersection propagation can be used to obtain conservative approximations suitable for data association, \cite{Tipaldi07iros}.

An exact method for sparse covariance recovery was proposed in \cite{Kaess09ras}. It is based on a recursive formula from \cite{Bjorck96book,Golub1980laa}, which calculates any covariance elements on demand from the other covariance elements and the elements of the Cholesky factor. It was implemented using a hash map, to provide for fast dependence tracking. The method, though, does not benefit from the incremental nature of the online problem.

In their paper, \cite{Prentice11isrr} proposed a covariance factorization for calculating linearized updates to the covariance matrix over arbitrary number of planning decision steps in a partially observable Markov decision process (POMDP). The method uses matrix inversion lemmas to efficiently calculate the updates. The idea of using factorizations for calculating inversion update is not new, though. A discussion of applications of the Sherman-Morrison and Woodbury formulas is presented in \cite{Hager89siam}. Specifically, it states the usefulness of these formulas for updating matrix inversion, after a small-rank modification, where the rank must be kept low enough in order for the update to be faster than simply calculating the inverse. In our latest work \cite{Ila15icra}, we proposed an algorithm which confirms this conclusion, and also proves its usefulness in online SLAM applications.

When optimizing over the entire robot trajectory, the SLAM solution is more accurate but the computational complexity grows every step and can become intractable for long runs. This can be alleviated by several techniques. One way is to optimize for a small window around the current pose, and only do expensive optimization when large loops are closed \cite{Huang11iros,Sibley08lnee}. This technique assumes that the robot is running in open loop for long periods of time and that the loop closure detection is strictly based on appearance based sensors and has no information about the current estimates. Without prior information about the robot position and the topology of the map, appearance based methods can be easily tricked by perceptual aliasing \cite{Ila10tro}. Other techniques consider only a subset of empirically selected key frames to perform global optimization, while the intermediate frames are referred to the optimized key frames \cite{Klein07ismar}. These techniques can reduce the run time, nevertheless they require more principled selection methods to automatically restrict the state estimation problem size to that of the area the robot operates in.

Recently, pose graph has received a lot of attention in the SLAM literature; when having a good estimation of the robot pose, the map can be retrieved by simply referring the relative measurements which can come from a large variety of sparse or dense sensors. Even if very efficient solutions to medium-size pose graph SLAM exist \cite{Kaess08tro, Kuemmerle11icra, Polok13icra}, it can become inefficient for very long runs. The size of the graph grows in time and it is not bounded by the size of the environment. Apart from the node count, the sparsity of the graph also affects the performance. Therefore, current literature in SLAM proposes reduction strategies which involve both, node marginalization and edge sparsification. Eliminating a node through marginalization actually introduces more edges which densify the graph. \cite{Carlevaris13icra} initially proposed a technique which approximates the dense clique generated by the node removal by using the Chow-Liu tree (CLT)~\cite{Chow68tit}. Those approximations produce overconfident estimates, therefore, in their latter work, \cite{Carlevaris14icra} proposed a sparse CLT approximations which result in conservative estimates. Conservative estimates are preferable in SLAM applications where data association is using the state estimate, or in applications where the map is used to plan a path to a goal point in the map. \cite{Carlevaris13icra} also pointed out the difference between composition and marginalization for the specific case where the pose to be removed is involved in a loop closure link. The strategy proposed in this paper automatically avoids this situation by always keeping the poses which are involved in potentially informative links.

Most of the existing graph pruning methods reduce the size of the graph in batch mode, after the graph was already built. \cite{Johannsson13icra} proposes a technique to temporally scalable SLAM that decides on the fly which nodes are added to the graph. This is done by introducing the concept of active nodes and re-uses the already existing poses in the graph when the robot revisits previously mapped areas. \cite{Huang13ecmr} considers the incremental nature of the SLAM problem and proposes a technique to keep consistent estimates by retaining all the information of the marginalized-out poses.

This paper shows how, based on the efficient covariance recovery strategy, a principled method that incrementally maintain a compact representation of a SLAM problem can be obtained. The system only keeps non-redundant poses and informative links. The result is a set of robot poses nicely distributed in the information space. This translates in highly scalable solutions and great speed-ups for large scale estimation problems, while the accuracy is marginally affected. The idea was previously introduced in a filtering framework \cite{Ila10tro} to maintain a compact representation of a 2D pose SLAM, and this paper extends it to large scale MLE estimation for 3D SLAM, addressing all the corresponding challenges. The method can be easily extended beyond pose SLAM, to problems with different types of measurements, for example trinary factors present in structure-less BA \cite{Indelman12bmvc}, or different types of variables, for example landmark SLAM, with the constrain that variables can be eliminated as long as measurement composition is possible.   
%
\section{Incremental Estimation}\label{Sec:Incr}

In this paper, the estimation problem is formulated as a maximum likelihood estimation of a set of variables $\boldsymbol \theta$ given a set of observations ${\bf z}$. The SLAM example is considered, where the vector \mbox{${\boldsymbol \theta} = \left[ \theta_1 \ldots \theta_n \right]$} gathers the variables corresponding to the robot poses and the map, and the vector \mbox{${\bf z}= \left[ z_1 \ldots z_m \right]$} gathers the available observations.
This estimation has to be done incrementally in an online application; every step a new variable and the associated measurements are integrated into the system and a new solution is calculated. In this section, we briefly show how the MLE problem is formulated and solved incrementally.

\subsection{State Estimation}\label{Subsec:State}

The goal is to obtain the maximum likelihood estimate (MLE) of a set of variables in $\boldsymbol \theta$ at every step, given the available observations in ${\bf z}$:
\begin{equation}
\boldsymbol{\theta}^{*}=\argmax_{\boldsymbol \theta}\: P(\boldsymbol{\theta}\mid{\bf z})=\argmin_{\boldsymbol{\theta}}\:\left\{-\log(P({\boldsymbol \theta}\mid{\bf z}))\right\}\;.
\label{eq:ML}
\end{equation}
It is well known that, assuming Gaussian distributed processes and measurements, the MLE has an elegant and accurate solution based on solving a NLS problem:
\begin{equation}
\boldsymbol{\theta}^{*}=\argmin_{\boldsymbol \theta} \left\{\frac{1}{2}\sum_{k=1}^{m} \displaystyle \left\Vert h_{k}( \theta_{i_k},\theta_{j_k}) \ominus z_{k}\right\Vert _{\Sigma_{k}}^{2} \right\}\;,
\label{eq:NLS}
\end{equation}
where $h({\theta}_{i_k},{\theta}_{j_k})$ are the nonlinear measurement function and $z_{k}$ are the measurements with normally distributed noise with covariance $\Sigma_{k}$. Finally,
$\ominus$ is the inverse composition operator, e.g. for 3D pose SLAM, if we have two poses $p, q \in {\mathbb R}^6\hookleftarrow \mathfrak se(3)$, and $\mathfrak se(3)$ is the Lie algebra of the special Euclidean group $SE(3)$, \mbox{$p \ominus q = \log(P\:Q\inv)$}, with \mbox{$P,Q \in SE(3)$} and $P=\exp(p)$ and $Q=\exp(q)$, defining the logarithm map as \mbox{$\log : SE(3) \rightarrow \mathfrak se(3)$} and the exponential map as \mbox{$\exp : \mathfrak se(3) \rightarrow SE(3) $}.  


Iterative methods, such as Gauss-Newton or Levenberg-Marquardt, are often used to solve the NLS in \eqref{eq:NLS}. This is usually addressed by solving a sequence of linear systems at every iteration. Linear approximations of the nonlinear residual functions around the current linearisation point $\boldsymbol \theta^i$ are calculated:
\begin{equation}
\tilde{\bf r}(\boldsymbol \theta^i) = {\bf r}(\boldsymbol \theta^i) + J(\boldsymbol \theta^i)(\boldsymbol \theta \ominus \boldsymbol \theta^i)\;,
\label{eq:linearisation}
\end{equation}
with ${\bf r}({\boldsymbol \theta})=\left[ r_1, \ldots, r_m\right]\tr$ being a vector gathering all nonlinear residuals of the type \mbox{$r_k=h_{k}( \theta_{i_k},\theta_{j_k}) \ominus z_{k}$} and $J$ being the Jacobian matrix which gathers the derivatives of the components of ${\bf r}(\boldsymbol \theta)$.
With this, the NLS in~\eqref{eq:NLS} is approximated by a linear one and solved by successive iterations:
\begin{equation}
{\boldsymbol \delta}^{*}=\argmin_{\boldsymbol \delta} \frac{1}{2} \left\Vert A\:{\boldsymbol \delta}-{\bf b} \right\Vert^{2}\;,
\label{eq:linearEq}
\end{equation}
where the matrix $A$ and the vector ${\bf b}$ are defined as $A\triangleq D\invs J $ and ${\bf b} \triangleq - D\invs {\bf r} $, with $D$ gathering all the $\Sigma_k$ measurement covariances~\cite{Dellaert06ijrr}. The correction ${\boldsymbol \delta}\triangleq \boldsymbol \theta \ominus \boldsymbol \theta^i$ towards the solution is obtained by solving the linear system:
\begin{equation}
A\tr \: A\: {\boldsymbol\delta} = A\tr{\bf b}\;,\;\text{or}\;\; \Lambda\boldsymbol{\delta}={\boldsymbol \eta}\;,
\label{eq:normalEq1}
\end{equation}
with $\Lambda$, the square symmetric positive definite system matrix and ${\boldsymbol \eta}$ the right hand side.
In the case of sparse problems such as SLAM, it is common to apply sparse matrix factorization, followed by backsubstitutions to obtain the solution of the linear system.
The Cholesky factorization of the matrix $\Lambda$ has the form $R\tr\:R=\Lambda$, where $R$ is an upper triangular matrix with positive diagonal entries.
The forward and backsubstitutions on $R\tr {\bf d} = {\boldsymbol \eta}$ and $R\:{\boldsymbol \delta}= {\bf d}$ first recover ${\bf d}$, then the actual solution ${\boldsymbol \delta}$.
After computing $\boldsymbol \delta $, the new linearisation point becomes \mbox{${\boldsymbol \theta}^{i+1}=\boldsymbol \theta^i \oplus \boldsymbol{\delta}$}, $\oplus$ being the vectorial composition operator, given two poses $p, q \in {\mathbb R}^6\hookleftarrow \mathfrak se(3)$, \mbox{$p \oplus q = \log(P\:Q)$}, with \mbox{$P,Q \in SE(3)$} and $P=\exp(p)$ and $Q=\exp(q)$.
The nonlinear solver iterates until the norm of the correction becomes smaller than a tolerance or the maximum number of iterations is reached.

\subsection{Incremental Updates}\label{Subsec:IncUp}
For large online problems, updating and solving the entire system at every step becomes very expensive. Therefore, online estimations need to be approached incrementally. At every step, a new variable and the corresponding observations are integrated into the system. This translates to new elements added to the summand in \eqref{eq:NLS} and consequently new rows added to the matrix $A$ in \eqref{eq:linearEq}. For example, in case of a new observation $h_k(\theta_i,\theta_j)$ involving two variables, the update of $A$ becomes:
\begin{equation}
\hspace{.025cm}\hat A =\left [\begin{array}{c} A \\ A_u \end{array}\right ],\;\text{with}\;\; A_u= \left[ 0\: \ldots{J}_{i}^j\:\Sigma_k\invs \ldots\: 0\: \ldots\ {J}_{j}^i\Sigma_k\invs \right ]\;.\hspace{-.4cm} 
\label{eq:rowA}
\end{equation}
This translates into additive updates of the system matrix $\Lambda$.
The sparsity of $A_u$ can be used to identify the blocks in $\Lambda$ as well as segments in the r.h.s ${\boldsymbol \eta}$ that change with this update. Considering $A_k = \left[ {J}_{i}^j\:\Sigma_k\invs \ldots\: 0\: \ldots\ {J}_{j}^i\Sigma_k\invs \right ] $ the part of $A_u$ which actually affects the system, we obtain:
\begin{equation}
\begin{array}{cc}
\hat \Lambda= \left [\begin{array}{cc}
 \Lambda_{00} & \Lambda_{10}\tr \\
 \Lambda_{10} & \Lambda_{11}+\Omega\end{array} \right ] \;,\;
 \hat {\boldsymbol \eta}=\left[\begin{array}{c} {\boldsymbol \eta}_{0}\\
{\boldsymbol \eta}_{1}+{\boldsymbol \omega}\end{array}\right],
\end{array}
\label{eq:updateLambda}
\end{equation}
where $ \Lambda_{00}$, $ \Lambda_{10}$ and ${\boldsymbol \eta}_{0}$ are the parts of the systems which remain unchanged, while $\Lambda_{11}$ and $ {\boldsymbol \eta}_{1}$ increment with $\Omega= A_k\tr\: A_k$ and ${\boldsymbol \omega}= -A_k\tr{r_k}$, respectively.
Theoretically, the solution ${\boldsymbol \theta}$ changes every step and in consequence the system matrix $\Lambda$ and the r.h.s. $\boldsymbol \eta$ change entirely. In practice, though, the changes in the state vector are very small, sometimes affecting only a small part of the vector, and in consequence the changes in the system can be isolated and treated accordingly. This is the key factor in updating and solving a nonlinear system incrementally. This allowed for fast algorithms to solve the incremental estimation problem \cite{Kaess08tro,Kaess11ijrr,Polok13rss}. If the changes in the linearisation point are substantial, the system matrix needs to be fully recalculated by computing the Jacobians using the new linearisation point, but this happens less frequently in an incremental estimation problem. The recently introduced data structure in~\cite{Kaess11ijrr}, the Bayes tree, offers the possibility to develop incremental algorithms where reordering and re-linearization are performed fluidly, without the need of periodic updates. In \cite{Polok13rss} we proposed an elegant and highly efficient approach which combines the efficiency of matrix implementation and considers the insights gained using the Bayes tree data structure.

\subsection{Incremental Solving}\label{Subsec:IncSolve}
As mentioned above, when incrementally calculating the solution to an NLS which continuously updates with new variables and observations, two situations can be distinguished; the smal changes in the linearisation point can be ignored and only the parts affected by the update need to be recalculated, or, less often, the linearization point changes significantly and the system needs to be recalculated entirely.

In our previous work \cite{Polok13rss}, we have shown that, in the same way as in $\Lambda$, the parts of the factorized form $R$ affected by the update, can also be identified. The updated $\hat R$ factor and the corresponding r.h.s. $\hat {\bf d}$ can be written as:
\begin{equation}
\begin{array}{cc}
\hat R= \left [\begin{array}{cc}
R_{00} & R_{01} \\
0 & \hat R_{11}
\end{array} \right ]\;,\; \hat {\bf d}=\left[\begin{array}{c} {\bf d_0}\\ \hat{\bf d_1}\end{array}\right] \;.
\end{array}
\label{eq:updateL}
\end{equation}
From $\hat \Lambda = \hat R \tr \:\hat R$, \eqref{eq:updateLambda} and \eqref{eq:updateL} the updated part of the Cholesky factor and the r.h.s can be easily computed:
\begin{align}
\hat R_{11}&= \text{chol}(R_{11}\tr \: R_{11}+\Omega) \label{eq:L22Omega0}\\
\hat {\bf d_1}&= \hat R_{11}\tr \setminus (\hat{\boldsymbol \eta}_{1} -R_{01}\tr\: {\bf d_0})\;.
\label{eq:L22Omega}
\end{align}
%

In order to update the $R$ factor and the r.h.s. vector $\bf d$, it is possible to use \eqref{eq:L22Omega0} and \eqref{eq:L22Omega}, respectively. Nevertheless, without a proper ordering, $R$ will quickly become dense, slowing down the computation. It is well known that $\Lambda$ can be reordered to reduce the fill-in. This has one major disadvantage that the factor $R$ changes completely with the new ordering, impeding the incremental factorization.
 The solution is to only calculate a new ordering for the parts of $R$ which are being affected by the update. \cite{Polok13rss} shows how an efficient incremental ordering can be obtained by considering a partial ordering on a sub matrix of $\hat\Lambda$, which is slightly larger than $\hat\Lambda _{11}=\Lambda_{11}+\Omega$ and which satisfies the conditions of being square and not having any nonzero elements above or left from it. This guarantees that the ordering heuristics such as approximate minimum degree (AMD) will have information about the nonzero entries in $\hat\Lambda_{10} = \hat\Lambda_{01}\tr$, which would otherwise cause unwanted fill-in. A similar fluid reordering approached was introduced in  \cite{Kaess11icra}, and was obtained by applying partial elimination on a Bayes tree data-structure which is a graph representation of the factorized matrix $\Lambda$. In contrast, our proposed technique operates directly on the sparse block-matrix avoiding matrix-graph conversions. 

Once the new ordering is calculated, a resumed factorization can be performed. The column, left-looking Cholesky calculates one column of the factor at a time, while only reading the values left to it. This algorithm can be used to ``resume'' the factorization of the right part of $R$ while only using the reordered part of $\Lambda$ and the unchanged part of the factor, $R_{00}$. The advantage of this approach is the overall simplicity of the incremental updates to the factor, while also saving substantial time by avoiding recalculation of $R_{00}$.

Back substitution is used to obtain the correction $\boldsymbol \delta$ and further the solution $\boldsymbol \theta$ from the updated $\hat R$ and the r.h.s. $\hat{\bf d}$. Maintaining a system representation updated with the new observations and variables every step and solved incrementally without affecting the quality of the estimation can highly increase the efficiency of the online MLE. Nevertheless, in many applications the uncertainty of the estimation is required. The following section describes how the required elements of the covariance matrix can also be calculated incrementaly.

\begin{figure*}[ht]
\includegraphics[width=1.0\linewidth]{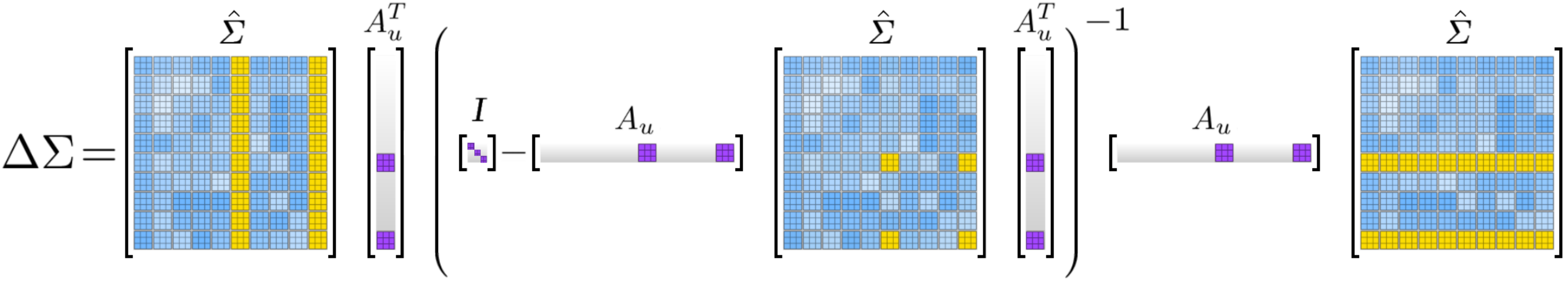}
\caption{Sparsity pattern of the matrices involved in calculating the increment on $\Sigma$ (best viewed in color).}
\label{fig:sparsityPattern}
\end{figure*}

\subsection{Covariance Recovery}\label{Subsec:Covariance}
The covariance matrix of a system is given by the inverse of the system matrix $\Sigma = \Lambda\inv$. For large systems, such inversion is prohibitive, since the result is a dense matrix. Nevertheless, most of the applications require only a few elements of the covariance matrix, eliminating the need for recovering the whole $\Sigma$. In general, the elements of interest are the block diagonal and the block column, corresponding to the last pose. Some other applications only require a few block diagonal and off-diagonal block elements.
In \cite{Bjorck96book,Golub1980laa}, it was shown how specific elements of the covariance matrix can be recursively calculated from the $R$ factor and \cite{Kaess09ras} shows a practical implementation of this formula.
For computation of multiple elements of the covariance matrix, such as the 
block diagonal, the recursive computation becomes efficient only if all the intermediate results are stored.

In \autoref{Subsec:IncUp}, we mentioned that most of the algorithmic speed-ups can be applied in case the linearisation point is kept unchanged or changes partialy. Then the effect of every new measurement can easily be integrated into the current system matrix $\Lambda$ by a simple addition (see \eqref{eq:updateLambda}). The matters get complicated when it is required to update its inverse:
\begin{equation}
\hat \Sigma=(\Lambda + A_u\tr\: A_u)\inv\;. \label{eq:upSigmaShort}
\end{equation}
By applying the Woodbury formula, it can be shown that in contrast to the information matrix which is additive, the covariance is subtractive:
\begin{equation}
\hat \Sigma = \Sigma + \Delta \Sigma \;,\; \Delta\Sigma = -\Sigma \: A_u\tr(I + A_u\: \Sigma A_u\tr)\inv A_u\: \Sigma \;. \label{eq:upSigma}
\end{equation}
Here, $S \triangleq I + A_u\: \Sigma A_u\tr$ is a square invertible matrix with the size equal to the rank of the update $A_u$. This rank is usually much smaller than that of $\Lambda$ and thus the cost of calculating this inverse is negligible compared to the full inverse in $\Sigma = \Lambda\inv$. Similarly to \eqref{eq:upSigmaShort}, one can downdate $\hat\Lambda$ to obtain $\Sigma$:
\begin{equation}
\Sigma=(\hat \Lambda - A_u\tr\: A_u)\inv \;,\label{eq:downSigmaShort}
\end{equation}
and by applying the Woodbury formula the increment can now be calculated in terms of the new covariance $\hat\Sigma$:
\begin{equation}
\Delta\Sigma = \hat\Sigma \: A_u\tr(I - A_u\: \hat\Sigma A_u\tr)\inv A_u\: \hat\Sigma\;.
\label{eq:downSigmaLong}
\end{equation}
Defining $U \triangleq I - A_u\: \hat\Sigma A_u\tr$, which is a matrix related to $S$, \eqref{eq:downSigmaLong} becomes:
\begin{equation}
 \Delta\Sigma = \hat B U \inv \hat B\tr \;,\;\text{with}\;\; \hat B = \hat\Sigma \: A_u\tr \;,
\label{eq:downSigmaShort2} 
\end{equation}
where looking at the sparsity pattern in the $\hat\Sigma \: A_u\tr$ product, it becomes apparent that only the block columns of $\hat\Sigma$, corresponding to the nonzero blocks in $A_u$ or the variables $\bf v$ being updated, are required. The sparsity pattern of all the matrices involved in the calculation of the increment is shown in \autoref{fig:sparsityPattern}.
We call this $\hat\Sigma_{\bf v}$ and can thus equivalently write $\hat B = \hat\Sigma_{\bf v} A_u\tr\:$, with
$\hat\Sigma\:_{\bf v}$ obtained by solving \mbox{$\hat \Lambda \hat\Sigma_{\bf v}\tr = I_{\bf v}$} or \mbox{$ \hat R\: \hat\Sigma_{\bf v} = \hat R\invtr I_{\bf v}$}, much like in \eqref{eq:normalEq1}. Here, $I_{\bf v}$ is a square matrix with identity diagonal block on columns corresponding to the variables ${\bf v}$ and zeros elsewhere.

Even though it sounds counterintuitive to compute an increment $\Delta\Sigma$ form the already (albeit partially) incremented value $\hat\Sigma\:_{\bf v}$, it allows us to update the covariance at any step from $A_u$ and $\hat \Lambda$ or $\hat R$, instead of having to store the old $\Lambda$ or $R$ in addition to the old $\Sigma$. In this way, it is not mandatory to update $\Sigma$ at each step: 
when performing an update to $\Sigma$ over several steps, $A_u$ will simply contain all the measurements since $\Sigma$ was last calculated. In this way, the covariance can be calculated incrementally, on demand whenever it is needed. Based on whether or not the linearisation point changed or the number of variables being updated gets very large, the algorithm for calculating the covariance incrementally has two branches: a) calculates sparse elements of the covariance matrix using the recursive formula as introduced in \cite{Golub1980laa}, and b) updates sparse elements of the covariance using the covariance downdate in \eqref{eq:downSigmaShort2}. The detailed algorithm can be found in \cite{Ila15icra}.

\subsection{The Sparsity and the Block Structure}\label{Subsec:SparseBlock}
The problems in robotics are in general \emph{sparse}, which means that the associated system matrix is primarily populated with zeros. Many efforts have been recently made to develop efficient implementations to store and manipulate sparse matrices. CSparse~\cite{CSparse}, developed by Tim Davis,~\cite{Davis06book} is one of the most popular sparse linear algebra libraries. It is highly optimized in terms of run time and memory storage and it is also very easy to use. CSparse stores the sparse matrices in compressed sparse column format (CSC) which considerably reduces the memory requirements and is suitable for matrix operations.

Furthermore, in many estimation problems, the random variables have more than one degree of freedom (DOF). For example, in $3D$-SLAM the poses ${\theta_i} \in {\mathbb R}^6\hookleftarrow \mathfrak se(3)$ have $6$~DOF and the landmarks $l_j \in \mathbb{R}^3$ have $3$~DOF. The associated system matrix can be interpreted as partitioned into sections corresponding to each variable, called \emph{blocks}, which can be manipulated at once. If the number of variables is $n$, the size of the corresponding system matrix is $N\times N$, where $N$ is a sum of the products of the number of variables of each type and their corresponding DOF~\cite{Blanco10se3}.

The block structure and the sparsity of the matrices can bring important advantages in terms of storage and matrix manipulation. Some of the existing implementations rely on sparse block structure schemes.
In g2o~\cite{Kuemmerle11icra}, matrices are represented as a vector column of blocks where each block is row-indexed associative array of matrix blocks. This is similar to sSBA~\cite{Konolige10iros1}, with the exception that g2o blocks can take any size. Notably, neither iSAM~\cite{Kaess08tro} nor iSAM2~\cite{Kaess11ijrr} employ any sparse block matrix representation at all. iSAM uses a modification of (element-wise) sparse compressed column format optimized for incremental element addition. iSAM2 is based on a graph data structure where each node is a variable which can have different size depending on its type. Google's Ceres solver~\cite{CeresSolver} has its own block matrix storage quite similar to g2o, with the difference that the blocks are stored in an array of matrix element values rather than each block separately. However, Ceres implements almost no operations on their block matrices and so those are merely in a role of intermediate storage before converting to compressed sparse column and passing the system to a linear solver such as Cholmod. 
In the existing schemes, the block structure is maintained until the point of solving the linear system. Here is where CSparse~\cite{Davis06book} or CHOLMOD~\cite{Davis97siam} libraries are used to perform the element-wise matrix factorization. Once it has been compressed, it becomes impractical and inefficient to change a matrix structurally or numerically and therefore these implementations need to convert their block matrices to CSC at each linear solving step. 

This motivated us to find efficient solutions for arithmetic operations on sparse block matrices, especially the matrix factorization, as well as solutions to sparse block matrix modification and storage. 
Our recent work maximally exploits the sparse-block structure of the problem. On one hand, the block matrix manipulation is highly optimized, facilitating convenient structural and numerical matrix changes while also performing arithmetic operations efficiently. On the other hand, the block structure is maintained in all the operations including the matrix factorization, variable ordering and covariance recovery, eliminating the cost of converting between sparse elementwise and sparse blockwise representation. Correct manipulation of the \emph{block matrices} enabled very efficient NLS and incremental NLS solutions \cite{Polok13icra,Polok13rss} implmented in SLAM++ library, which outperformed other similar state-of-the-art implementations, without affecting the precision in any way. What is interesting about the block matrix format employed in SLAM++ is the use of template meta-programming for further acceleration using loop unrolling and SIMD instruction sets such as SSE. This is a novel feature in the context of sparse block matrix work which sets our work apart and yields a considerable performance advantage. In this paper, we will show that, based on the previously proposed block-based data structure in \cite{Polok13icra}, we can also efficiently recover the marginal covariance matrices incrementally to be used in a state-based loop closure detection and compact representation of the SLAM problem.

\section{Information Based Compact 3D Pose SLAM}\label{Sec:CompactSLAM}
Pose SLAM is a variant of SLAM where only the robot trajectory is estimated and the sensor measurements are only used to produce relative constraints between the robot poses. In this case the state vector $\boldsymbol \theta = \left[ {\theta_1},{\theta_2} \ldots {\theta_n} \right]$ gathers only the variables corresponding to the robot poses. To reduce the computational cost of the Pose SLAM and to facilitate its application to very large scale problems, we previously introduced an approach that only takes into account highly informative loop closure links and non-redundant poses in an information filtering framework \cite{Ila10tro}.  A more compact representation of the SLAM problem reduces the memory requirements to store the entire state of the robot as well as it is more computationally efficient, since the systems to be solved are small. Note that maintaining the sparsity of the problem is an important factor when generating efficient solutions. In all existing approaches, maintaining the sparsity comes at a price of introducing some approximations but in general those approximations try to minimize the loss of information in the system. 

Filtering is well known to produce less accurate solutions for the SLAM problem, therefore, in this paper, we extend the approach introduced in \cite{Ila10tro} to maximum likelihood estimation. The existing strategies in this direction focus on how to select the measurements and the variables to be removed from the state representation, and on how the removing process unfolds. 

Mutual information of the laser scans is used in \cite{Kretzschmar12ijrr} to decide which measurements and nodes should be removed from the pose graph representation of the SLAM. The problem of removing a variables and measurements form the SLAM state representation is from a graph is slightly different from the problem of deciding on the fly whether or not a a variable or a measurement are added to the state. The former involve variable marginalization which in turn produce a dense rather than sparse state representation. Therefore, the existing approaches resort to local approximations such as Chow-Liu trees to obtain a sparse graph. Several aspects on how to apply graph sparsification by marginalization in the context of MLE SLAM are discussed in \cite{Carlevaris14tro}. \cite{Johannsson13icra}, on the other hand, introduced an incremental strategy which avoids marginalization and which is similar to the one proposed in \cite{Ila10tro} and its extension to MLE-SLAM proposed in this paper, with the difference that their method requires relocalization when revisiting parts of the map and the fact that it is applied in the context of visual SLAM. 

The strategy introduced in this paper is incremental and therefore suitable for online SLAM, has a general application to any type of pose SLAM problems, can be easily extended to landmark SLAM or SFM, it is complete in the sense that provides not only the way to calculate the distance and information measures but also the corresponding methods to obtain the required thresholds. 
The compact representation is achieved by computing two measures, the proximity in terms of sensor range of current pose to any other previous poses and the information gain for each candidate link. Apart from maintaining a compact representation of the state, the calculation of the proximity of two poses in terms of sensor range can provide a more efficient loop-closure detection than the actual registration of all the sensor readings or even the appearance-based techniques.

\begin{figure}[t]
\includegraphics[trim=25mm 2mm 20mm 10mm, clip, width=1.0\linewidth]{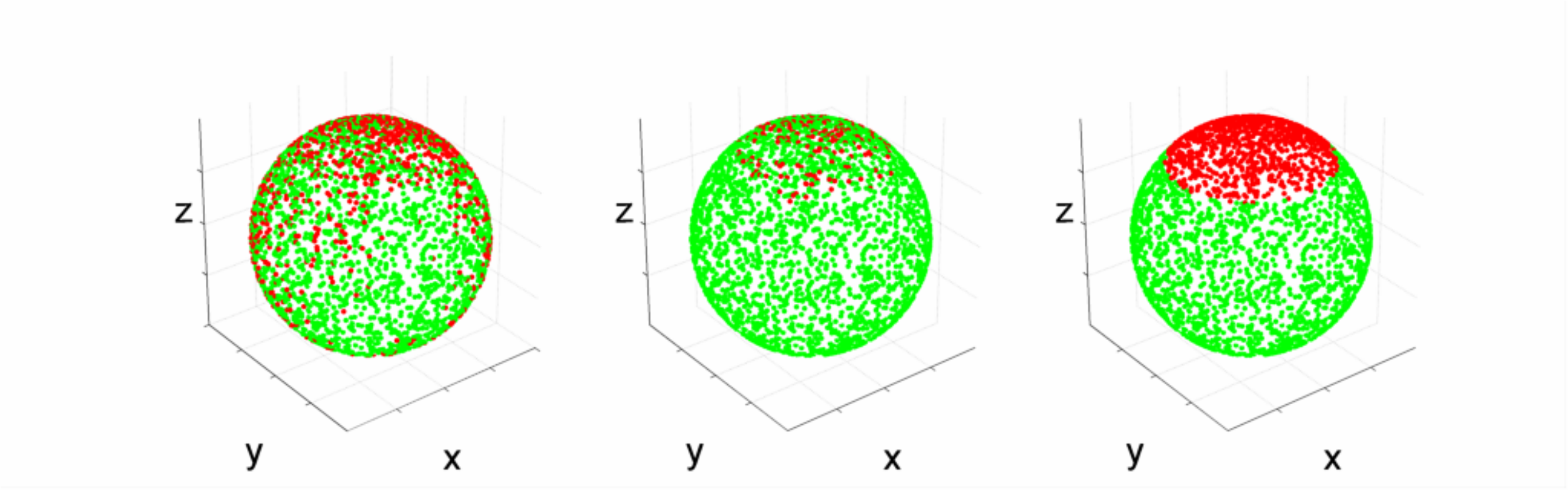}
\caption{Rotation threshold applied to 5000 normally distributed random rotations. Each point is a rotated view direction vector; roll is however not visible. The z+ axis is the forward direction. Distance threshold is applied to: left) the elements of the axis-angle vector, middle) the rotation angle (note the dependence on roll) and right) the angular change of view direction. The angular threshold is $\frac{\pi}{4}$ in all cases.}
\label{fig:rotThreshold}
\end{figure}
%

\subsection{Distance Measure}\label{SubSec:Distance}
A measure of proximity of two poses is the relative displacement calculated from the current estimation~\cite{Ila10tro}. In the following formulations, the vectors corresponding to a single state variable, $\theta_i$, are demoted in lower case to be consistent with the previous notation. The scalar elements of those vectors will be marked with corresponding subscripts. In an online application, the displacement between the current pose,~${\theta_n}$, to any other previous pose in the trajectory,~${\theta_i}$, can be estimated as a Gaussian with parameters:
\begin{align}
\mu_d&= D(\mu_i, \mu_n)\\
{\Sigma}_d&= [J_i\:J_n]\: \left[ \begin{array}{cc} {\Sigma}_{ii} & {\Sigma}_{in} \\ {\Sigma}_{in}\tr & {\Sigma}_{nn} \end{array} \right] [J_i \: J_n]\tr\;, \label{eq:SigmaEdge}
\end{align}
where $D(\cdot)$ calculates the relative displacement between the mean estimates of the two poses, ${\Sigma}_{ii}$ and ${\Sigma}_{nn}$ are the marginal covariances and $ {\Sigma}_{in}$ is the cross correlation between the $i^{th}$ and the current ($n^{th}$) pose.

In an online application, at each step, one can test the proximity of the current pose with any of the previously visited poses to determine if there is a possibility of the sensor range overlap. This can be obtained by calculating the probability of a pose ${\theta_i}$ being closer than $v$ to
the pose ${\theta_n}$ along each dimension, $v$ being the sensor range. We marginalize the distribution
on the displacement for each of its dimensions, $r$ to get a one-dimensional Gaussian distribution $\mathcal{N}(\mu_r,\sigma_r^2)$ that
allows to compute the probability:
\begin{align}
p_r&=\int_{-v_r}^{+v_r} \: \mathcal{N}(\mu_r,\sigma_r^2) \nonumber \\
&=\frac{1}{2} \left( \text{erf}\left( \frac{v_r-\mu_r}{\sigma_r\:\sqrt{2}} \right) - \text{erf}\left(\frac{-v_r-\mu_r}{\sigma_r\:\sqrt{2}} \right) \right) \;.
\label{eq:similarity}
\end{align}
If, for all dimensions, $p_r$ is above a given threshold, $s$, then the pose ${\theta_i}$ is considered close enough to the current robot pose,~${\theta_n}$.
We include ${\theta_n}$ in the state only if no other poses in the representation are close to it.

The thresholds $v$ are derived from the sensor characteristics: the field of view for cameras, the maximum distance for the laser scan alignment, etc.. In general,
it is simpler to define a threshold for each dimension separately than to define a single threshold for a measure integrating the distances along all dimensions (e.g., a weighted norm). Therefore, separate thresholds for translation and rotation are defined.

The translation component of the $v$ threshold can be easily determined from the sensor range. For the rotational component, however, there are several ways to define the threshold. Given the rotation component $q_i = \mu_i(4:6)$ and $q_j=\mu_j(4:6)$ then $q_d = \mu_d(4:6)$ is the relative rotation of the two poses, represented by an axis angle vector, obtained by taking unit length axis of rotation and multiplying it by the rotation angle in radians. 
The easiest is to calculate the probability of each element in $q_d$ to be below a threshold $v_a$ but this is incorrect due to the strong correlation between the elements of the rotational components. This can be seen in \autoref{fig:rotThreshold}, left.
A correct way is to compute the probability of the magnitude of the relative rotation $|q_d|$ to be smaller than a threshold $v_a$, \autoref{fig:rotThreshold}, middle. In this particular application though, we do not want just to limit the relative rotation but to see if the fields of view of the sensor overlap. Therefore, in the case of using cameras, a more permissive threshold can be considered, a threshold invariant to roll.
We can compute the probability of the angle of the relative \emph{view direction rotation} to be smaller than a threshold $v_a$. 
This threshold is shown in \autoref{fig:rotThreshold}, right. 

The last one, although being desirable, involves calculation of nontrivial Jacobians. Since we deal with uncertainty in the estimate, the probability depends on the marginal covariances. These covariances need to be transformed from covariances on the state space, $\mathbb{R}^6 \times \mathbb{R}^6$, where the 6 DOF are $\left[x, y, z \right]$ and a $\mathbb{R}^3$ axis angle rotation, to the threshold space, $\mathbb{R}^4 \times \mathbb{R}^4$, where the 4 DOF are $\left[dx, dy, dz\right]$ and a (scalar) angle of the relative view direction rotation. 

To calculate the view direction vectors, one can convert the rotations from axis angle representation to rotation matrices, where the columns of the matrix give the directions of the principal axes of the coordinate frame, \mbox{$Q_i = \textproc{Rot}(q_i)$} and \mbox{$Q_j = \textproc{Rot}(q_j)$}. Assuming that the view direction coincides with the 'z+' axis, the view directions are \mbox{$d_i = Q_i(1:3, 3)$} and \mbox{$d_j = Q_j(1:3, 3)$}. Finally, the view direction angle is \mbox{$\alpha = \arccos(d_i \cdot d_j)$}. This can also be calculated directly from the relative rotation as:
\begin{align}
\alpha &= \arccos(\textproc{Rot}(r_d)(1:3, 3) \cdot \left[0\: 0\: 1\right]\tr) \nonumber \\
&= \arccos(\textproc{Rot}(r_d)(3, 3)) \;.
\end{align}
By expanding Rodriguez' rotation formula, we can see that:
\begin{equation}
\textproc{Rot}(r_d)(3, 3) = \cos(\|r_d\|) + (1 - \cos(\|r_d\|)) \frac{r_d(3)^2}{\|r_d\|^2}\;.
\end{equation}

This finally gives us the lower dimension (4 DOF) distance \mbox{$\hat \mu_d = \left[\mu_d(1:3), \textproc{Rot}(\mu_d(4:6))(3, 3)\right]$}. The Jacobian of this transformation is $J_{tr} = \frac{\partial \hat \mu_d}{\partial \mu_d}$, which is a $4 \times 6$ matrix. The covariance is then transformed with $\hat\Sigma_d = J_{tr} \Sigma_d J_{tr}\tr$ and the result is a $4 \times 4$ matrix. 
Distance threshold can be applied in this space in order to determine whether the poses have overlapped field of view and in consequence loop closure links can be obtained.

\subsection{Information Measure}\label{SubSec:Information}
The mutual information quantifies the Entropy reduction in the system after the integration of an observation.
For Gaussian distributions, it is given by the logarithm of the ratio of determinants of
prior and posterior state covariances~\cite{Dissanayake02ar,Sim05icra,Ila10tro}.
In \cite{Ila10tro} has been shown that by algebraically manipulating this ratio of determinants, one can easily obtain the mutual information from the uncertainty of the observation, $\Sigma_k$ and $S = I + A_u\: \Sigma A_u\tr$ the innovation matrix:
\begin{align}
\mathcal{I}=\frac{1}{2}\:\ln \frac{|\Lambda + A_{u}\tr\:A_{u}|}{|\Lambda|} = \frac{1}{2}\:\ln | \Sigma_{k}\inv | \cdot |S| \;.\label{eq:infoSimple}
\end{align}
In case of pose SLAM, an observation is given by a relative transformation between two poses. Therefore, the estimated mutual information can be written in terms of
the uncertainty of the observation, $\Sigma_k$ and the uncertainty of the edge $\Sigma_d$:
\begin{align}
\mathcal{I}= \frac{1}{2}\:\ln | \Sigma_{k}\inv | \cdot |\Sigma_{k} + \Sigma_{d}| \;,\label{eq:infoCompactSLAM}
\end{align}
where $\Sigma_d$ is calculated as in \eqref{eq:SigmaEdge}. In this way, the mutual information of a potential observation can be estimated by specifying the marginal covariances of the poses involved in the observation and using an initial guess for the uncertainty of the observation. After sensor registration, the exact uncertainty of the observation is known and the mutual information of the link can be evaluated precisely.

\renewcommand{\algorithmicindent}{.375cm} 

\begin{algorithm}[t!] 
\caption{\label{alg:compactSLAM} Incremental Compact SLAM Estimation}
\begin{algorithmic}[1]
\Require thresholds: $v$, $s$, $g_{pose}$, and $g_{loop}$
\Require expected sensor covariance: $\bar\Sigma_{y}$
\Require initial state: ${\boldsymbol \theta}_0, \Sigma_0$
\State $ ({\boldsymbol \theta},{S}) = \Call{InitSystem}{{\boldsymbol \theta}_0, \Sigma_0}$ 
\State $keepPose = \textsc{True}$ 
\State $n=1$
\State $I_{n-1} = \Call{GetNextData}{}$
\While {$I_n = \Call{GetNextData}{}$}
	\State $(\mu_u,{\Sigma}_u) = \Call{Registration}{I_n,I_{n-1}}$ \label{algline:sensorReg} 
	\If{$keepPose$} 
		\State $(\mu_e,\Sigma_e) = (\mu_u,\Sigma_u)$ 
		\State $({\boldsymbol \theta},{S},{\Sigma}_{M}) = \Call{IncUp}{{\boldsymbol \theta},{S},{\Sigma}_{M}, ({\mu}_e, \Sigma_e})$ \label{algline:incUp} 
	\Else
		\State $(\mu_e,\Sigma_e) = \Call{ConcatenatePose}{(\mu_e,\Sigma_e),(\mu_u,\Sigma_u)}$ \label{algline:poseConcat} 
		\State $({\boldsymbol \theta},{S},{\Sigma}_{M}) =\Call{ReplUp}{{\boldsymbol \theta},{S},{\Sigma}_{M}, ({\mu}_e, \Sigma_e})$ \label{algline:overUp}
	\EndIf
	
	\State $C = \Call{SearchLoopClosure}{{\boldsymbol \theta},{\Sigma}_{M}, v, s}$ \label{algline:searchLoopClosures}
	\State $\mathcal{I} = \Call{MInfo}{C,{\boldsymbol \theta},{\Sigma}_{M}, \bar\Sigma_{y}}$ \label{algline:mutualInfo}
	\State $loopClosed = \textsc{false}$
	\While {$C \neq \emptyset$} 
		\State $i= \Call{argmax}{\mathcal{I}}$
		\If{$C(i)<n-1$ {\bf and} $\mathcal{I}(i) > g_{loop}$}
			\State $(\mu_y,{\Sigma}_y) = \Call{Registration}{I_n,I_{C(i)}}$ 
			\If{$\textsc{NotVoid}(\mu_y)$}
				\State $\mathcal{I}(i)= \Call{MInfo}{C(i),{\boldsymbol \theta}_{C(i),n},{\Sigma}_{{M}_{C(i),n}}, \Sigma_{y}}$ \label{algline:redoMInfo} 
				\If{$\mathcal{I}(i)>g_{loop}$} \label{algline:loopGainTest}
					\State $({\boldsymbol \theta},{S},{\Sigma}_{M}) = \Call{IncUp}{{\boldsymbol \theta},{S},{\Sigma}_{M}, ({\mu}_e, \Sigma_e})$ \label{algline:incUpAfterLoop}
					\State $\mathcal{I} = \Call{MInfo}{C,{\boldsymbol \theta},{\Sigma}_{M}, \bar\Sigma_{y}}$ \label{algline:mutualInfoUpdate} 
					\State $loopClosed = \textsc{True}$
				\EndIf
			\EndIf
		\EndIf
		\State $(C,\mathcal{I} ) = (C,\mathcal{I} ) \setminus \{i\}$
	\EndWhile
	\State $keepPose = (loopClosed$ {\bf or} $\Call{min}{\mathcal{I}} > g_{pose})$ \label{algline:toBeOrNotToBe} 
	\State $n= n+1$
\EndWhile
\end{algorithmic}
\end{algorithm}

\subsection{Compact 3D Pose SLAM -- The algorithm}\label{Sec:AlgCase1}
This section describes the algorithm to obtain a compact representation of the 3D pose SLAM problem in the context of maximum likelihood estimation. The algorithm is meant for online applications, therefore it involves incremental estimation strategies. The efficient incremental solving summarised in \autoref{Subsec:IncSolve} and detailed in \cite{Polok13rss} is used together with incremental covariance recovery summarised in \autoref{Subsec:Covariance} and detailed in \cite{Ila15icra}. The two measures introduced above, the estimated relative transformation between two poses and the mutual information are used to select only those poses which are relevant and those edges that are informative. The result is an incremental algorithm for compact pose SLAM, which automatically maintains a sparse representation of the state in the information space. Its efficiency comes from its sparsity, the incremental computations and the fact that all the matrix computations are done by blocks. 

The algorithm is detailed in \autoref{alg:compactSLAM}. It requires four parameters, the sensor range, $v$, the probability to accept a pose as having overlapping field of view with the current one, $s$, and the minimum information gain to add poses, $g_{pose}$, and to close loops, $g_{loop}$.

The algorithm starts with an initial state and loops while there are measurements (edges) to be integrated into the system. The measurements are relative transformations between the robot poses obtained by registering either images or laser scans (line \ref{algline:sensorReg}). Registration function can also account for other sensors such as odometry or IMU. In case that the previous pose was not integrated into the system, the new relative transformation is concatenated to it at line \ref{algline:poseConcat}. The concatenation is done in both mean and covariance space. The mean is obtained by pose composition and the covariance of the composed poses can be obtained as in \cite{Smith86ijrr} and requires the Jacobians of the pose composition function, or as in \cite{Barfoot14tro} to handle the uncertainty of the $SE(3)$ pose composition. 
Correct computation of the composed measurement covariance is important to yield conservative estimates compared to the full solution, otherwise the resulting estimates could end up being  overconfident (low covariance) and potential loop closures would be lost, or conversely not confident enough (high covariance) which would deteriorate performance by performing too many unnecessary sensor registration attempts. The method used in our approach leads to a conservative estimate, as demonstrated is \autoref{SubSec:CovNorms}.
The resulting transformation updates the system either by incrementing the state (line \ref{algline:incUp}) in case that the pose is to be kept in the system, or by replacing the previous pose in case that it is deemed redundant (line \ref{algline:overUp}).

At line \ref{algline:searchLoopClosures}, the algorithm searches for potential loop-closures by applying the distance test introduced in \autoref{SubSec:Distance}. The search returns $C$, a set of candidate pose ids. Before proceeding with the sensor registration, the relevance of the possible links that can be established with the candidates $C$ is determined by calculating the mutual information as in \eqref{eq:infoCompactSLAM}, at line \ref{algline:mutualInfo}. In here, an estimate of the sensor noise, $\bar \Sigma_{y}$, is used -- the actual value is only available after the registration (and then the mutual information is updated, line \ref{algline:redoMInfo}). Nevertheless, given known sensor characteristics, in practice, using an estimated value offers good means to select the relevant links.

Starting with the most informative candidate, the algorithm performs sensor registration to obtain the link that will update the system. Observe that the mutual information of each of the remaining candidates is recalculated after the update (line \ref{algline:mutualInfoUpdate}), and eventually only a few or no further candidates are registered. This is due to the fact that informative links substantially change the entropy of the system, and the other candidates become irrelevant.

At line \ref{algline:toBeOrNotToBe}, the algorithm decides whether or not the current pose is added to the system. A pose is added to the system if it is part of an informative link or if the possible links it can establish are informative. In this way the algorithm manages to maintain a set of poses and links uniformly distributed in the information space. Tests on real and simulated datasets described in the next section will show how this strategy considerably reduces the computational costs while maintaining a good accuracy of the estimate. Note that only poses that do not involve possible informative loop closure are removed, therefore marginalizing out those poses can be done by simply concatenating the edges that involve the removed pose. This has the advantage that marginalization is guaranteed to not introduce extra-edges in the graph and therefore, there is no requirement for complementary strategies to sparsify the resulting subgraph as in \cite{Carlevaris13icra}.

The main difference between the algorithm proposed in this paper and the one in \cite{Ila10tro} is the fact that in here the state pruning is performed in a MLE framework, whereas in \cite{Ila10tro} it was integrated in a filtering approach. Filtering is, in general, much simpler than the current incremental approach, mainly because the linearisation point stays fixed. Nevertheless, this is also an important source of errors in the estimation. The current strategy keeps this error low by updating the linearisation point when needed. Another difference between the two approaches is that the latter is applicable to any variable dimensions (2D or 3D SLAM) as well as it is easy to extend to landmark SLAM or even structure from motion problems.


\begin{figure*}[t]
\begin{center}
\begin{tabular}{c}
\begin{minipage}{.5cm}
	\centering
	a)
\end{minipage}
\begin{minipage}{3.2cm}
	\centering
	\includegraphics[width=2.8cm]{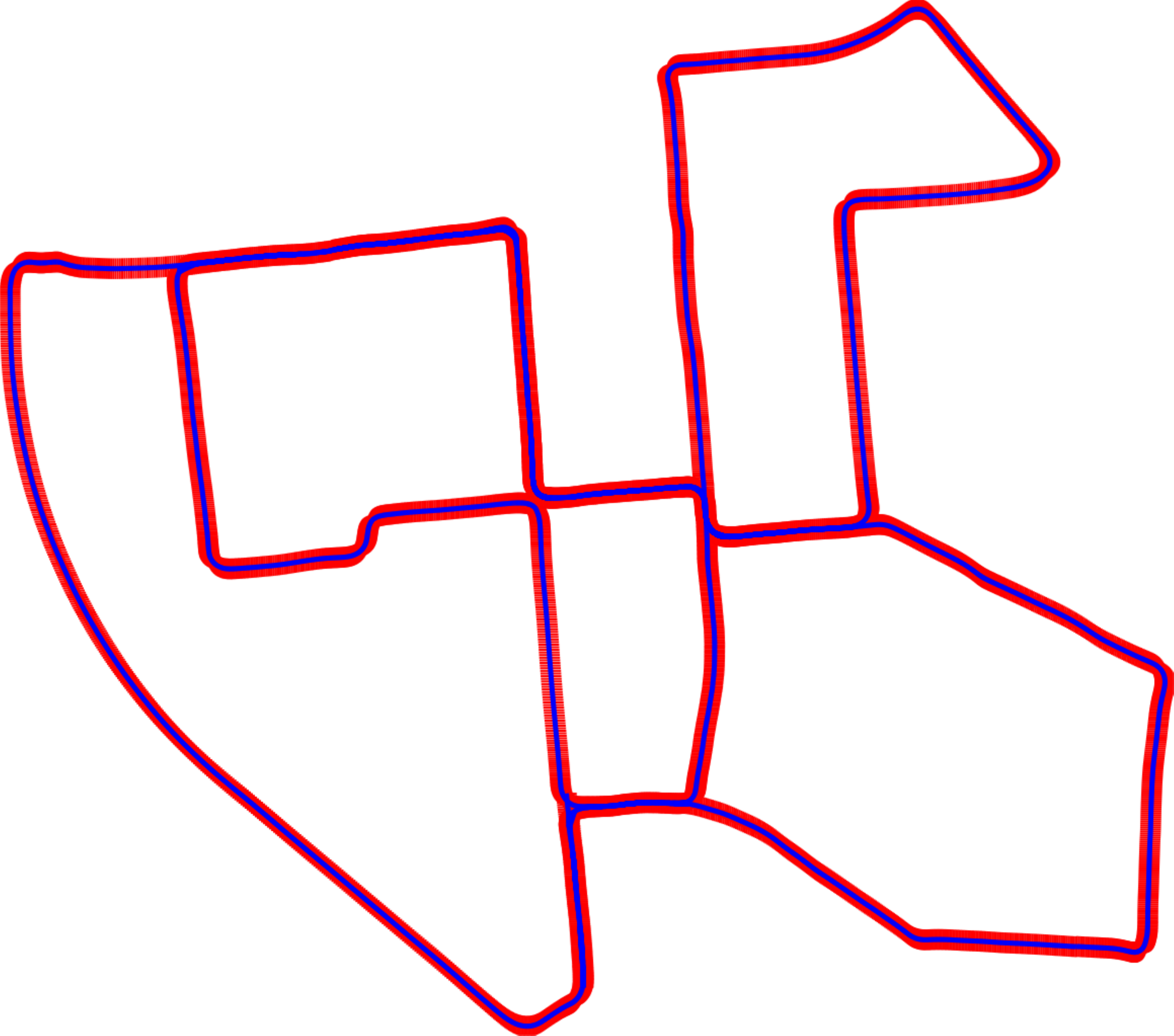} 
\end{minipage}\hfill
\begin{minipage}{3.2cm}
	\centering
	\includegraphics[width=2.8cm]{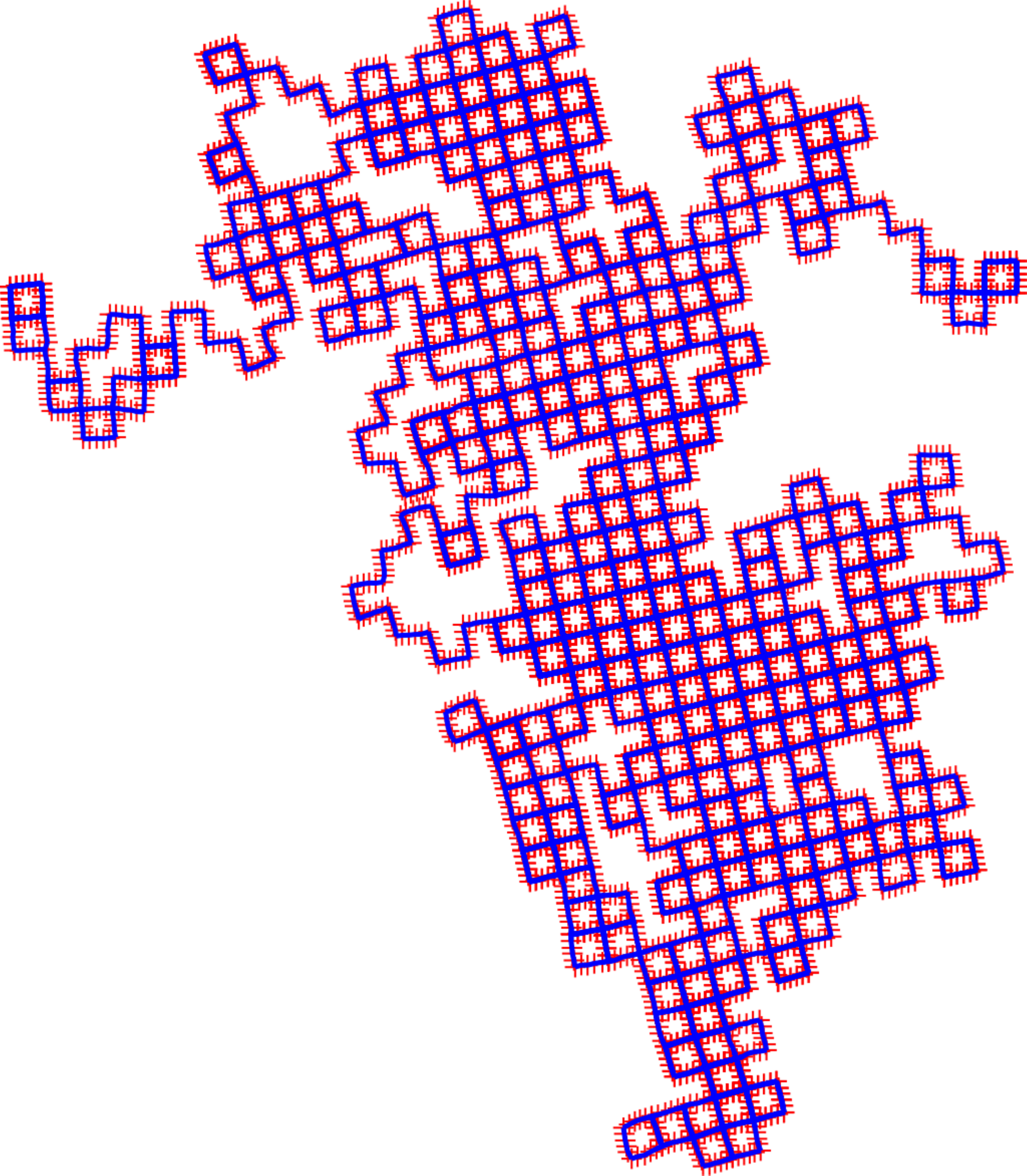}
\end{minipage}\hfill
\begin{minipage}{3.2cm}
	\centering
	\includegraphics[width=2.8cm]{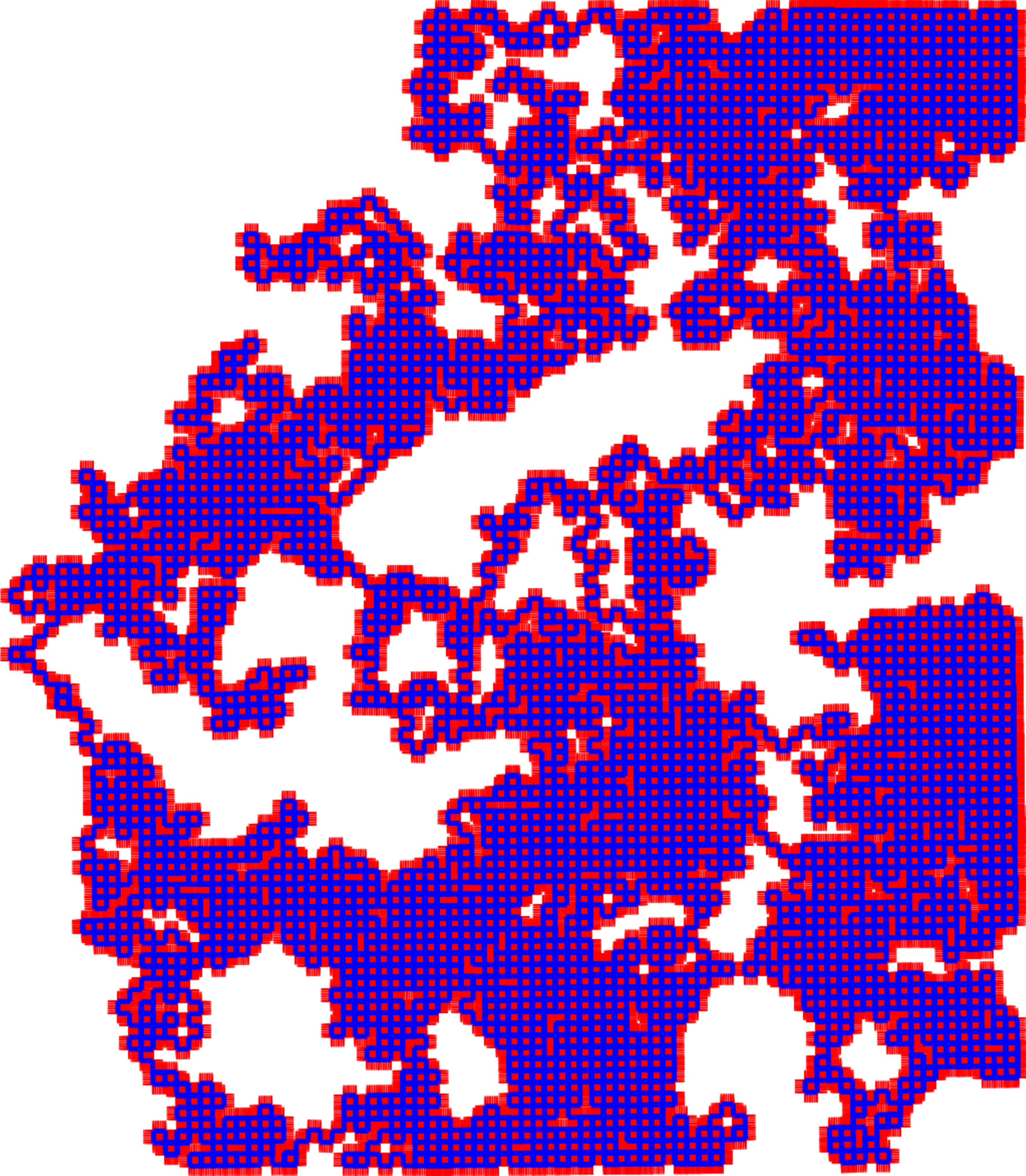} 
\end{minipage}\hfill
\begin{minipage}{3.2cm}
	\centering
	\includegraphics[width=2.8cm]{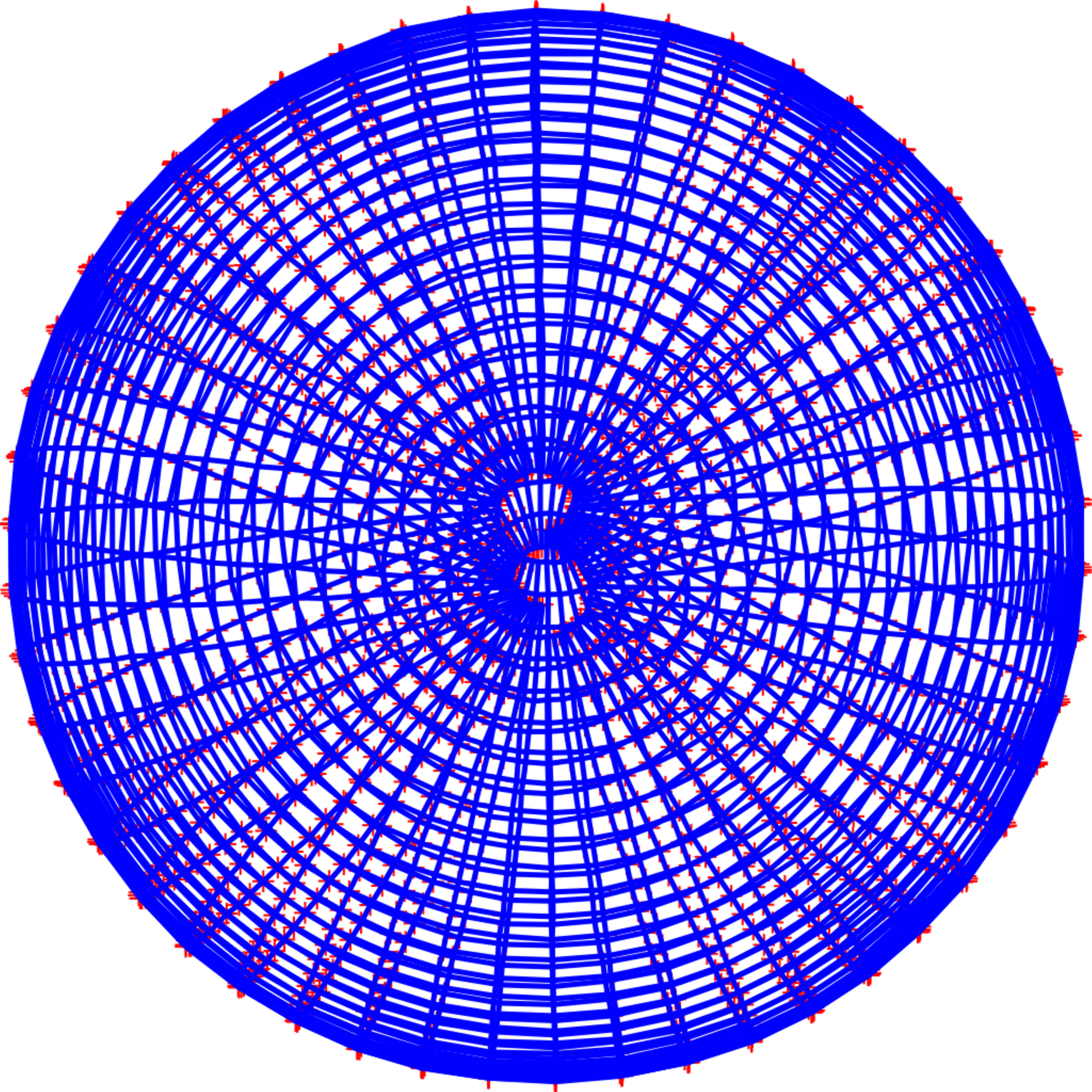}
\end{minipage}\hfill
\begin{minipage}{3.2cm}
	\centering
	\includegraphics[width=2.8cm]{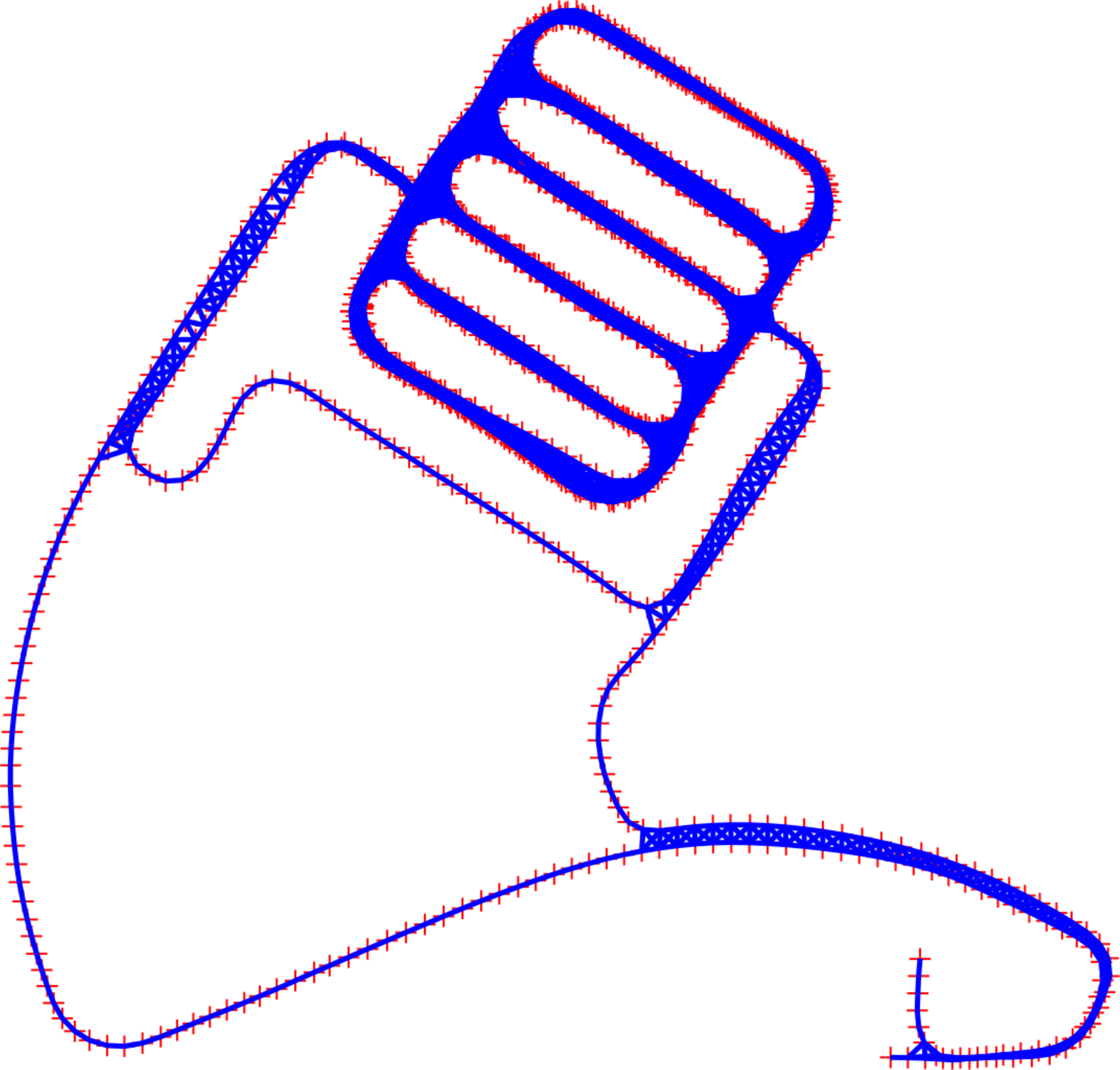}
\end{minipage}\hfill
\\
\begin{minipage}{.5cm}
	\centering
	b)
\end{minipage}
\begin{minipage}{3.2cm}
	\centering
	\includegraphics[width=2.8cm]{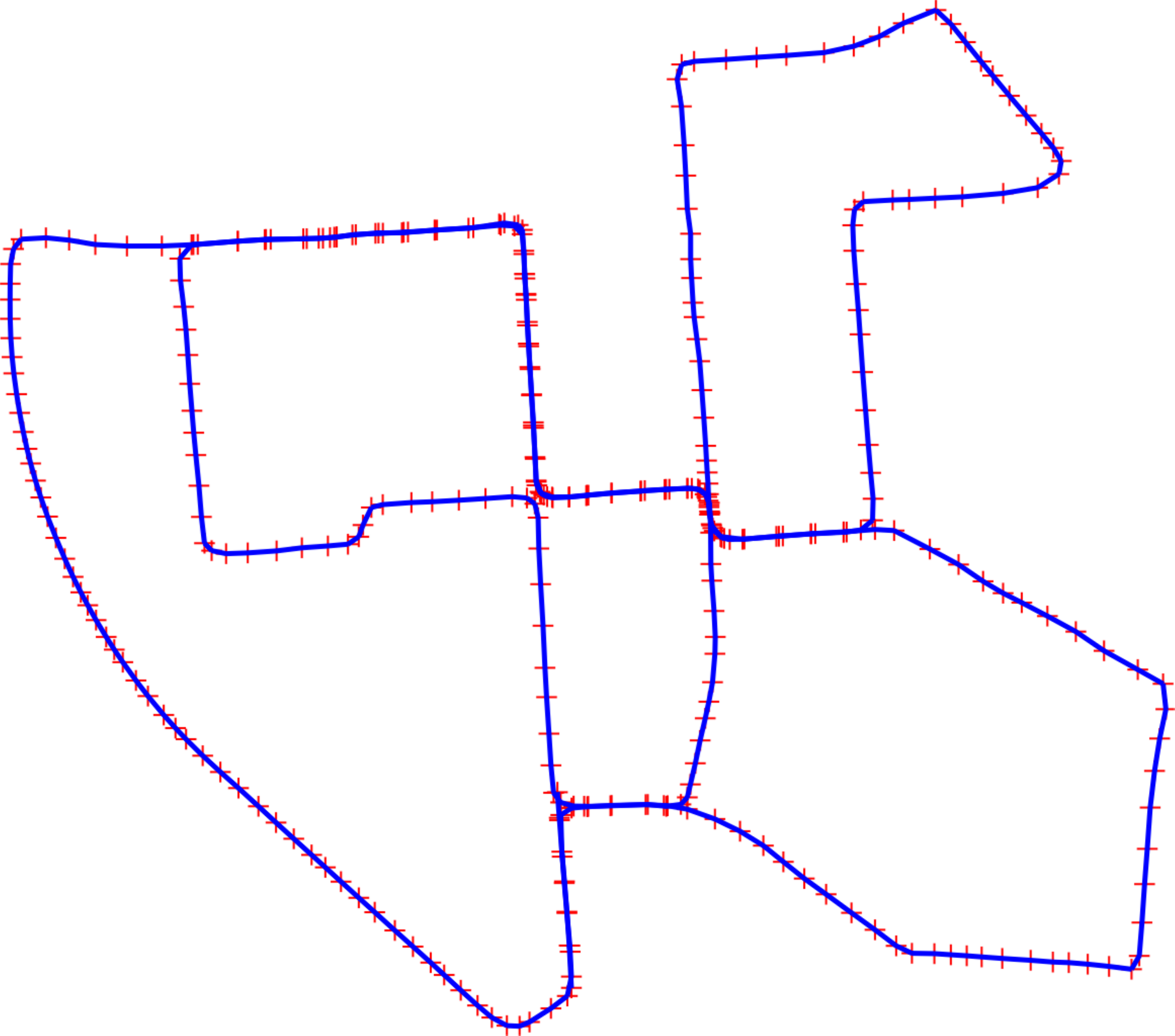}  
\end{minipage}\hfill
\begin{minipage}{3.2cm}
	\centering
	\includegraphics[width=2.8cm]{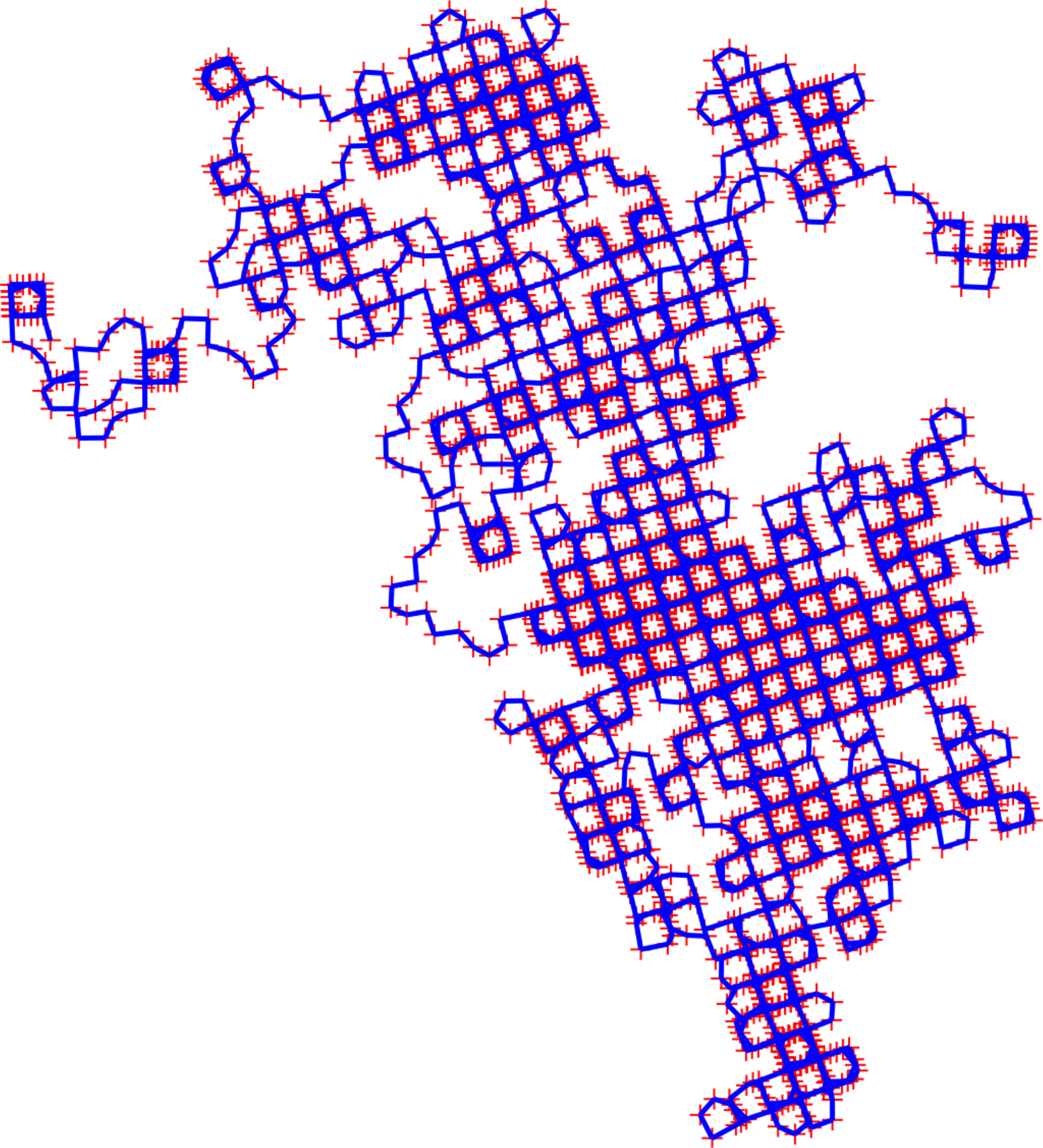}
\end{minipage}\hfill
\begin{minipage}{3.2cm}
	\centering
	\includegraphics[width=2.8cm]{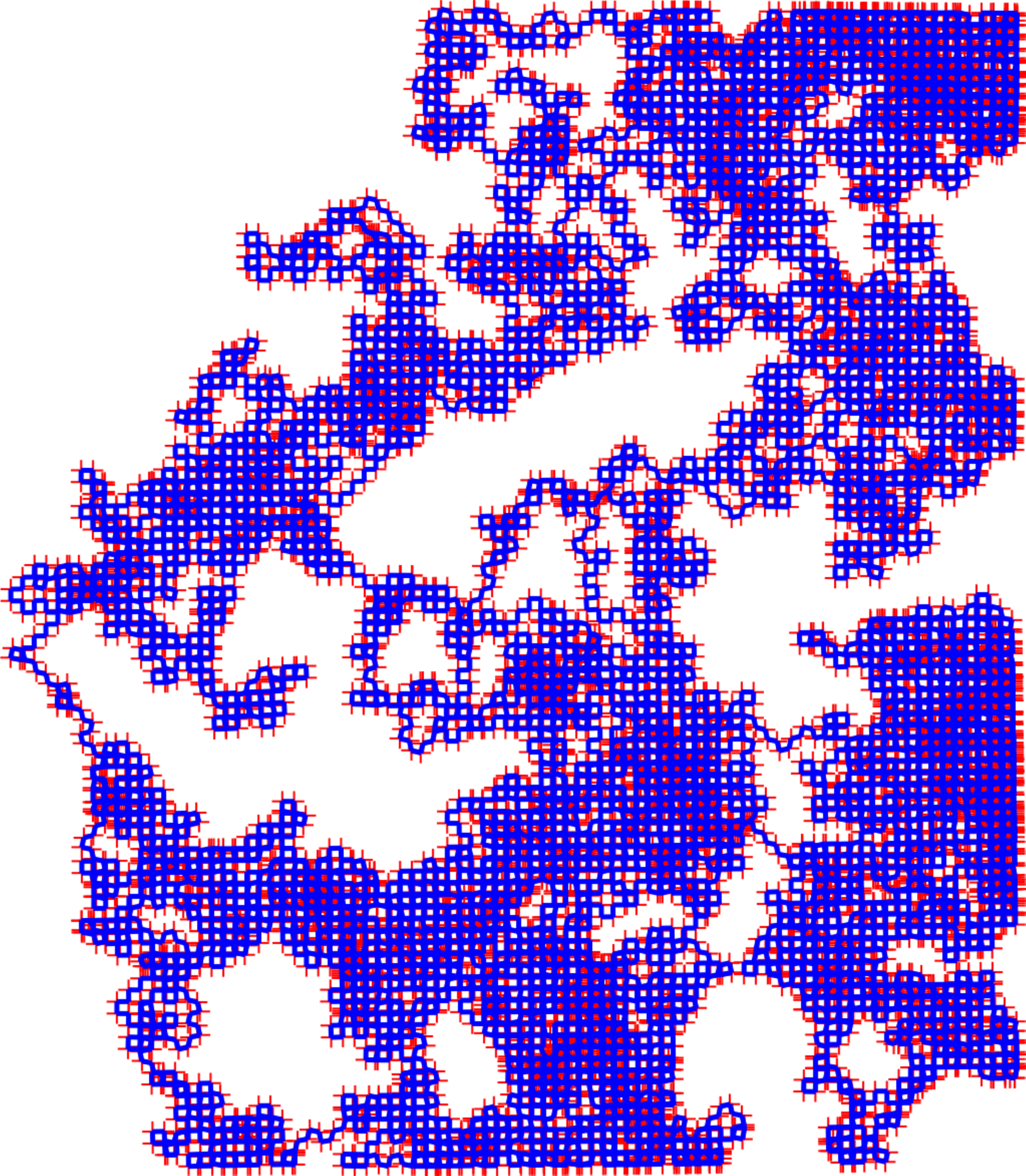} 
\end{minipage}\hfill
\begin{minipage}{3.2cm}
	\centering
	\includegraphics[width=2.8cm]{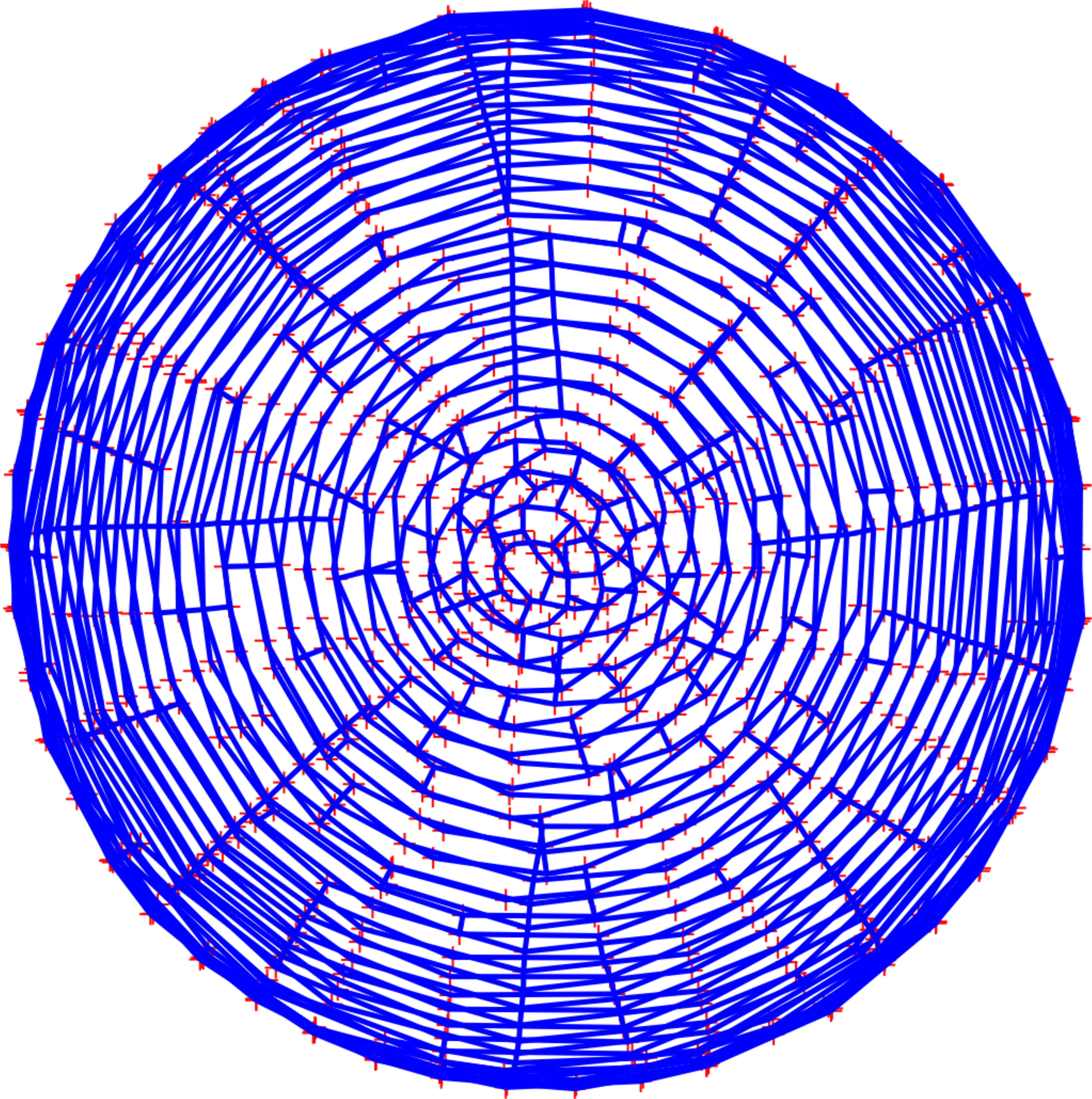}
\end{minipage}\hfill
\begin{minipage}{3.2cm}
	\centering
	\includegraphics[width=2.8cm]{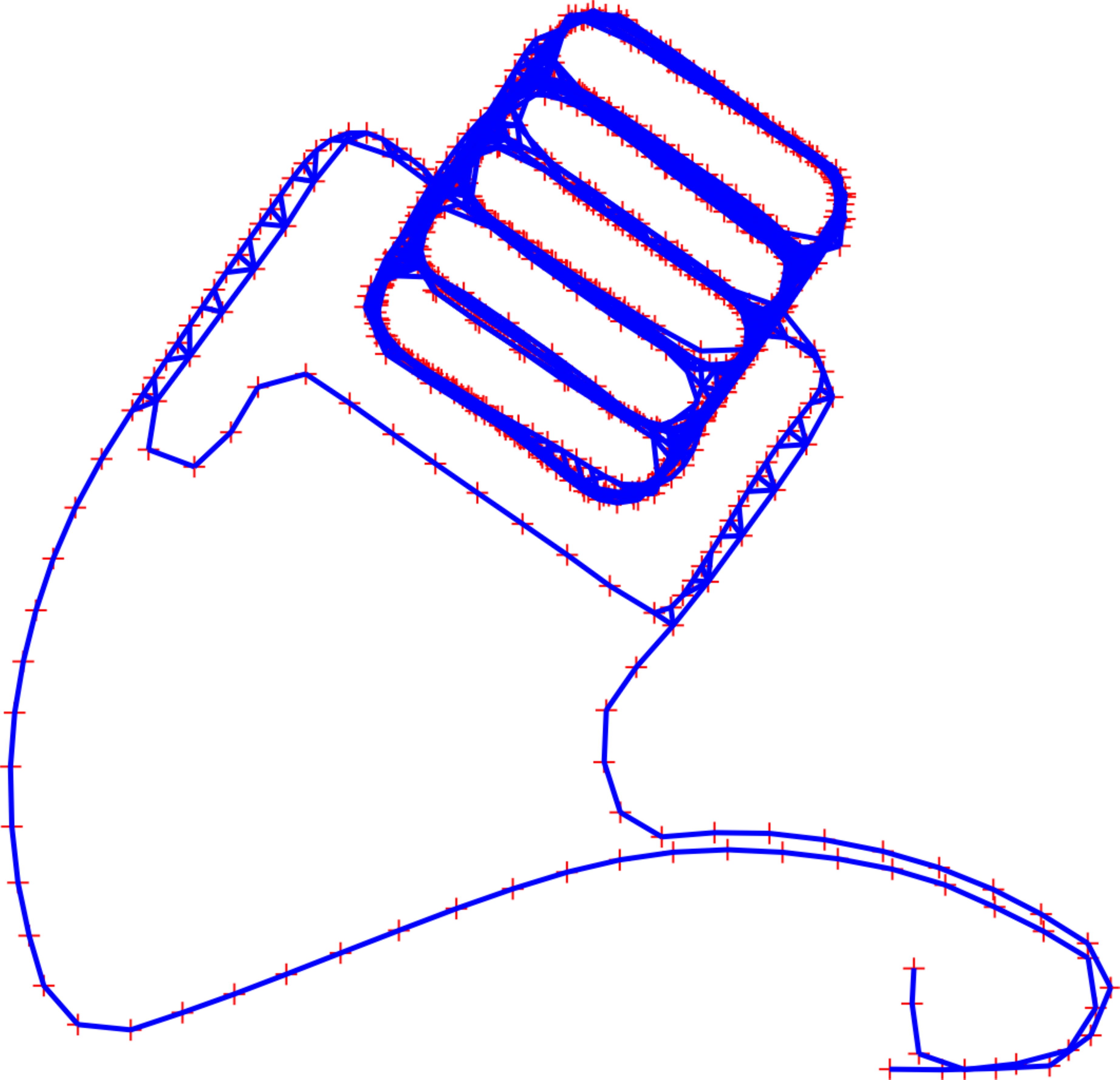}
\end{minipage}\hfill
\\
\begin{minipage}{.5cm}
	\centering
	c)
\end{minipage}
\begin{minipage}{3.2cm}
	\centering
	\includegraphics[width=2.8cm]{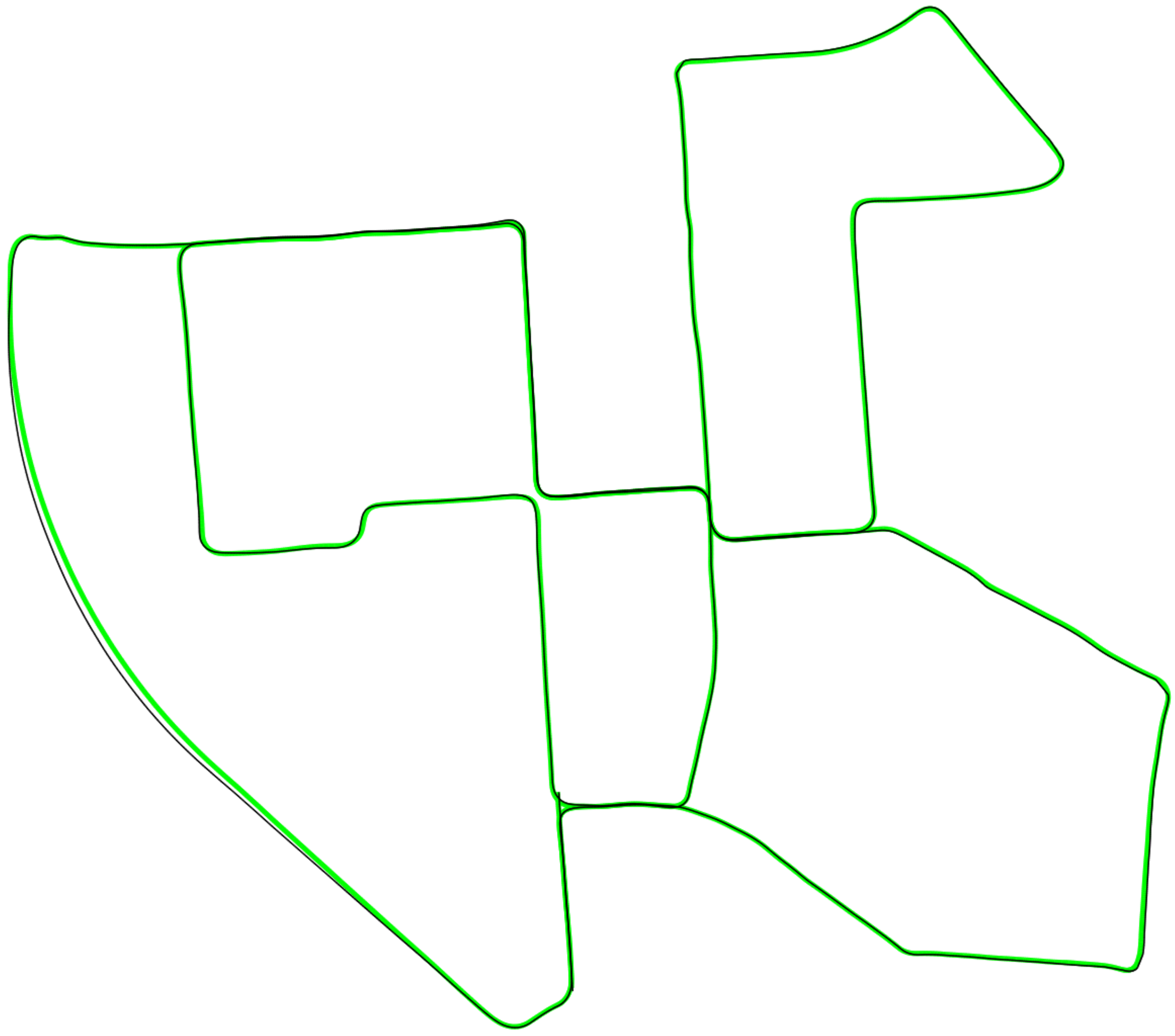}  
\end{minipage}\hfill
\begin{minipage}{3.2cm}
	\centering
	\includegraphics[width=2.8cm]{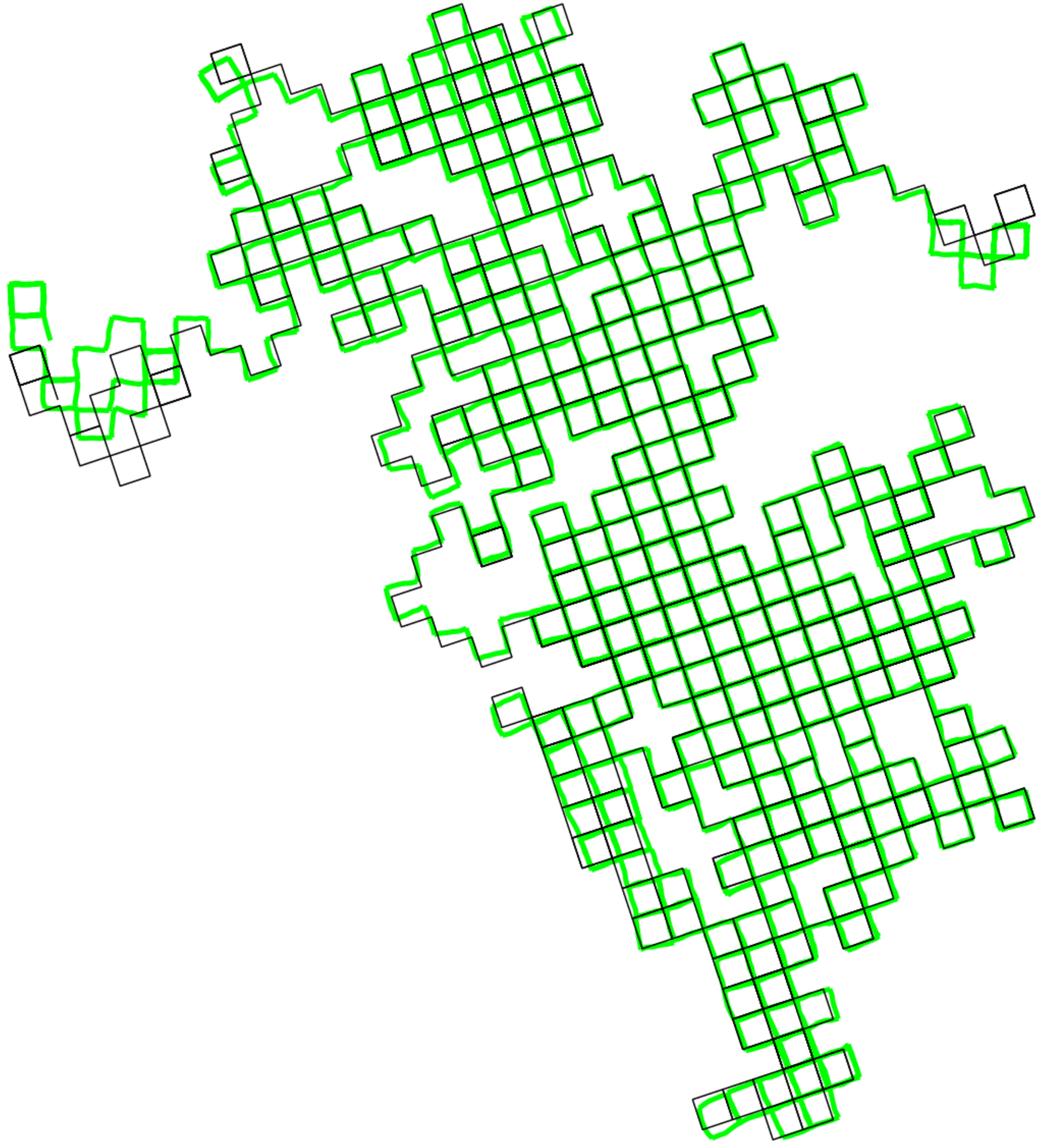}
\end{minipage}\hfill
\begin{minipage}{3.2cm}
	\centering
	\includegraphics[width=2.8cm]{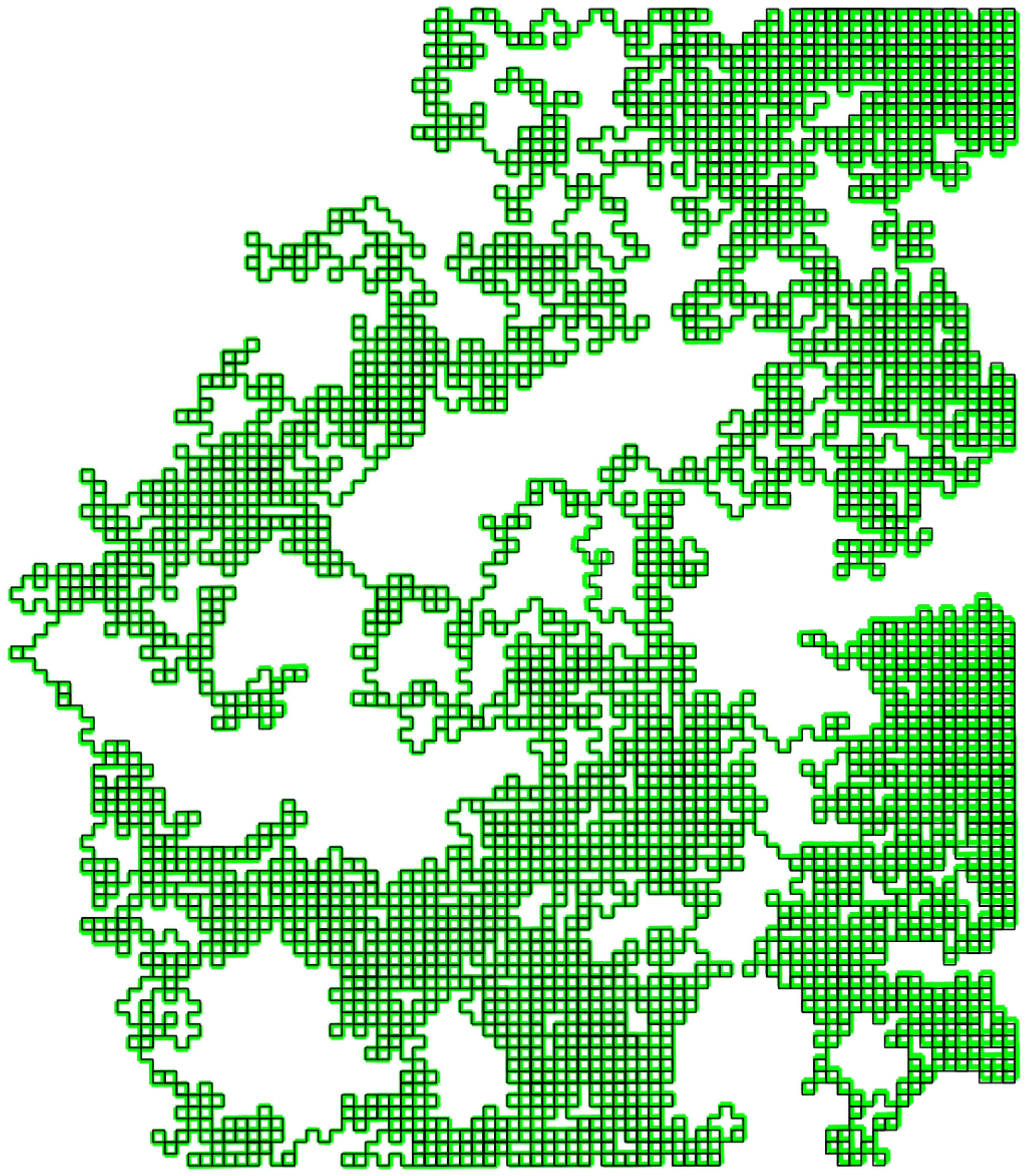} 
\end{minipage}\hfill
\begin{minipage}{3.2cm}
	\centering
	\includegraphics[width=2.8cm]{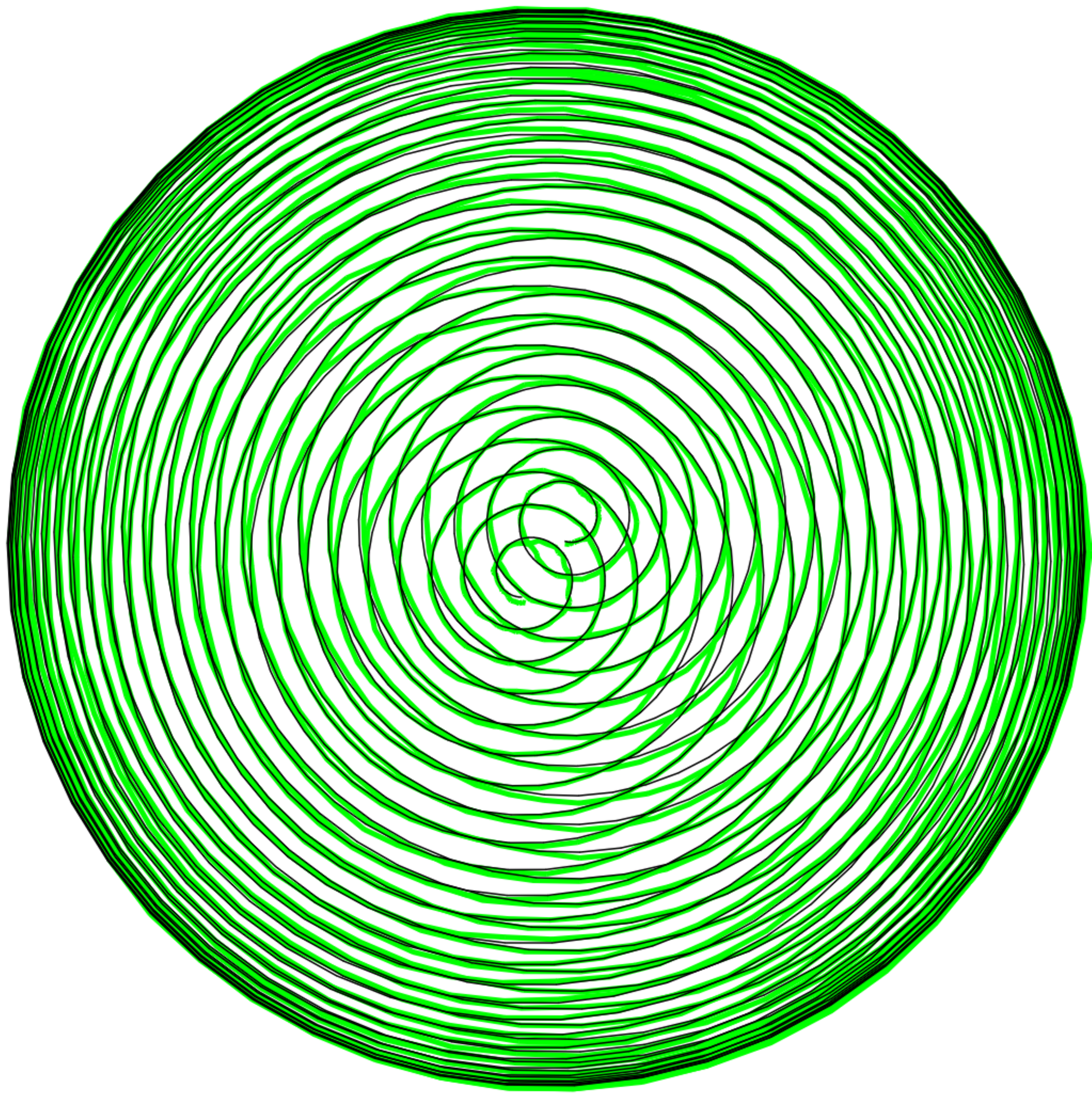}
\end{minipage}\hfill
\begin{minipage}{3.2cm}
	\centering
	\includegraphics[width=2.8cm]{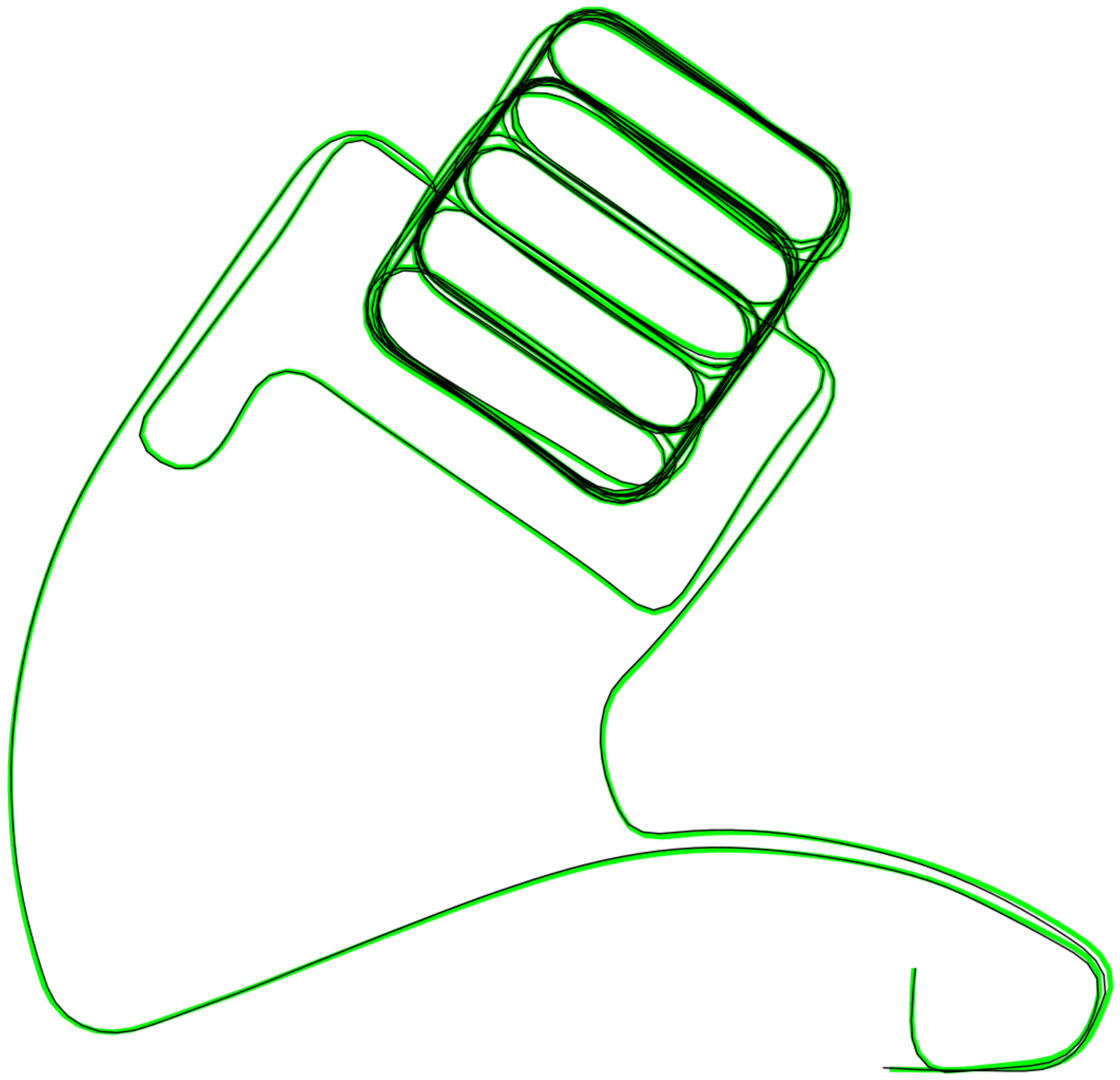}
\end{minipage}\hfill
\\
\hspace{.5cm}
\begin{minipage}{3cm}
	\centering \emph{kitti00}
\end{minipage}\hfill
\begin{minipage}{3cm}
	\centering \emph{10kHog-man}
\end{minipage}\hfill
\begin{minipage}{3cm}
	\centering \emph{100k}
\end{minipage}\hfill
\begin{minipage}{3cm}
	\centering \emph{sphere2500}
\end{minipage}\hfill
\begin{minipage}{3cm}
	\centering \emph{parking-garage}
\end{minipage}\hfill
\end{tabular}
\end{center}
\caption{The datasets. The tiny red crosses mark the robot poses over the blue estimated trajectory in a) full state APAL-SLAM vs. b) compact FPFL-SLAM. c) in green the recovered full state compared with the ground truth in black.}
\label{fig:Datasets}
\end{figure*}

\section{Experimental Validation} \label{Sec:Experiments}
This section evaluates both the state-based loop closure detection and the online state reduction strategy proposed in \autoref{Sec:CompactSLAM}. The incremental SLAM implementation integrates the incremental solving with fluid reordering and resumed Cholesky factorisation proposed in \cite{Polok13rss} and the incremental covariance recovery introduced in \cite{Ila15icra} together with the information-based pose and links selection proposed in this paper. The code and the help bash scripts associated to the experiments can be found in Extensions 2 -- 4 associated to this paper. The included Readme files explain how to perform the compilation and the execution. Extension 2 comprises SLAM++\footnote{https://sf.net/p/slam-plus-plus/} nonlinear least squares library implemented in C++, and also the proposed algorithm. The library performs highly efficient operations on sparse block matrices which have proven to outperform most of the existing state of the art incremental SLAM software \cite{Polok13icra,Ila15icra}.
The tests were performed on a computer with Intel Core $i5$ CPU $661$ running at $3.33$ GHz and $8$ GB of RAM. This is a quad-core CPU without hyperthreading and with full SSE instruction set support.

\subsection{Datasets}\label{SubSec:Datasets}
The tests were performed on several simulated and real datasets which can be seen in \autoref{fig:Datasets} and \autoref{fig:ellipse_plots} and can be found in Extension 5 in a graph file format, which is a popular format in SLAM community. The following subsections describe the datasets and the evaluations in more detail.

\subsubsection{Simulated Datasets}\label{SubSec:SimData}
Several simulated datasets are used to test the methods proposed in this paper. We used four simulated datasets with available ground truth, one generated by our code, called \emph{ellipse3D} and three other datasets which are publicly available, \emph{sphere2500} dataset~\cite{Kaess08tro} 
and \emph{10kHog-man} and \emph{100k}~\cite{Grisetti07rss}.
While the first two, \emph{ellipse3D} and \emph{sphere2500} are 3D datasets, the last two are 2D pose SLAM datasets. 

The \emph{ellipse3D} dataset was generated from a ground truth trajectory in form of two concentric 3D ellipses, the first one with semi-axes of $10$ m and $6$ m and the second with semi-axes of $20\:m$ and $6\:m$, respectively. The length of the total trajectory is of $170$ robot poses over $72.29\:m$. 
The relative transformations between the robot positions are measured with a sensor with $5\%$ error in translation and $5\%$ error in orientation. The sensor is able to establish a link between any two poses closer than $3\:m$ and $0.2\:rad$ in orientation. 

\subsubsection{Real Datasets}\label{SubSec:RealData}
Two real datasets were also used to validate both the state-based loop closure detection as well as the algorithm to maintain a compact representation of the SLAM problem.
We used the \emph{kitti} dataset with available GPS ground truth~\cite{Geiger2013ijrr}. This dataset contains several trajectories from which we selected the longest one, \emph{kitti00} ($4541$ poses). The stereo images are processed by a front-end connected to the SLAM++ nonlinear optimizer. In an on-line processing scenario, the connection is be bidirectional; the front-end provides relative measurements between camera poses to the compact pose SLAM (\autoref{alg:compactSLAM}, line \ref{algline:sensorReg}), and the compact pose SLAM provides the indices of pairs of images that are matching candidates. The stereo processing pipe\-line follows a standard stereo processing algorithm, including feature detection and matching, triangulation and 3D point cloud alignment followed by least square reprojection error refinement.

However, the \emph{kitti00} dataset is used to validate the state-based loop closure strategy proposed in this paper which is compared with the appearance based methods. The tests analyze the threshold sensitivity of both methods, therefore repeated runs with different thresholds need to be performed. For that, we generated a file containing all possible loop closures, by exhaustively matching all-to-all images, and computing the relative transformations within a RANSAC approach. Only the transformations with the inlier ratio greater than $0.35$ are kept (using a higher threshold leads to having only highly relevant loops, which would make the work of the loop-closing algorithm a simple one). In this way all the tests in \autoref{SubSec:LoopClose} were done on the same dataset.

The uncertainty of each relative pose measurement in the \emph{kitti00} dataset is estimated using Monte Carlo approach: each feature point used in the calculation of the relative transformations is corrupted using zero mean Gaussian error with variance equal to $1\:px$ and samples are drawn from that distribution. Each sample is propagated through the triangulation and relative camera pose estimation to find the measurement covariance. The images can be used to detect loop closures based on appearance, therefore this dataset is used to compare the loop closure strategy based on state estimation to an appearance-based one.

A second 3D dataset called \emph{parking-garage} was also used in our tests~\cite{Kuemmerle11icra}. This dataset is a 3D pose graph of a multi-level parking garage and has about $1661$ robot poses and 6275 edges. Being a graph file dataset, this dataset assumes loops are already detected, therefore it will be used only in testing the compact SLAM representation. Since this dataset does not provide ground truth, it was processed by one hundred iterations of a batch solver and the results were used as a de-facto ground truth.

\subsection{Compact Pose SLAM Evaluation}\label{SubSec:CompactSLAMeval}
Maintaining a compact state representation in online SLAM can lead to great computational savings. Three strategies were tested: a) all the poses and all the possible loop closures are integrated into the system (denoted as all poses, all loops (APAL)), b) all the poses but only informative loop closure links are integrated into the system (all poses, few loops (APFL)) and c) relevant poses and informative loops are integrated into the system (few poses, few loops (FPFL)). The selection of one or another strategy can be easily implemented by appropriately setting the distance and information thresholds ($v$, $s$, $g_{pose}$ and $g_{loop}$) in \autoref{alg:compactSLAM}. In particular, setting $s$ to zero and both $g_{pose}$ and $g_{loop}$ negative infinity, respectively, leads to APAL strategy with all the loop closures detected by the distance test. Setting up a higher value for the probability threshold $s$ reduces the number of loop closure candidates.
Setting $g_{loop}$ to a minimum mutual information a loop needs to have to be added to the system leads to APFL strategy, and setting $g_{pose}$ to a minimum of information a link connecting a pose must have, leads to FPFL. Again, we offer an automated solution to select the adequate thresholds for the strategy to use. The three strategies were tested on above mentioned datasets. Execution time and translational and rotational errors are provided in Tables \ref{tab:Time} and \ref{tab:ErrorEval}, respectively.

To select suitable values of thresholds $v$, $s$ and $g_{pose}$ and $g_{loop}$,
a representative part of each dataset is used for setting the thresholds. The value of $60\%$ (by the number of vertices) was used with all the datasets to make sure that a major loop closure is included in this sample, except for the \emph{100k} dataset which is highly repetitive and only $20\%$ sample was used.
On this sample, poses and measurements are incrementally added into the system and, at each step, $\sigma^2_d$ and $\mu_d$ are recorded for each loop closure with the current pose. At the end, it is possible to find such a sensor range $v$ so that all loops would have probability above a certain threshold $s$, using \eqref{eq:similarity}. The choice of $s$ is arbitrary and we chose the value of $0.1$ to stay in the region where the number of proposed candidates is stable, with some space for increasing this value if required. Conversely, starting with a known sensor range (e.g. from the physical characteristics of the given sensor and the capabilities of the corresponding sensor registration algorithm), it is possible to find such value of $s$ that no loop closures are lost.

For the APAL scenario, $g_{pose}$ and $g_{loop}$ are simply set to negative infinity in order to accept all the poses and loops as having sufficiently high mutual information. A run with this configuration is then performed while recording the mutual information of all the ground truth loop closures edges as well as the mutual information of the edges linking every new pose (recorded in \autoref{alg:compactSLAM}, lines \ref{algline:loopGainTest} and \ref{algline:toBeOrNotToBe}).

To generate the APFL configuration, $g_{loop}$ is then set to \mbox{$e^{(1.36 ln(l_{90}+1))}-1$ where $l_{90}$} is a $90$-percentile of all the recorded mutual informations of the loop edges.
It is possible to run with this configuration and verify that no important loops are lost and at the same time enough loops are being discarded. If that is not the case, it is possible to manually adjust the threshold before selecting the $g_{pose}$ threshold. To this end, the Extension 4 contains a simple script which takes the initial $g_{pose}$ as an input and runs several tests using the multiples of this value, distributed in the range $[g_{pose}\frac{1}{10}, 10 g_{pose}]$ in such a way that the ratio of the adjacent thresholds is a constant. Then, it is just a matter of choosing the preferred trade-of between speed and precision and using the corresponding threshold.

To obtain FPFL we fix $g_{loop}$ at the chosen value and set $g_{pose}$ to \mbox{$e^{(1.7 ln(p_{90}+1))}-1$} where $p_{90}$ is $90$-percentile of all the recorded pose mutual information. Another run with this configuration is performed to make sure that the pose graph sparsity is as expected. Again, if the result is not satisfactory, it is possible to execute several runs with scaled values of $g_{pose}$ and to choose a suitable value for the threshold. Note that setting the thresholds happens only on the \emph{sample} rather than on the entire dataset. Also note that in most cases the thresholds proposed by the above-mentioned heuristic do not require further fine-tuning.

The above-mentioned heuristics were developed by first running exhaustive tests on all the datasets, then manually choosing the preferred thresholds and finally finding a function which would yield values close to the manually chosen ones. The thresholds used in our tests are listed in \autoref{tab:Time}. Out of all the thresholds, only two had to be manually modified. Specifically, in \emph{parking-garage} the $g_{loop}$ was reduced in order to allow more loop closures and decrease the error, and in \emph{sphere2500}, $g_{pose}$ was increased in order to obtain a more compact representation. The suggested thresholds would still work in both cases, except that in the former case of \emph{parking-garage} the solution would be less precise (accepted only $115$ out of $4615$ loops rather than $964$ with the decreased threshold) and in the latter case of \emph{sphere2500}, it would be less compact (accepted $2500$ poses rather than only $959$ with the increased threshold).


\subsection{Error Evaluation}\label{SubSec:Error}
In order to evaluate the compact pose SLAM algorithm, we want to compare the rotational and translational errors of the final estimate for all the three cases mentioned above. The problem when evaluating the accuracy is the fact that the size of the state varies in all cases. Several types of errors are proposed in the literature for evaluating the SLAM problem. Relative pose error (RPE) was used in \cite{Kummerle09ar} and \cite{Sturm12iros} and was shown to be useful in the evaluation of the graph based SLAM. A more intuitive way is to compare the absolute trajectory error (ATE) after registering the two configurations: the ground truth and the estimated graph \cite{Sturm12iros}.
For calculating the relative pose error or absolute trajectory error, we first need to find a way to recover the poses corresponding to the poses in the initial graph which are missing in the compacted representation.

One way of doing this is by applying linear interpolation to each edge in the compact representation, while using the contribution of the corresponding edges in the initial representation as weights. The cumulative weights can be calculated given the measurements $z_{\cdot,\cdot}$ as follows: 
\begin{equation}
w_u = \frac{\sum_{k=i}^{i+u- 1}{\left\Vert z_{k,k+1} \right\Vert}_2} {\sum_{l=i}^{i+q-1}{\left\Vert z_{l,l+1} \right\Vert}_2}\;, 
\end{equation} 
where $q$ is the the length of the path in the graph between variables $^{I}\theta_{i}$ and $^{I}\theta_{i+q}$ in the initial ($I$) state, where $^{I}\theta_{i}$ corresponds to $^{C}\theta_{j}$ in the compacted ($C$) graph and similarly $^{I}\theta_{i+q}$ corresponds to $^{C}\theta_{j+1}$, and $0 \leq m \leq q$ is index of a pose in this path.
The poses $^{I}\theta_{i}$ through $^{I}\theta_{i+q}$ can now be interpolated as:
\begin{equation}
^{I}\theta_{i+u} = ^{C}\theta_{j} \oplus w_m \cdot (^{C}\theta_{j+1} \ominus ^{C}\theta_{j})\;. \label{eq:v1_reconst}
\end{equation}

Another way is to apply the weighting in the error space. For that we calculate the relative displacement (error) $d_{j, j+1}$ between the initial and optimised estimation of each odometric edge in the compacted graph:
\begin{equation}
d_{j,j+1} = h( ^{C}\theta_{j},^{C}\theta_{j+1}) \ominus (z_{i,i+1} \oplus \ldots \oplus z_{i+q-1,i+q})\;,
\end{equation}
and the poses $^{I}\theta_{i}$ through $^{I}\theta_{i+q}$ can be approximated as:
\begin{equation}
^{I}\theta_{i+m} = ^{C}\theta_{j} \oplus (z_{i,i+1} \oplus \ldots \oplus z_{i+m-1,i+m}) \oplus w_m \cdot d_{j,j+1}\;. \label{eq:v2_reconst}
\end{equation}

\begin{figure}[t]
\begin{center}
\includegraphics[width=0.8\linewidth, height=0.65\linewidth]{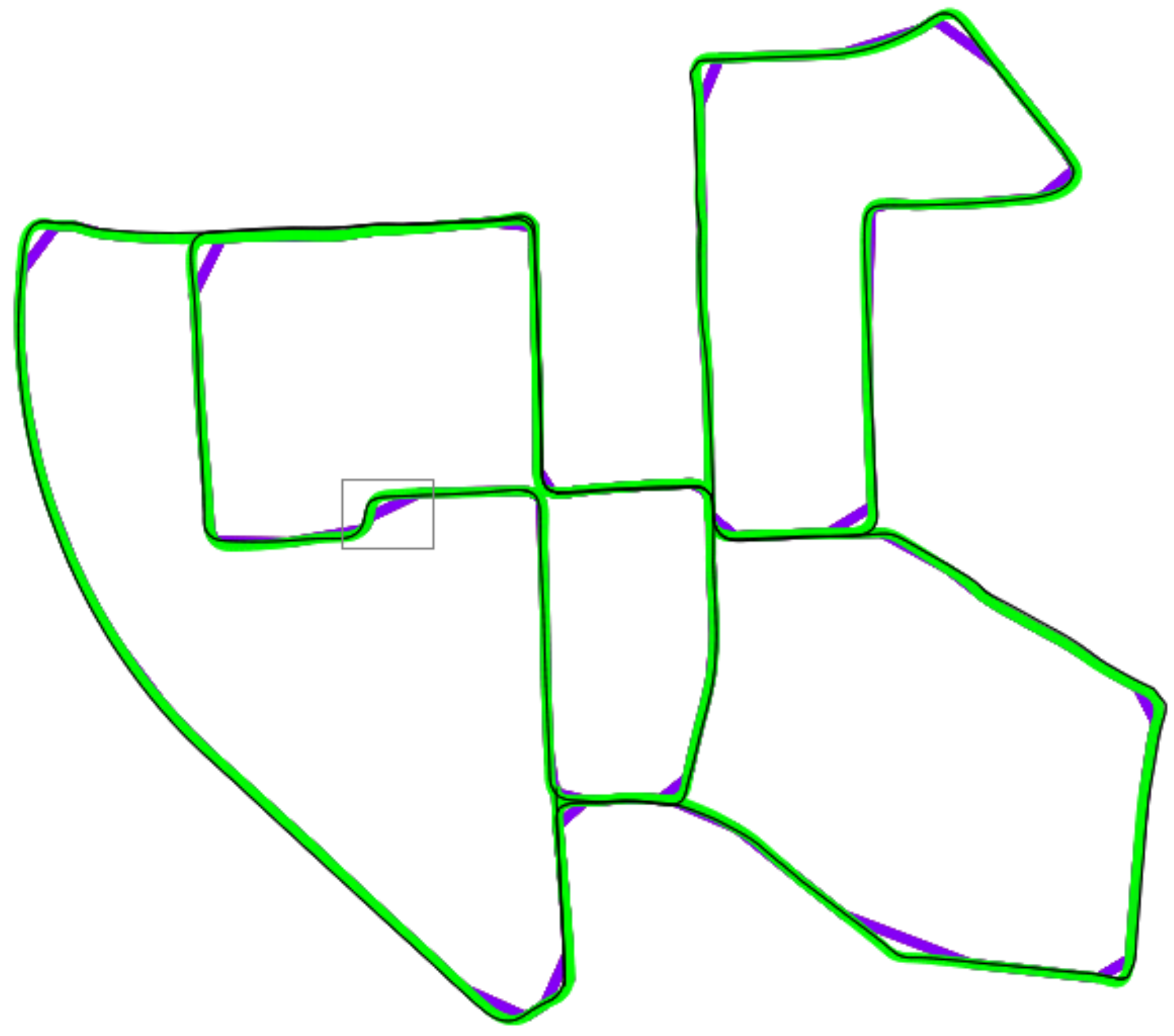} \\ 
\begingroup\vspace{2mm}
\setlength{\fboxsep}{0mm}
\fbox{\includegraphics[trim=0mm 10mm 0mm 0mm, clip, width=0.75\linewidth]{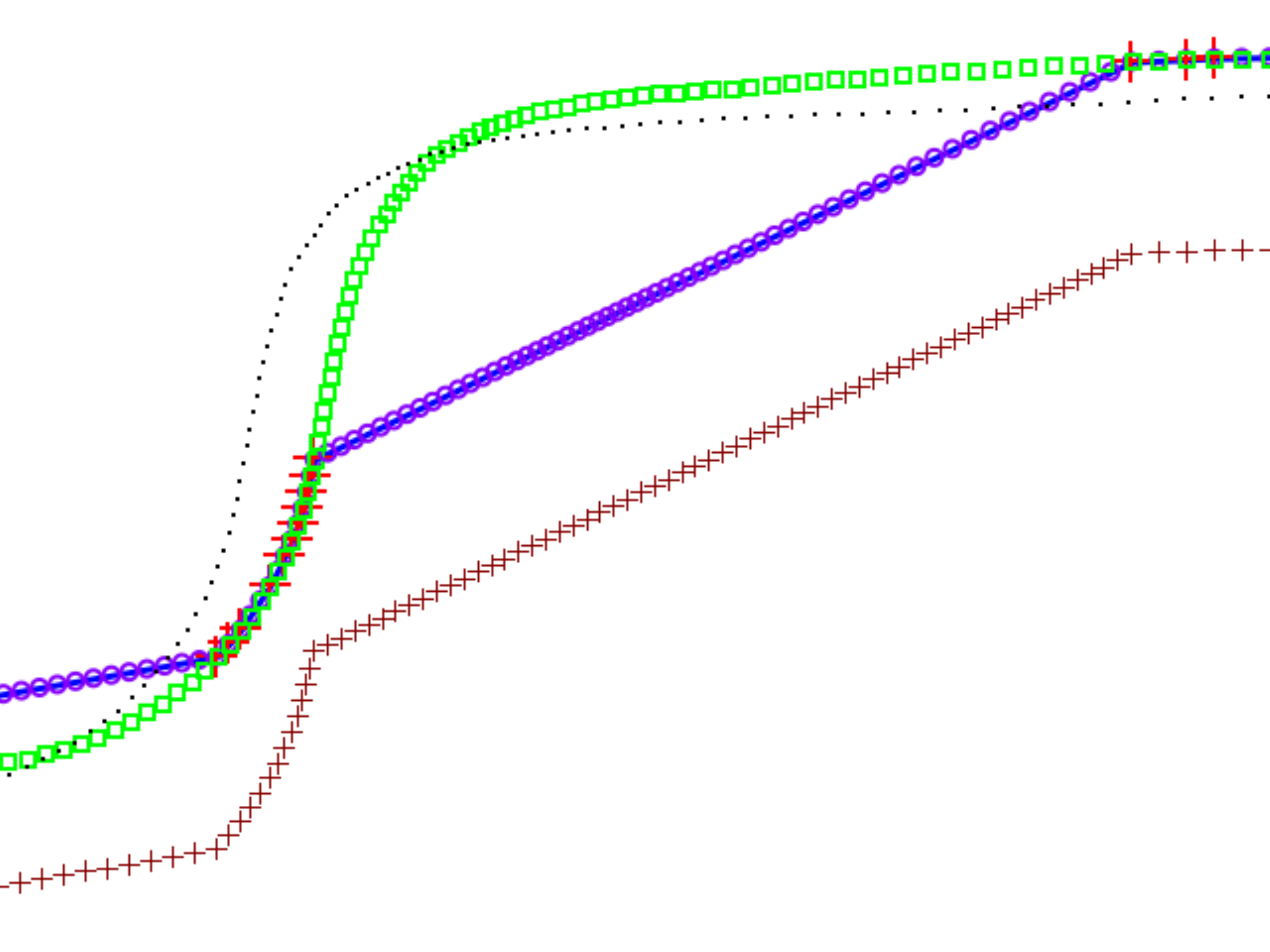}} \\
\endgroup\vspace{-2mm}
\end{center}
\caption{\emph{Kitti00} trajectory compacted to only 355 poses out of 4541. v0 and v1 in violet, v2 in green, ground truth in black (top), a detailed view of the trajectory, optimized poses marked by large red crosses, v0 small dark red crosses (shifted vertically to not overlap with v1), v1 violet circles, v2 green rectangles, ground truth in black (bottom).}
\label{fig:Interpol}
\end{figure}

Note that in both \eqref{eq:v1_reconst} and \eqref{eq:v2_reconst}, $^{I}\theta_{i} = ^{C}\theta_{j}$ and $^{I}\theta_{i+q} = ^{C}\theta_{j+1}$ holds. \autoref{fig:Interpol} shows that the interpolated trajectory using \eqref{eq:v2_reconst} represented by the green squares nicely follows the ground truth represented by the black dots. In the captions, this strategy corresponds to v3. On the other hand, the violet circles representing the interpolated trajectory using \eqref{eq:v1_reconst} and denoted v1, stay on the segments defined by the compact representation (big red crosses). \autoref{fig:Interpol} also shows interpolation when the weights in \eqref{eq:v1_reconst} are uniform in small dark red crosses, denoted v0. Note how the violet circles concentrate near the curve, while the red crosses do not.

\setcounter{topnumber}{1}

\begin{table}[t]
\begin{center}
\begin{tabular}{|c|c|c|c|}\cline{2-4}
\multicolumn{1}{c|}{} & \multicolumn{3}{c|}{RMSE Error} \\ \hline
alg.	& ATE & RPE & RPE all-all \\\hline
v0	& $3.853 m, 3.392 \textdegree$	& $0.076 m, 0.459 \textdegree$	& $11.782 m, 3.374 \textdegree$	\\
v1	& $3.818 m, 3.598 \textdegree$	& $0.064 m, 0.501 \textdegree$	& $12.363 m, 3.581 \textdegree$	\\
v2	& \bm{$3.093$}$m$, \bm{$2.250$}$\textdegree$ & \bm{$0.029$}$m$, \bm{$0.119$}$\textdegree$ & \bm{$7.297$}$m$, \bm{$2.223$}$\textdegree$	\\ \hline 
\end{tabular}
\end{center}
\caption{Error evaluation for the \emph{kitti00} dataset. }
\label{tab:Interpol}
\end{table}

\setcounter{topnumber}{\mttopnumber}

Now that we have recovered the full state, we can rigidly align it to the ground truth trajectory e.g. using the Kabsch algorithm \cite{Kabsch76ac} and apply the error metrics as described in (2) and (5) in \cite{Sturm12iros}. These are the absolute trajectory error (ATE) -- the error between the corresponding poses in the estimated and the ground truth trajectory, the relative pose error (RPE) -- an error between the corresponding relative transformations between the consecutive poses in the estimated and the ground truth trajectory and finally the relative pose error all to all (RPE all-all) -- an error between the corresponding relative transformations between all the possible pairs of poses in the estimated and the ground truth trajectory. The translational and rotational components of these errors are reported separately in \autoref{tab:ErrorEval}. The results for the trajectory in \autoref{fig:Interpol} can be found in \autoref{tab:Interpol}. The v2 strategy always leads to the lowest error, therefore it will be further used for the rest of the evaluations.


\subsection{State-based Loop Closure Detection}\label{SubSec:LoopClose}
The \autoref{alg:compactSLAM} integrates a loop closure search scheme based on the estimated state at each step. This is done at line \ref{algline:searchLoopClosures} and it is based on the distance test described in \autoref{SubSec:Distance}.

Alternatively, in case that the robot is equipped with an image sensor, appearance can be used to detect loop closures. \emph{FAB-MAP2} is an appearance based method which classifies the place the robot is currently seeing as new or already seen before from a different pose \cite{Cummins10ijrr}. If a current place is categorised as seen before, the online SLAM algorithm attempts to close the loop by matching the similar views. \emph{FAB-MAP2} uses visual words to represent the appearance, which were obtained a priori from a training set. \emph{FAB-MAP2} is a vision-baded technique suitable for closing very large loops under good lighting condition in non-repetitive environments. On the other hand, the proposed state-based loop closure detection algorithm works for any exteroceptive sensors (lasers, sonars, etc) which can be registered to obtain relative transformations and is independent of the environment, although it requires relatively small loops and good estimation in order to be highly efficient. The probability thresholds need to be set in concordance with the size of the loops and the errors in the estimation.

In the case of the \emph{distance test}, one needs to provide a trusted sensor range threshold $v_r$ for each measurement dimension and a probability threshold $s \in [0,1]$. If $s$ is set too low, more loop closure candidate are generated by the test, and this is not desired. If the threshold is set too high, some loop closures might be lost. Our tests show that, once the sensor range threshold $v$ is set correctly, $s$ drastically affects the number of candidates only in the very close vicinity of $0$ and $1$, being relatively conservative, otherwise. This characteristic favours the automatic selection of the threshold, and therefore allowed us to actually implement it in our code. This has a great benefit in real robotic applications, a robot can sample a small part of the environment and based on that, automatically decide on which thresholds to use for the rest of the long-run mission. To our best knowledge, this is the only existing loop closure detection strategy that allows a high level of automation of the process.

\begin{figure}[t!]
\begin{center}
\begin{tabular}{c}
\includegraphics[trim=40mm 28mm 40mm 20mm, clip, width=0.9\linewidth]{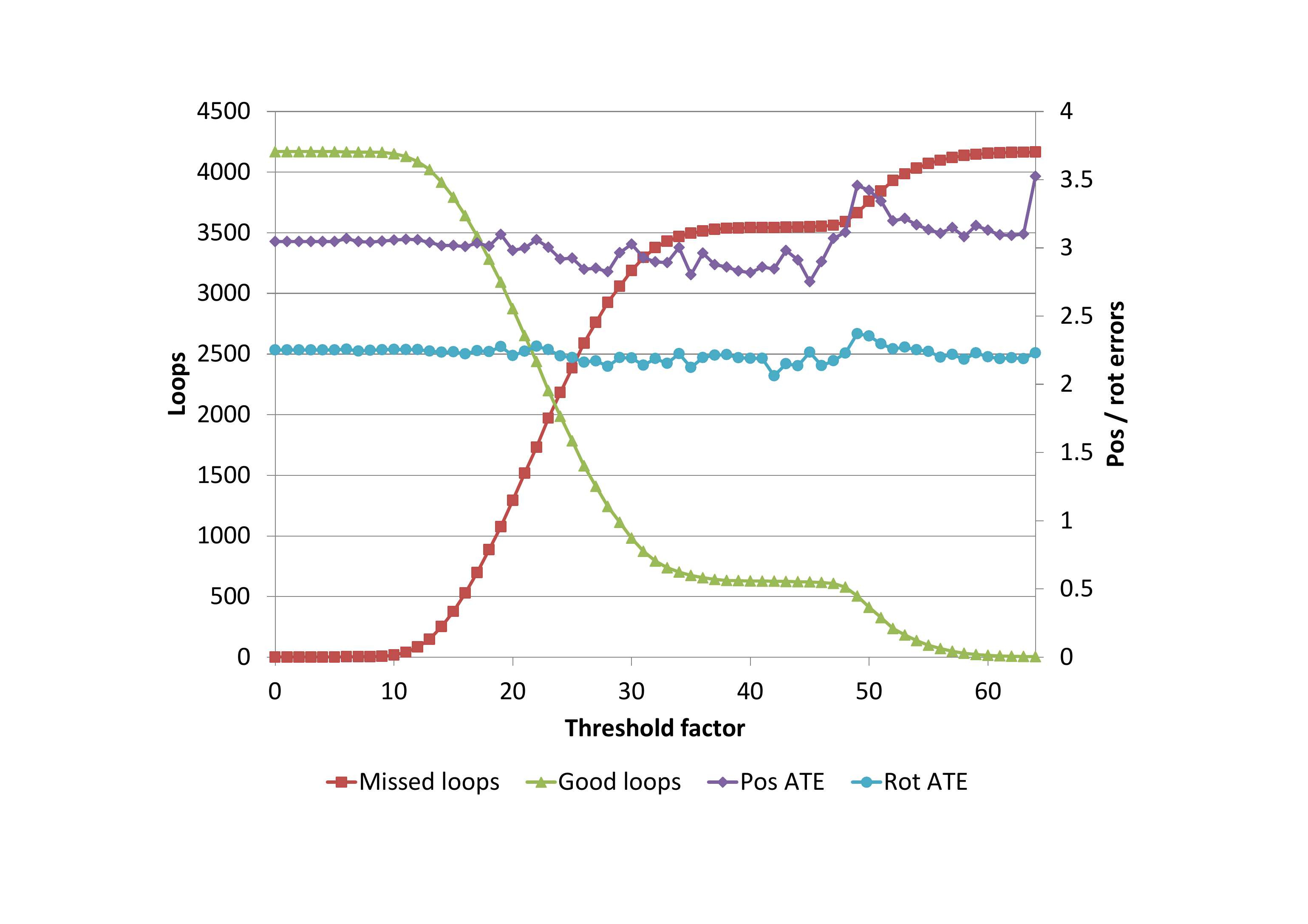} \\
\includegraphics[trim=40mm 28mm 40mm 20mm, clip, width=0.9\linewidth]{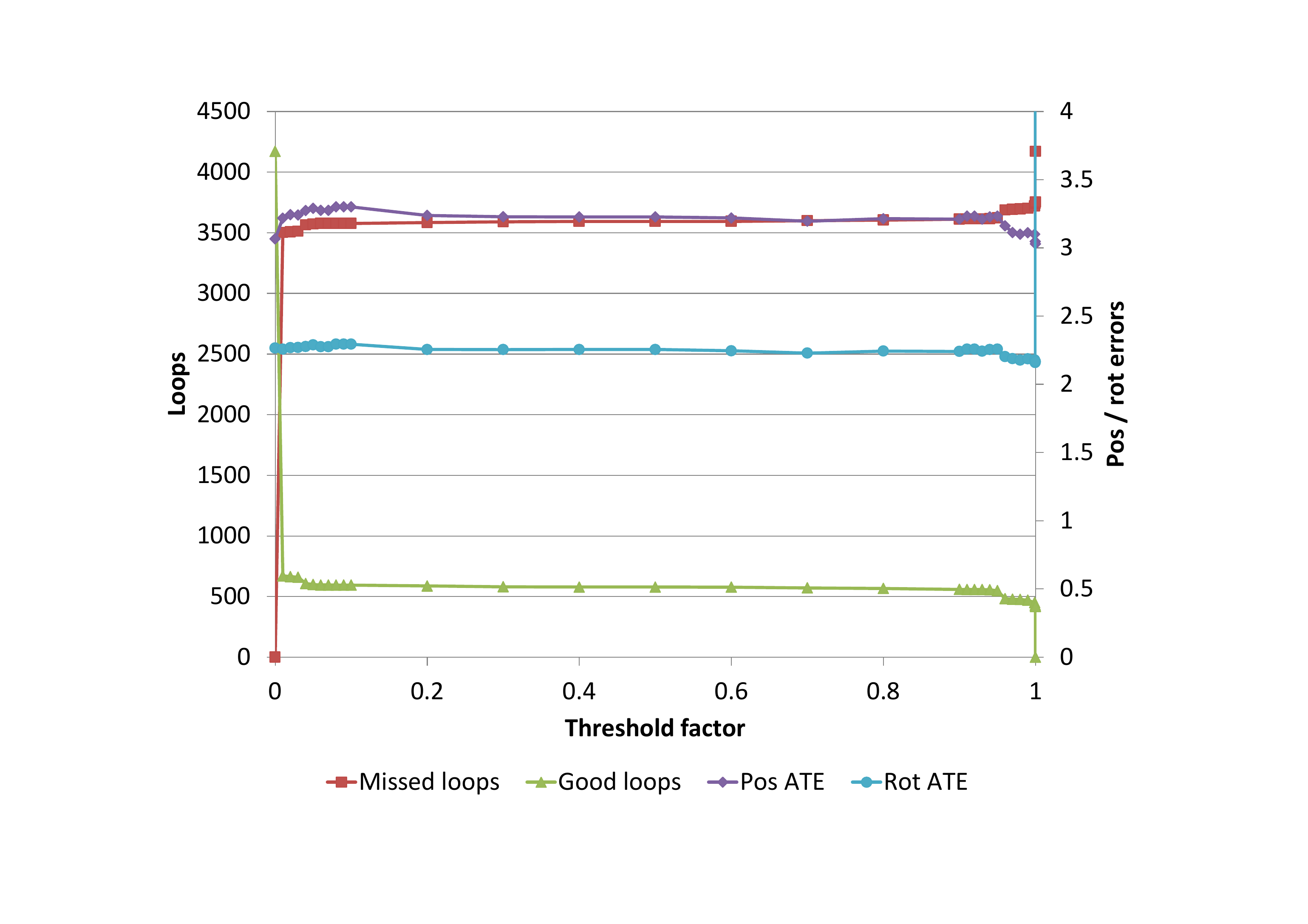} \\
\includegraphics[trim=40mm 28mm 40mm 20mm, clip, width=0.9\linewidth]{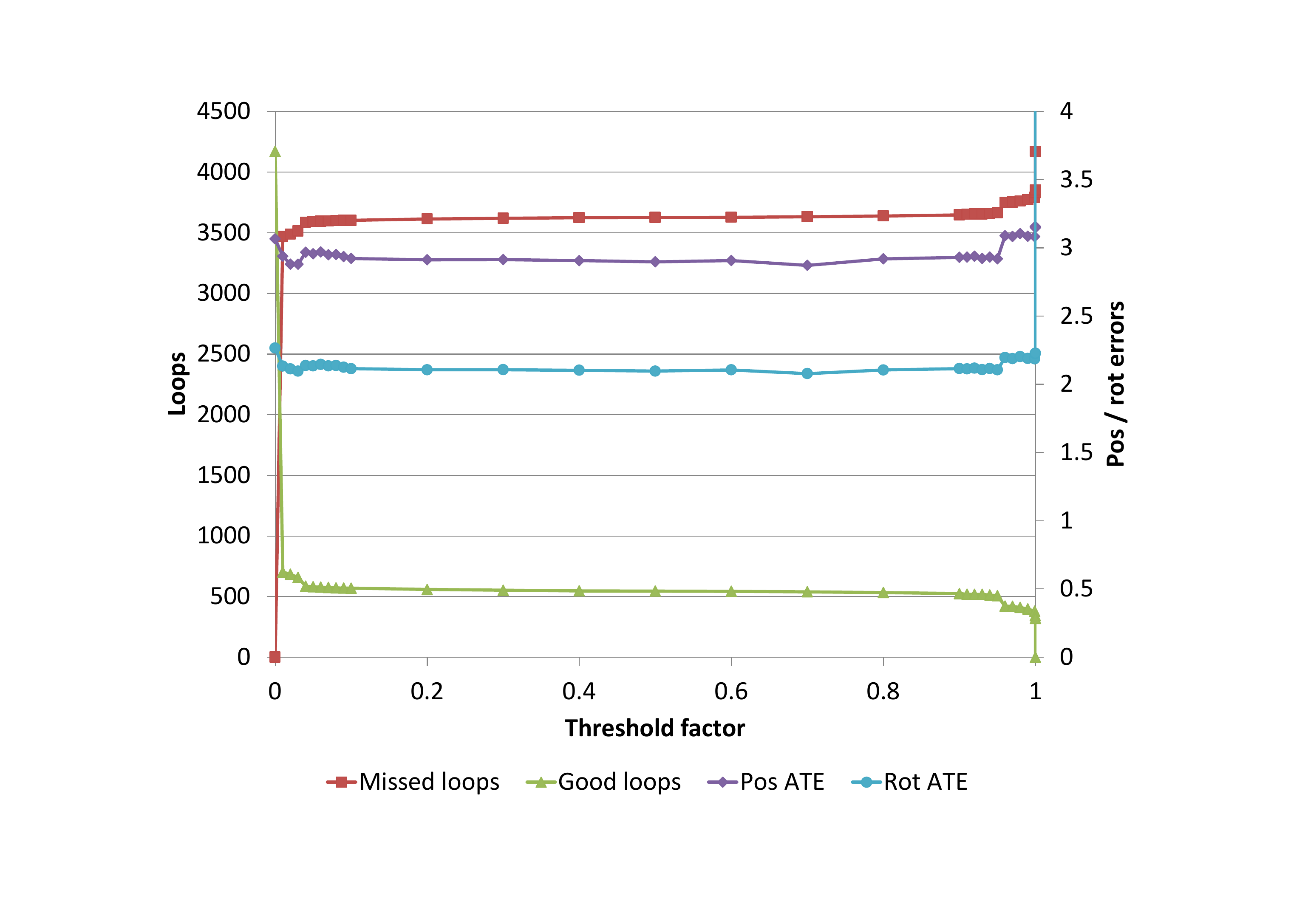}
\end{tabular}
\end{center}
\caption{Comparison of sensitivity of different loop closing methods on their respective thresholds, distance-based loop closing (top), \emph{FAB-MAP2} on the Oxford dictionary provided by the authors (middle) and \emph{FAB-MAP2} on the dictionary we trained on the \emph{kitti} dataset (bottom).} 
\label{fig:fabmapThrAnal}
\end{figure}
We further evaluated the dependence of the two loop closing methods on their respective thresholds and the effect it has on the precision. We also evaluated the number of detected valid loops. In order to do that, we processed each pair of images by standard sparse relative pose estimation procedure and applied a threshold of $0.35$ on the matched keypoint inlier ratio. \autoref{fig:fabmapThrAnal} (top), shows the dependence of the number of detected loops and the corresponding solution error on the loop closure information gain threshold. Note that the horizontal axis is a factor of loop gain, with value $32$ being equal to $l_{90}$ (the $90\%$ percentile of loop gains in the sample of the dataset) and the loop gain changes $100$-fold down to value $0$ or up to value $64$, respectively, with the ratio between the consecutive gains being constant. This is because the information gain is a logarithmic quantity. It would also be possible to use a logarithmic plot, but then the vertical axis would be in the middle. Note that the number of detected loops changes smoothly, in two intervals: in $[10, 35]$ the relatively short loops with low information gain are being culled, whereas in $[48, 60]$ the long loops with high information gain are being culled. The threshold we applied in our evaluations falls in the plateau in between those two intervals. The error varies slightly and in value $65$ it would spike up since that is the point where all the loop closures are culled. Also note that the number of missed loops is zero on the left and it reaches the number of the loops on the right.

On \autoref{fig:fabmapThrAnal} (middle) there is a plot of the number of \emph{FAB-MAP2} candidates and the solution error, depending on the probability threshold, using the Oxford dictionary. Only the left frames (out of the stereo pairs in the \emph{kitti} dataset) were used for all the \emph{FAB-MAP2} evaluations. Note that the number of loops is relatively constant and changes abruptly in the $[0.0, 0.1]$ interval and, more importantly, also in the $[0.9, 1.0]$ interval which is the typical working point. Note that \emph{FAB-MAP2} misses many of the loops with good inlier ratio, even though the threshold applied is very low (it only accepts all the loops if zero threshold is specified). Also note that the error is worse than that of the distance-based approach.

Since the results on the Oxford dictionary were worse than ours, we decided to train a new dictionary specifically for the \emph{kitti} dataset. We tried to match the procedure described in \cite{Cummins10ijrr}. We used $2.5$~millions of SURF features extracted from the frames of left camera sequences \emph{kitti01} -- \emph{kitti21}, deliberately skipping \emph{kitti00}. We used frames spaced approximately every $20$m (only $1951$ out of all the frames). K-means clustering was trained on Intel Core i5-4590 with $8$~GB RAM while Chow-Liu tree was computed on dual Intel Xeon E5-2665 ($16$~cores in total) with $64$~GB RAM because of its high memory requirements. The calculation of the K-means took 1~day 06:00:30 and required $2.5$~GB of memory, The subsequent calculation of Chow-Liu tree took 12:42:01 and required $25$~GB or memory (using the compact method -- the fast one would require $45$~GB of RAM).

On \autoref{fig:fabmapThrAnal} (bottom) there is a plot of the number of \emph{FAB-MAP2} candidates and the solution error, depending on the probability threshold, using the \emph{kitti} dictionary. Note that the number of loops is still relatively constant and again changes abruptly at the borders of the plot. Note that using a custom dictionary did not help closing all the loops with good inlier ratio, even with low thresholds. The error improved somewhat, compared to the Oxford dictionary.


\subsection{Conservativeness of the Compact Pose Estimate}\label{SubSec:CovNorms}
An important property of the compact SLAM algorithm is conservativeness of the computed pose estimates. If the algorithm produces an over-confident estimate, the robot poses could be imprecise by more than what their covariance suggests, which could lead to bad decisions in data association, loop closure detection, motion planning, etc.. If the estimate is, on the other hand, over-conservative, the result would be equally difficult to use.

To evaluate the conservativeness of the estimate, the norms of marginal covariances of all the variables in the system are calculated, at each step. These indicate the uncertainty in the poses of the trajectory. Additionally, the norms of full covariances (i.e. both marginal and cross-covariances) are calculated at each step. These add more information about the correlation of the variables in the system. These two norms are calculated for three scenarios: the compact SLAM algorithm (FPFL), the SLAM algorithm including all the possible poses and loops (APAL) and also a variant of APAL where the redundant poses not present in the compact representation are marginalised out using the Schur Complement. This essentially compares the effects of \emph{measurement composition} in the case of compact SLAM with the effects of \emph{variable marginalisation}. 

\begin{figure}[t]
\begin{center}
\includegraphics[trim=40mm 28mm 40mm 20mm, clip, width=1.0\linewidth]{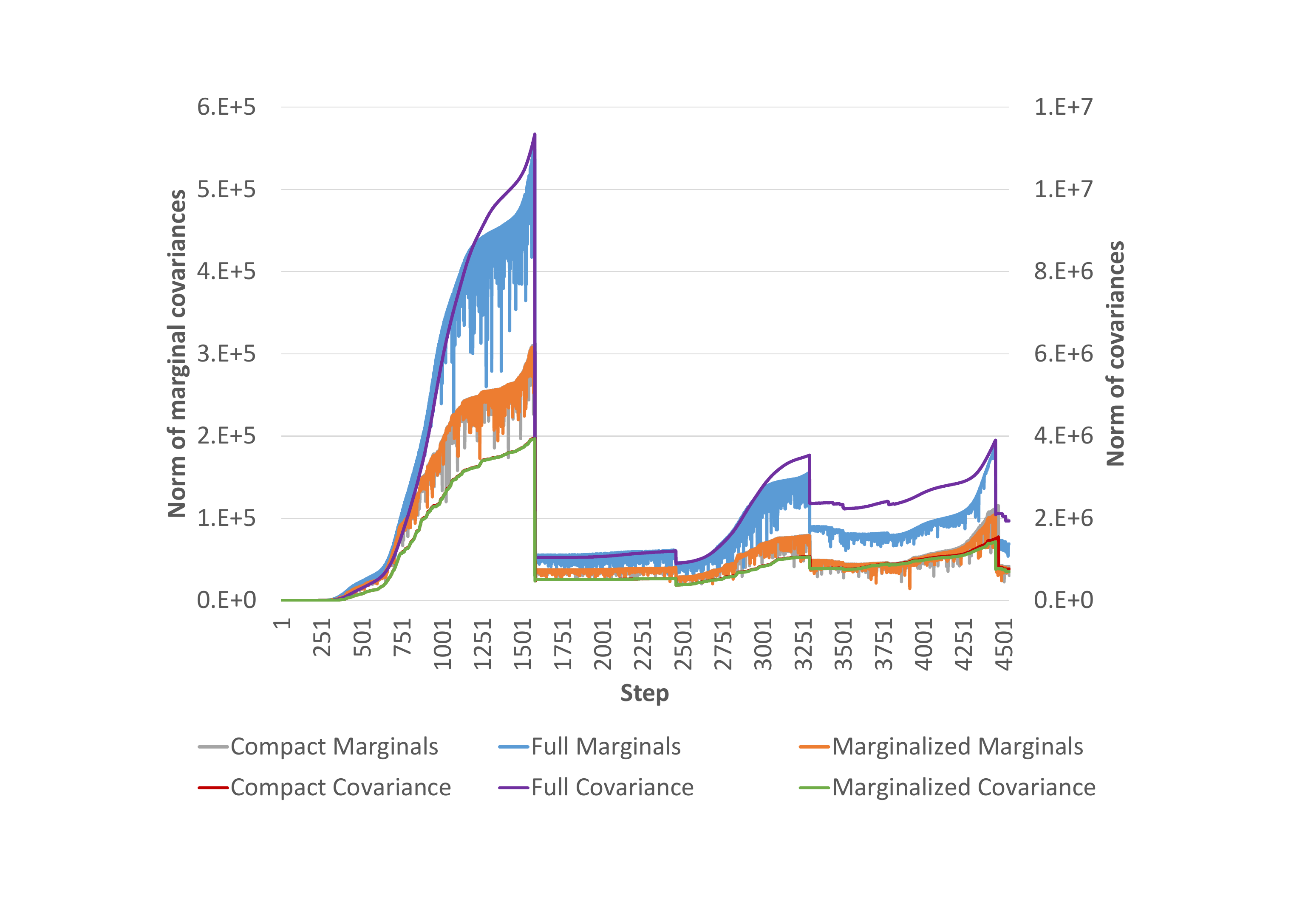} \\
\includegraphics[trim=40mm 28mm 40mm 20mm, clip, width=1.0\linewidth]{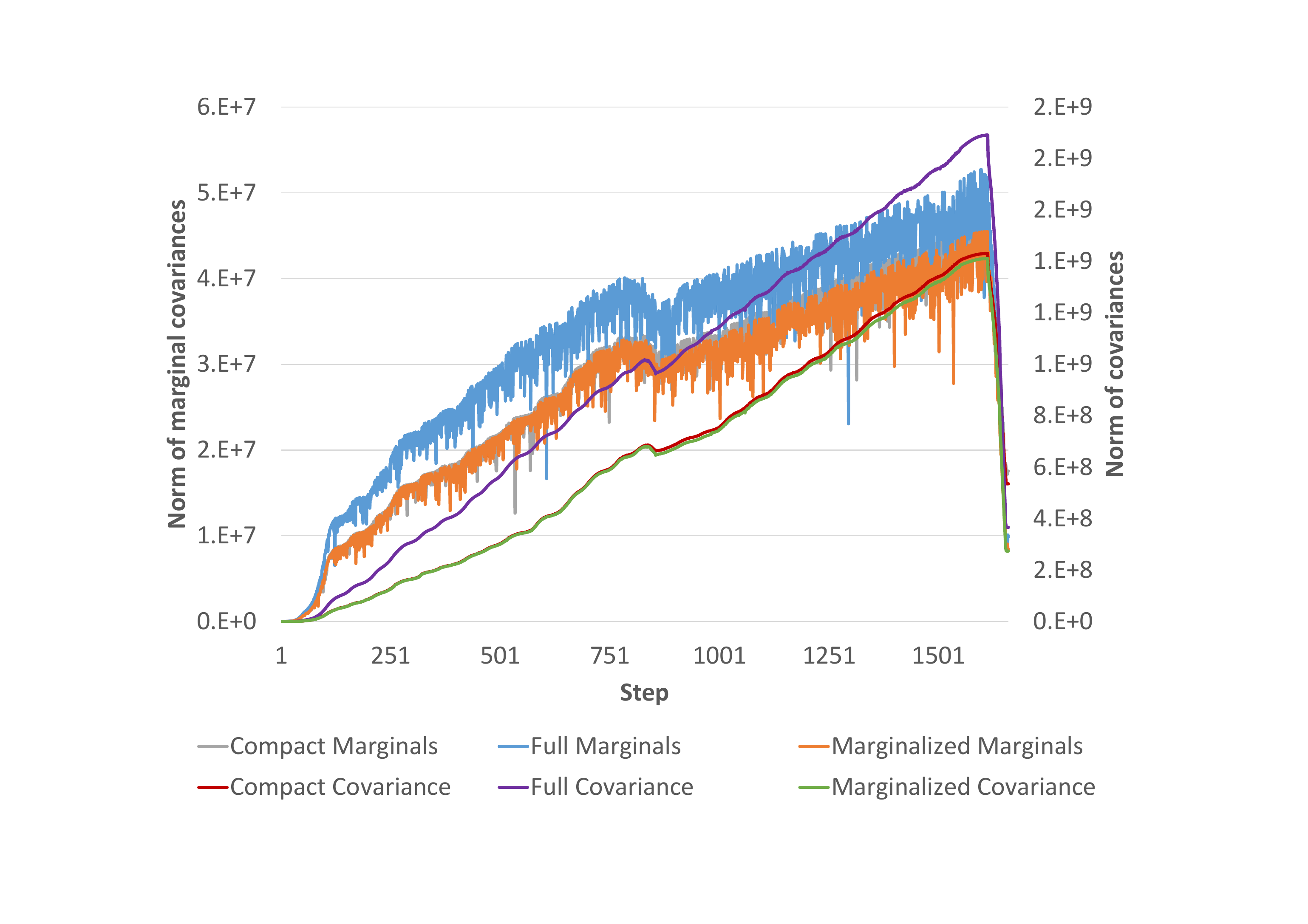}
\end{center}
\caption{Comparison of the norms of covariances of the incremental solutions of the \emph{kitti00} (top) and \emph{parking-garage} (bottom) datasets. Best viewed in color.}
\label{fig:conservative}
\end{figure}

\autoref{fig:conservative} plots the evolution of the covariance norms for the \emph{kitti00} and \emph{parking-garage} datasets. The highest norm corresponds to the APAL SLAM. This is followed by the FPFL and the marginalised system. This indicates that the compact SLAM is slightly more conservative than marginalisation of the variables not present in the compact system. A similar result is obtained when comparing the norms of marginal covariances -- again, the full system has the greatest norm of marginal covariances and the compact and marginalised systems have approximately the same norms. The slight difference in norm stems from the fact that the covariances of the composed measurements are calculated using an approximate function as described in \cite{Smith86ijrr}. The same evaluation was performed on the other datasets as well, supporting the same conclusions, but were omitted from this paper to save space.


\begin{figure}[t]
\begin{center} \hspace{-.25cm}
\begin{tabular}{c}
\begin{minipage}{.5cm}
	\centering
	a)
\end{minipage}
\begin{minipage}{0.42\linewidth}
	\centering
	\includegraphics[width=1\linewidth]{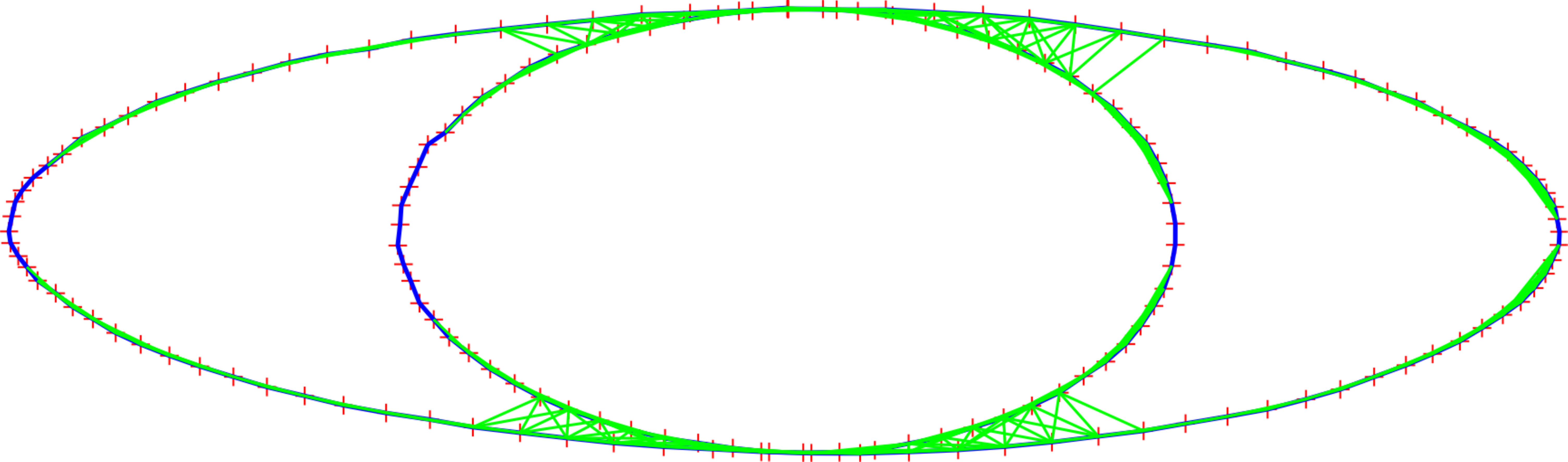}
\end{minipage} \hspace{.25cm}
\begin{minipage}{0.42\linewidth}
	\centering
	\includegraphics[width=1\linewidth]{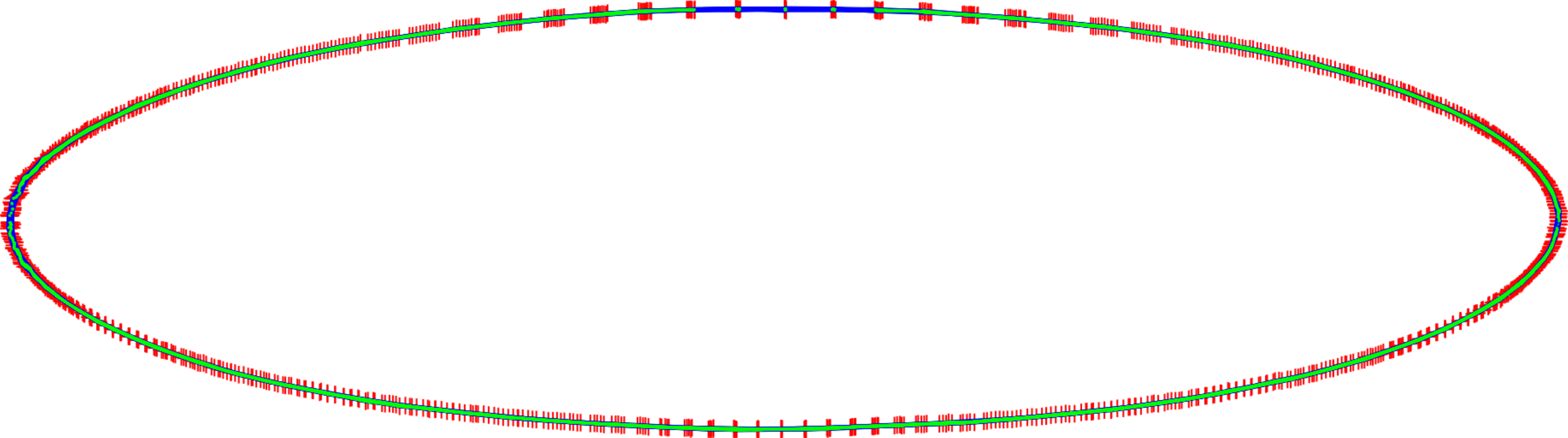}
\end{minipage} \hfill \vspace{.2cm} \\
\begin{minipage}{.5cm}
	\centering
	b)
\end{minipage}
\begin{minipage}{0.42\linewidth}
	\centering
	\includegraphics[width=1\linewidth]{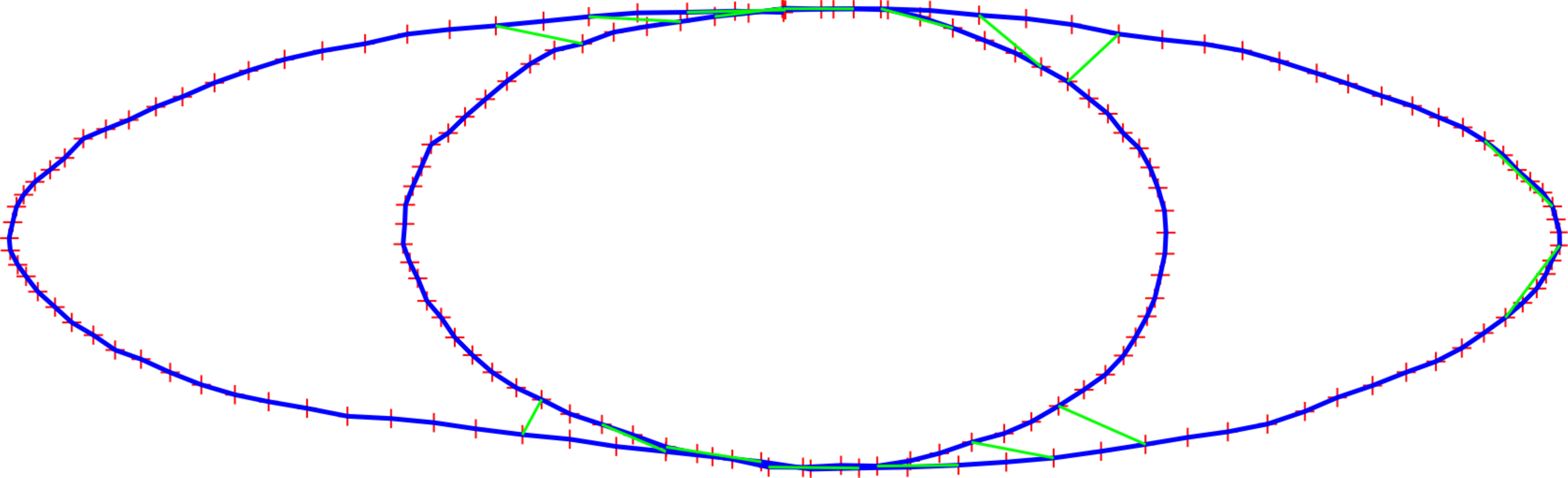}
\end{minipage} \hspace{.25cm}
\begin{minipage}{0.42\linewidth}
	\centering
	\includegraphics[width=1\linewidth]{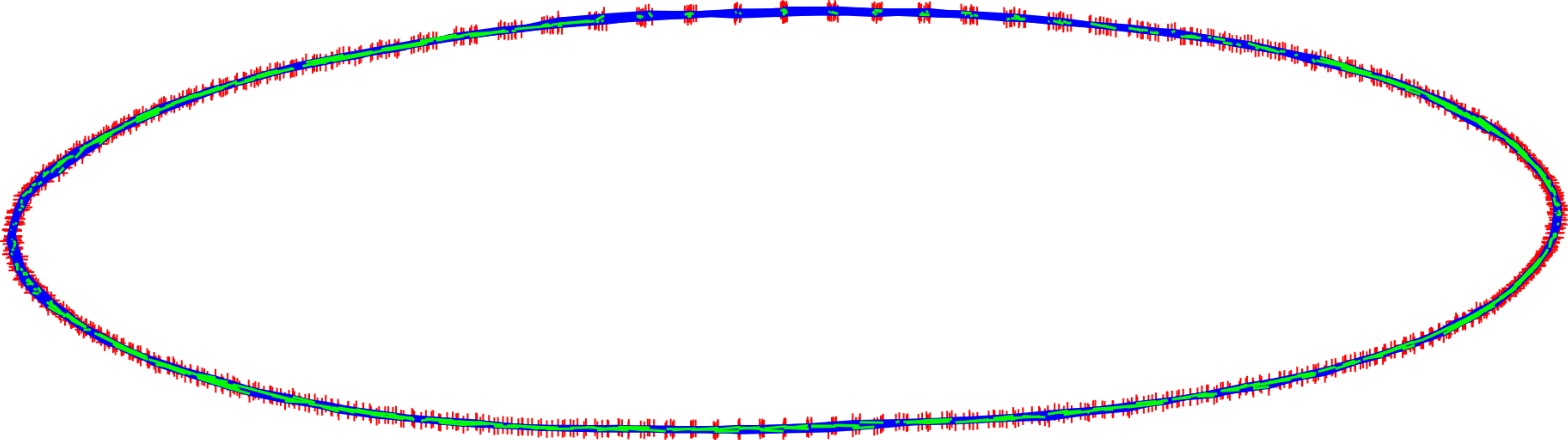}
\end{minipage} \hfill \vspace{.2cm} \\
\begin{minipage}{.5cm}
	\centering
	c)
\end{minipage}
\begin{minipage}{0.42\linewidth}
	\centering
	\includegraphics[width=1\linewidth]{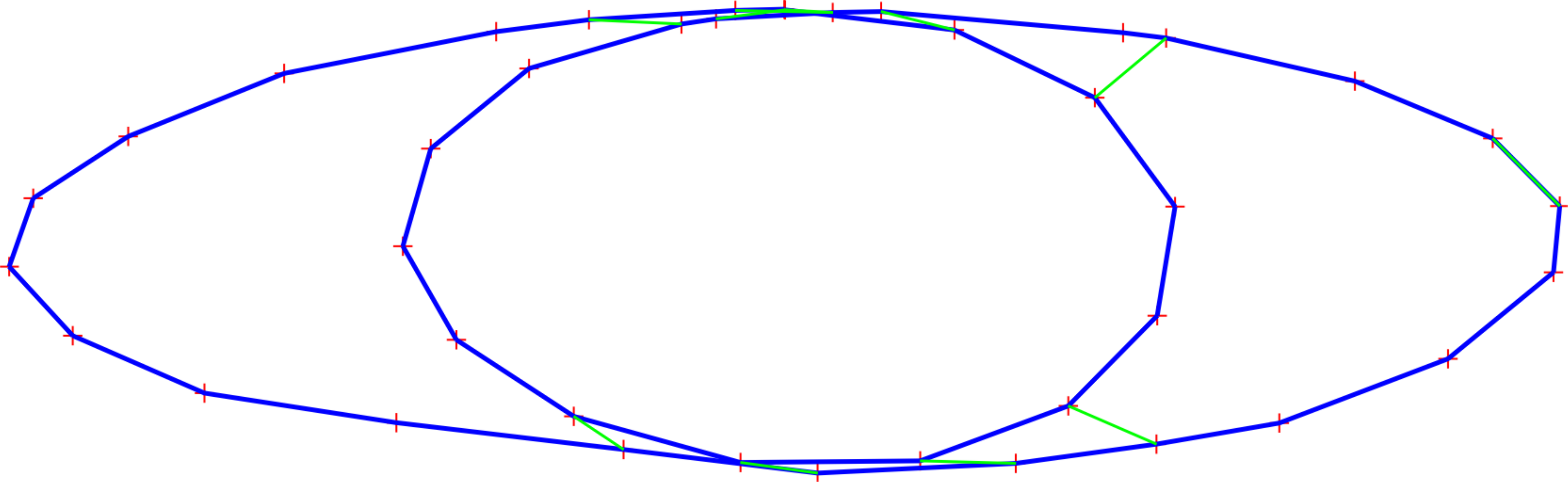}
\end{minipage} \hspace{.25cm}
\begin{minipage}{0.42\linewidth}
	\centering
	\includegraphics[width=1\linewidth]{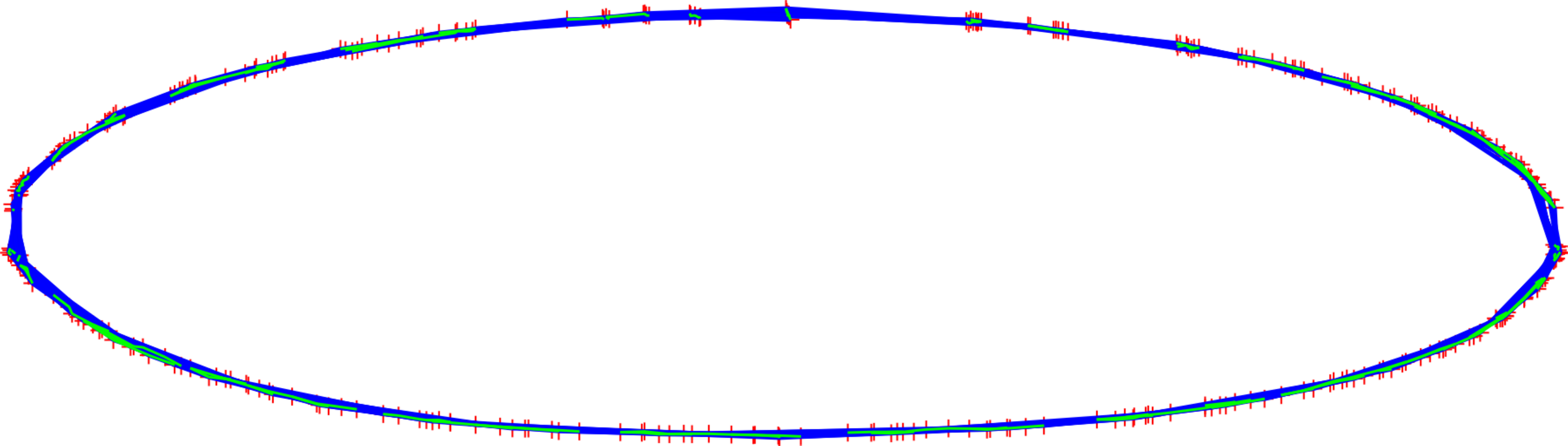}
\end{minipage} \hfill
\end{tabular}
\end{center}
\caption{\emph{Ellipse} datasets; poses (red), and loop closures (green); a) using all possible loop closures, b) using only relevant loop closures and c) compact pose SLAM with reduction of the number of poses.}
\label{fig:ellipse_plots}
\end{figure}
\subsection{Performance and Accuracy Analysis}\label{SubSec:Performance}

A compact representation of the SLAM problem translates into more efficient estimation both from the point of view of the memory occupied by the state as well as from the point of view of the execution time. To validate this we compare the cumulative time at the end of the processing of each of the datasets mentioned above.
We first define the configuration of the thresholds used to obtain the APAL and FPFL solutions for each of the datasets.

For example, for the \emph{ellipse3D} dataset, following the procedure described at the end of \autoref{SubSec:CompactSLAMeval}, we set \mbox{$v=[2.54\:m, 2.54\:m, 2.54\:m, 0.36\:\textdegree]$}, \mbox{$s = 0.125$}, \mbox{$g_{pose} = -\infty$} and \mbox{$g_{loop} = -\infty$} to obtain an estimation problem which includes all poses and all loops (APAL). 
We then increased the \mbox{$g_{loop} = 5.50$} to obtain an estimation problem which contains all the poses but only the informative links (APFL). Similarly, increasing \mbox{$g_{pose} = 5.74$} leads to a compact estimation problem, yet containing maximal amount of information. The resulting trajectories in all three cases are shown in \autoref{fig:ellipse_plots} left, and the timing and accuracy are reported in the Tables \ref{tab:Time} and \ref{tab:ErrorEval}, respectively. Out of $377$ possible loop closure links, the algorithm selected only $11$ relevant ones, resulting in a factor of $7.6 \times$ reduction in the run time. Nevertheless, only $0.228 \%$ of the accuracy in position and $1.140\%$ in rotation is lost when reducing the number of links to only informative. 

Similar tests were performed with the rest of the datasets. This paper reports only the APAL and FPFL cases, being the most relevant in our comparisons. \autoref{fig:Datasets} shows the solutions of several SLAM datasets processed by allowing all poses and all loop closures to be added to the state representation (\autoref{fig:Datasets}~a)) and by selecting only informative links and non-redundant poses in an incremental processing (\autoref{fig:Datasets}~b)). The corresponding timing results are shown in \autoref{tab:Time}. We can see that for all the datasets there is a considerable time reduction when performing compact SLAM. \autoref{tab:ErrorEval} on the other hand, shows that the translational and rotational errors increase only slightly, in the case of the compact SLAM. To provide some perspective on the values of the errors, we also performed \emph{random} selection of poses and loop closures on the \emph{kitti00} and \emph{parking-garage} datasets, in order to get a solution with the same sparsity as the one in the FPFL case. Those results are denoted RFPFL and it is visible that they are much worse in both cases. 

Runtime evaluation is provided in \autoref{tab:Time}. The APAL and FPFL strategies are integrated into the SLAM++ library and compared against the solution of the SLAM++ \cite{Polok13rss}, g2o \cite{Kuemmerle11icra}, and iSAM \cite{Kaess08tro} solvers with neither loop detection nor compact representation. The time required to obtain the marginal covariances is also provided, except for the $100k$ dataset processed with g2o and iSAM where it takes several days. The plain nonlinear least squares solver and marginalisation times are used only as reference. Note that there is an important difference on how the incremental processing is performed in APAL and FPFL strategies. While in plain nonlinear solving (SLAM++, g2o and iSAM columns in \autoref{tab:Time}) the incremental updates occur \emph{every new vertex}, in APAL and FPFL the updates happen \emph{every new measurement} (see \autoref{alg:compactSLAM}, line \ref{algline:incUpAfterLoop}). This is due to the fact that the mutual information of every measurement is calculated. In general the number of edges in the system is much higher than the number of vertices. Nevertheless, due to highly efficient block matrix solvers and covariance recovery algorithms implemented into SLAM++ library, the APAL strategy has comparable runtime and has the benefit of providing state-based loop closure detection. At the same time, the FPFL strategy remains efficient by maintaining a compact representation of the state, but at the same time integrating the state-based loop closure detection technique. \autoref{tab:Time} also reports percentage of loops and poses kept in the compact representation. 


\begin{table}[t]
\begin{center}
\begin{tabular}{|c|c|c|c|c|}\hline
Dataset 			& $v$ 				& $s$ 				&$g_{pose}$ &$g_{loop}$ \\\hline
ellipse3D	& $\{2.5\}^3, 0.4$	& $\frac{1}{8}$	& $5.74$ & $5.50$ \\
ellipseN	& $\{1.1\}^3, 0.3$	& $\frac{1}{10}$	& $5.70$ & $5.11$ \\
kitti00 	& $\{25.5\}^3, 1.1$	& $\frac{1}{10}$	& $8.51$	& $5.13$ \\
parking-garage	& $\{95.0\}^3, 1.1$	& $\frac{1}{10}$	& $5.77$	& $2.45^*$ \\
sphere2500 
			& $\{2.9\}^3, 0.1$	& $\frac{1}{10}$	& $4.33^*$	& $9.14$ \\
10kHog-man		& $\{8.9\}^2, 6.3$	& $\frac{1}{10}$	& $2.36$	& $1.94$ \\
100k			& $\{27.6\}^2, 6.2$	& $\frac{1}{10}$	& $2.36$	& $4.08$ \\
\hline
\end{tabular}
\end{center}\caption{Compact SLAM thresholds. Note that the $\{X\}^Y$ notation in the $v$ column means merely $Y$ repetitions of $X$, and was introduced to save space. $^*$ The thresholds marked by asterisk were manually modified, as described at the end of \autoref{SubSec:CompactSLAMeval}.}
\label{tab:Thresh}
\end{table}

\begin{table*}[t]
\begin{center}
\begin{tabular}{|c|c|c|c||c|c|c|c|}\cline{2-8}
\multicolumn{1}{c|}{} & \multicolumn{3}{c||}{Time of covariance calculation / nonlinear solving [s]} & \multicolumn{2}{c|}{Time [s]} & \multicolumn{2}{c|}{FPFL [\%]} \\ \hline
Dataset 			& SLAM++ 			& g2o 				& iSAM				& APAL			& FPFL 			& loop & vert. \\ \hline
ellipse3D			& $0.112 / 0.051$		& $1.197 / 0.160$		& $1.156 / 1.868$			& $0.502$			& $0.066$			& $2.91$ & $23.52$ \\
ellipseN			& $30.564 / 8.058$		& $419.229 / 25.884$	& $480.056 / 150.264$		& $146.954$			& $5.387$		& $3.20$ & $31.00$ \\
kitti00 			& $33.185 / 77.873$		& $679.592 / 80.148$	& $733.914 / 688.287$		& $203.045$			& $6.163$		& $1.63$ & $7.81$ \\ 
sphere2500 		& $29.589 / 85.870$		& $5474.351 / 207.284$	& $5965.062 / 281.754$		& $655.016$			& $19.558$	& $16.32$ & $38.36$ \\
parking-garage		& $11.717 / 13.589$	& $212.725 / 20.302$	& $243.867 / 147.649$		& $102.989$			& $30.602$			& $20.86$ & $64.29$ \\
10kHog-man			& $201.498 / 246.005$	& $5765.850 / 552.462$	& $5955.160 / 1431.893$		& $3102.154$		& $401.587$			& $7.11$ & $46.76$ \\
100k				& $4h53m / 18h22m$				& $- / 22h03m$				& $- / 40h24m$					& $200h 28m$				& $26h15m$		 		& $4.26$ & $25.30$ \\ \hline
\end{tabular}
\end{center}
\caption{Time performance in seconds. APAL and FPFL, both include state-based loop closure detection. SLAM++, g2o and iSAM columns show the solving time for given data association.}
\label{tab:Time}
\end{table*}

\begin{table*}[t]
\begin{center}
\begin{tabular}{|c|c|c|c|c|c|c|c|}\cline{3-8}
\multicolumn{2}{c|}{} & \multicolumn{6}{c|}{RMSE Error} \\ \hline
\multirow{2}{*}{Dataset} 			& \multirow{2}{*}{Mode} & \multicolumn{2}{c|}{ATE} & \multicolumn{2}{c|}{RPE} &\multicolumn{2}{c|}{RPE all-all}	\\\cline{3-8}
 & & translation & rotation & translation & rotation & translation & rotation \\\hline
\multirow{2}{*}{ellipse3D}			& APAL	& $0.173$ & $3.666$ & $0.061$ & $1.643$ & $0.521$ & $3.121$ \\
							& FPFL	& $0.338$ & $5.815$ & $0.100$ & $3.156$ & $0.839$ & $5.778$ \\ \hline
\multirow{2}{*}{ellipseN}				& APAL	& $0.138$ & $1.599$ & $0.039$ & $1.102$ & $0.338$ & $1.535$ \\
							& FPFL	& $0.235$ & $4.820$ & $0.095$ & $2.737$ & $0.841$ & $4.033$ \\ \hline
\multirow{3}{*}{kitti00 
				}			& APAL	& $3.046$ & $2.251$ & $0.037$ & $0.143$ & $7.007$ & $2.188$ \\
							& FPFL	& $3.093$ & $2.250$ & $0.029$ & $0.119$ & $7.297$ & $2.223$ \\
							& RFPFL	& $12.017$ & $3.702$ & $0.028$ & $0.118$ & $17.448$ & $3.623$ \\ \hline
\multirow{2}{*}{sphere2500 
				}			& APAL	& $0.203$ & $1.397$ & $0.166$ & $1.582$ & $0.905$ & $1.391$ \\
							& FPFL	& $0.457$ & $2.440$ & $0.244$ & $2.179$ & $1.580$ & $2.397$ \\ \hline
\multirow{3}{*}{parking-garage$^\dag$}	& APAL	& $0.193$ & $0.787$ & $0.016$ & $0.270$ & $2.563$ & $0.777$ \\
							& FPFL	& $0.661$ & $0.757$ & $0.012$ & $0.109$ & $1.167$ & $0.647$ \\
							& RFPFL	& $6.197$ & $9.761$ & $0.247$ & $1.688$ & $9.910$ & $7.264$ \\ \hline
\multirow{2}{*}{10kHog-man}			& APAL	& $0.917$ & $3.325$ & $0.081$ & $1.612$ & $3.707$ & $3.318$ \\
							& FPFL	& $1.479$ & $4.902$ & $0.139$ & $2.583$ & $5.754$ & $4.896$ \\ \hline
\multirow{2}{*}{100k}				& APAL	& $0.913$ & $0.534$ & $0.009$ & $0.270$ & $1.644$ & $0.522$ \\
							& FPFL	& $1.205$ & $0.814$ & $0.020$ & $0.473$ & $2.559$ & $0.802$ \\ \hline

\end{tabular}
\end{center}
\caption{Error evaluation. $^\dag$Note that the \emph{parking-garage} dataset does not come with ground truth and a de-facto ground truth was obtained by batch solving until convergence.} 
\label{tab:ErrorEval}\vspace{-1mm} 
\end{table*}

%
\begin{figure}[t!]
\begin{center}
\begin{tabular}{c}
\includegraphics[trim=40mm 28mm 40mm 20mm, clip, width=0.9\linewidth]{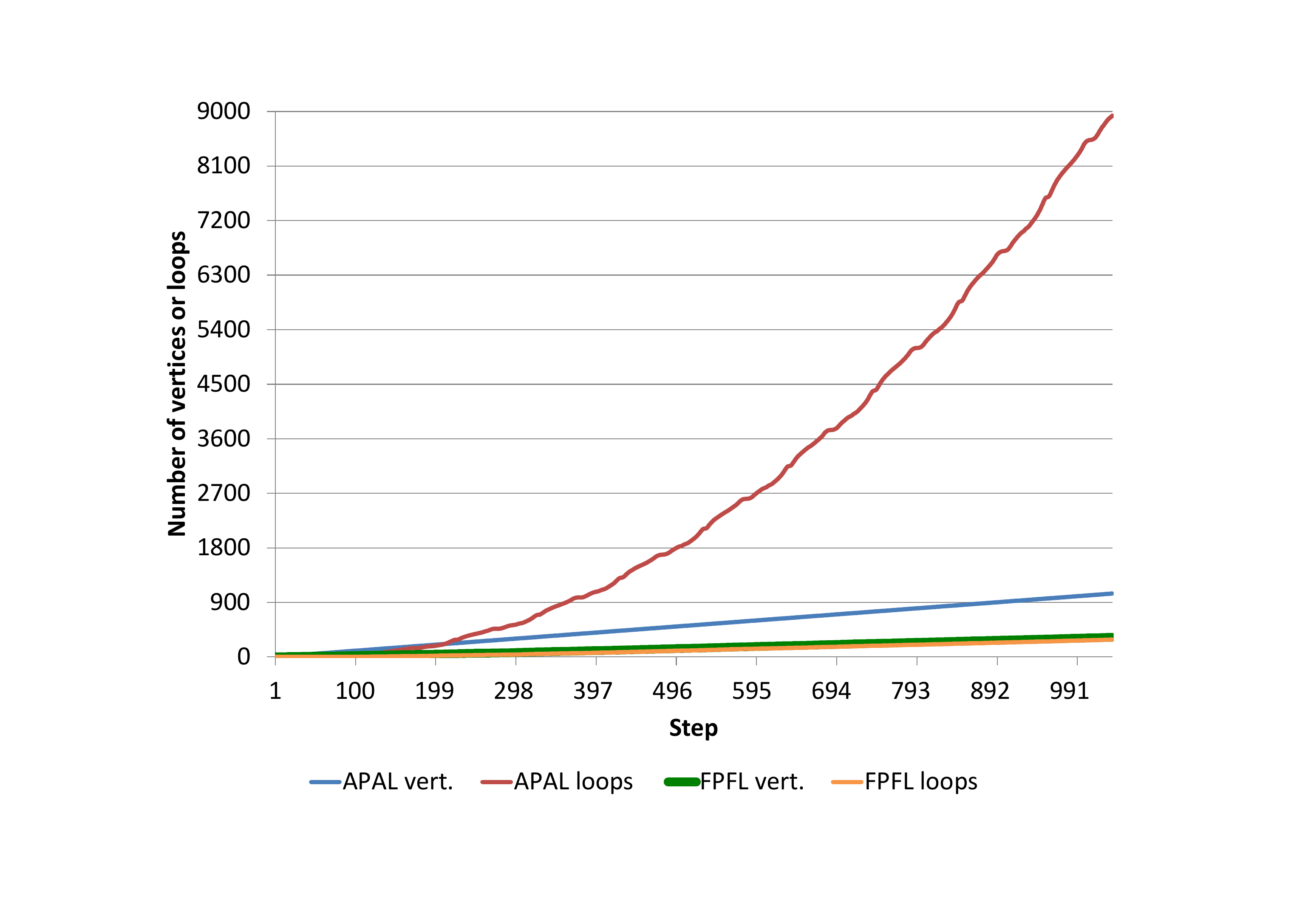}\\ 
\includegraphics[trim=36mm 28mm 40mm 20mm, clip, width=0.9\linewidth]{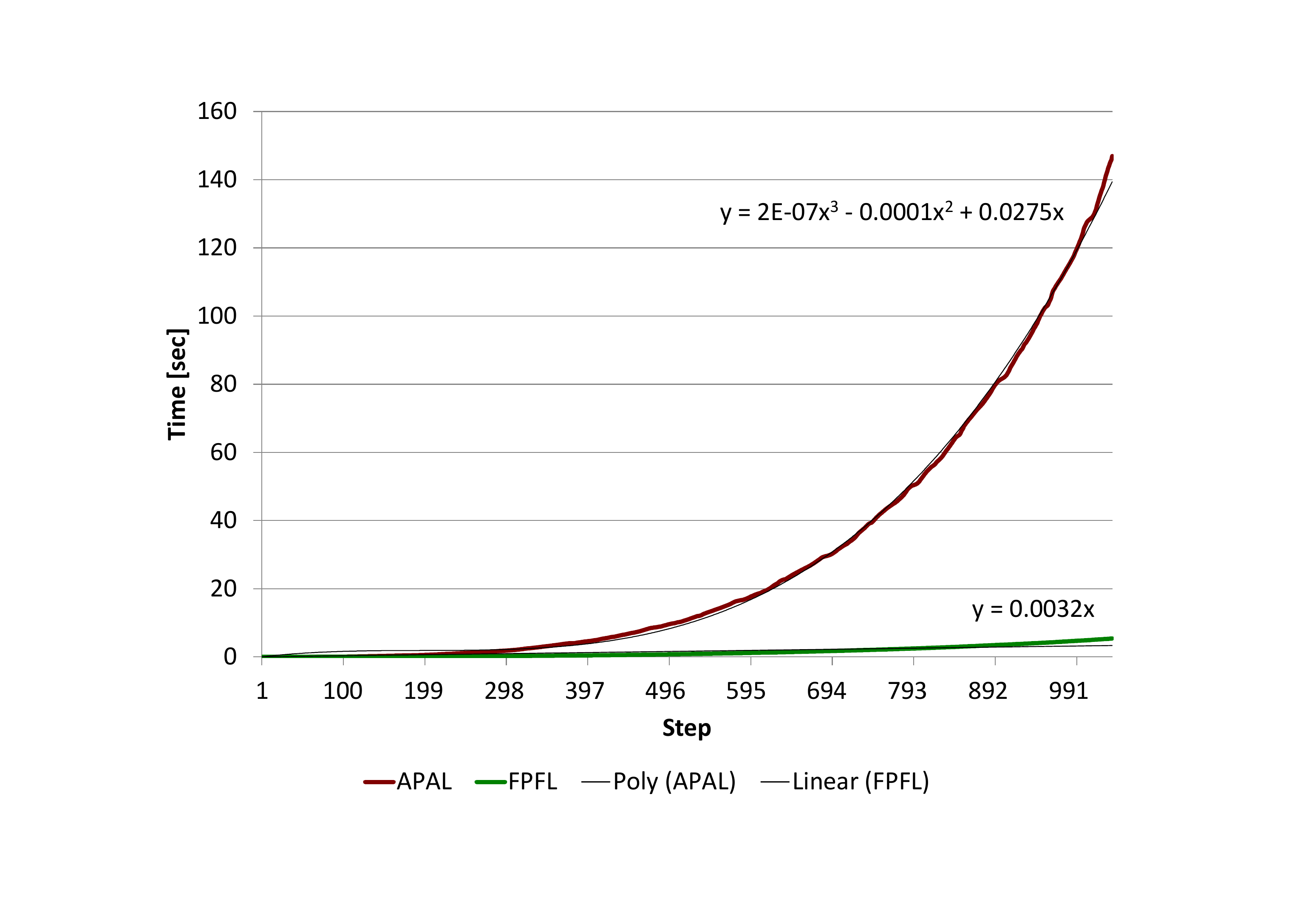} 
\end{tabular}
\end{center}
\caption{Evolution of the number of poses and loop closures for both strategies APAL and FPFL (top) and cumulative time of distance based loop-closing (bottom) on the \emph{ellipseN} dataset.}
\label{fig:ellipseN} \vspace{-.25cm} 
\end{figure}
\begin{figure}[t!]
\begin{center}
\begin{tabular}{c}
\includegraphics[trim=40mm 28mm 40mm 20mm, clip, width=0.9\linewidth]{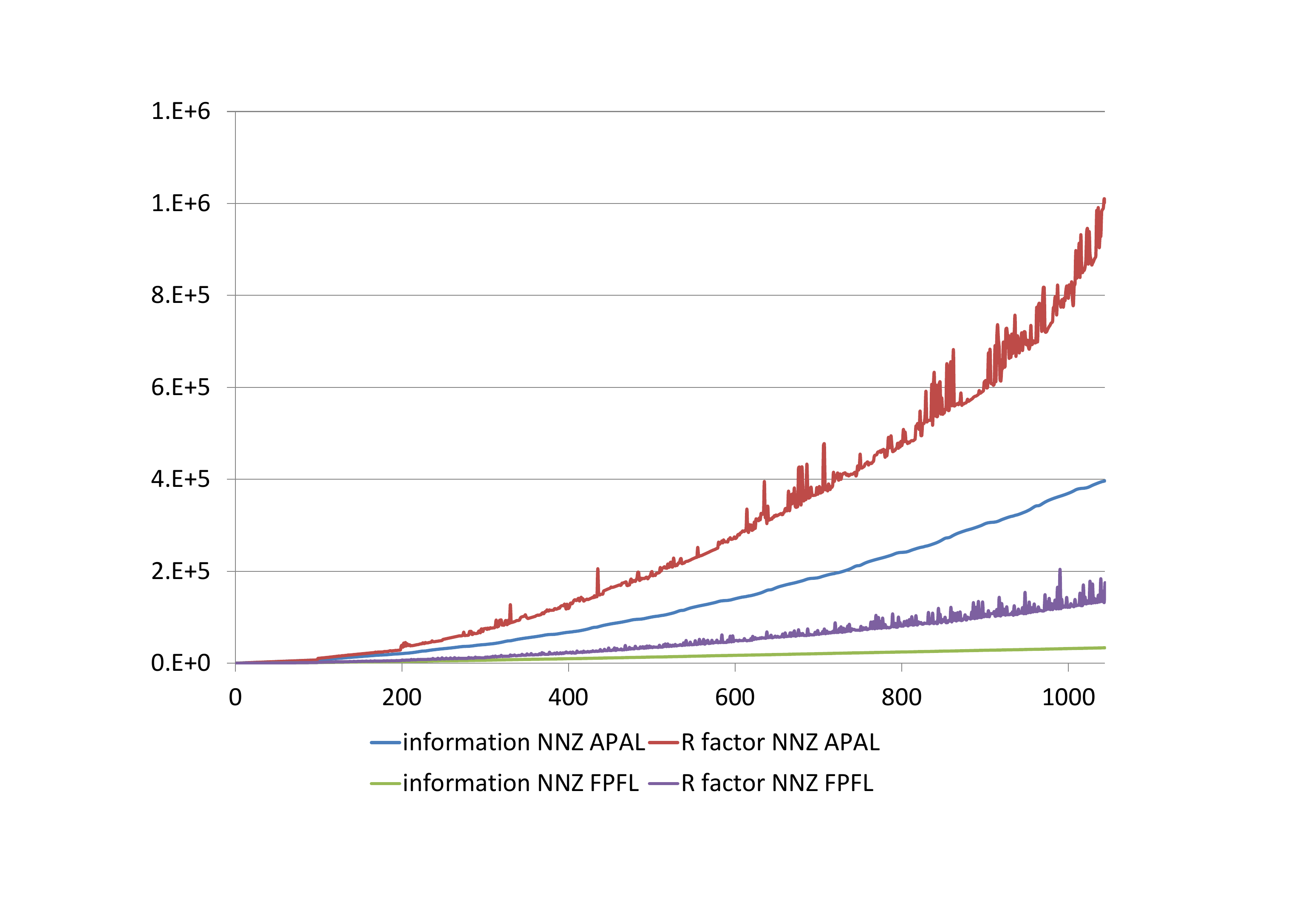}
\end{tabular}
\end{center}
\caption{Evolution of number of nonzero system matrix entries on the \emph{ellipseN} dataset.}
\label{fig:musageEllipseN} \vspace{-.3cm} 
\end{figure}
The compact pose SLAM was also tested on a multiple loops dataset. Similar to \emph{ellipse3D} we created \emph{ellipseN} dataset which loops $N$-times around an ellipse with semi-axes of $20\:m$ and $6\:m$, respectively, see \autoref{fig:ellipse_plots} right. \autoref{fig:ellipseN} (top) shows the evolution of number of poses and loop closures over $N=10$ loops around the ellipse. We can see that, while in APAL the number of loop-closures increases exponentially, the FPFL strategy maintains a linear trend with a low slope increase in number of both poses and loops. While looping $10$ times took $146.954$s to run incrementally with the APAL strategy, it took only $5.387$s with the compact representation. \autoref{fig:ellipseN} (bottom) shows the cumulative time for APAL and FPFL strategies, and includes the solving and search for loop closures times. While APAL runs in polynomial time the FPFL runs in linear time.

We have also evaluated the memory requirements of the algorithm, again on the \emph{ellipseN} dataset. \autoref{fig:musageEllipseN} shows the evolution of the number of nonzero elements in the information matrix and its factorisation at each step. Note that for all poses all loops, the size of both matrices grows exponentially while in the compact case the size of the information matrix follows a linear trend. The number of nonzeros in the R factor is slightly higher due to fill-in and is a bit noisy due to the incremental reordering strategy described in \cite{Polok13rss}.

\begin{figure}[t!]
\begin{center}
\begin{tabular}{c}
\includegraphics[trim=40mm 28mm 40mm 20mm, clip, width=0.9\linewidth]{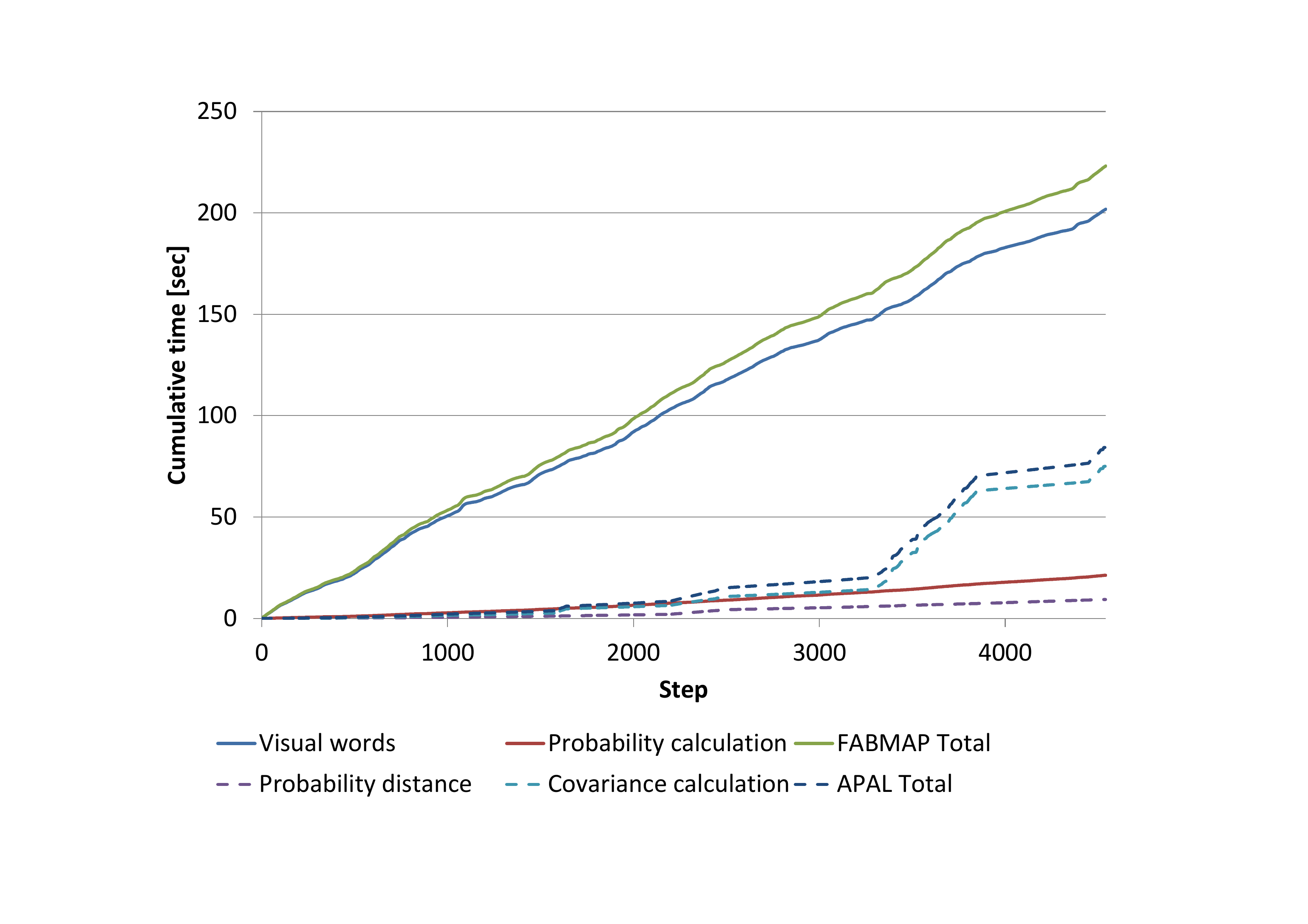}
\end{tabular}
\end{center}
\caption{Comparison of cumulative time of appearance-based data association on the \emph{kitti00} dataset using \emph{FAB-MAP2} and the Oxford dictionary provided by the authors (solid lines) and distance-based data association using the proposed method (dashed lines).} 
\label{fig:fabmapCumTime}
\end{figure}
The distance loop closing strategy was compared to appearance-based loop-closing implemented in the \emph{FAB-MAP2} library \cite{Cummins10ijrr}. \autoref{fig:fabmapCumTime} shows the execution time comparison of the two methods. For FAB-MAP2, the plot shows a sum of the time of transforming the feature descriptors into visual words and the match probability matrix calculation. We have used an incremental approach, as described by the authors, and at each step, one new column of the match probability matrix is added. Note that the time complexity is dominated by the visual words formation, which in turns depends on the number of features detected in the image and on the size of the vocabulary. For the distance based approach, we time the equivalent computation, consisting of finding the loop closure candidates and calculating the information gains for them. 
For the \emph{kitti00} dataset, our method significantly outperforms FAB-MAP2, even if including the time it takes to calculate the covariances. Note that this test was performed on the Oxford dictionary which was provided by the authors of FAB-MAP2. For the following tests, we have also trained our own dictionary on the \emph{kitti} dataset, as described below. The loop closing run time with this vocabulary has the same complexity with higher constant factor. The final total time with this dictionary is $2107.71$s (out of that $2043.00$s visual words and $64.72$s probability calculation) and was omitted from \autoref{fig:fabmapCumTime} otherwise the bottom part of the plot would be illegible.


%

\section{Conclusions} \label{Sec:Conclusions}

This paper addressed both the efficiency and temporal scalability of the online SLAM. SLAM++ nonlinear least square solver based on efficient sparse block matrix operations has already proven its superiority over the existing solutions for incremental processing in SLAM \cite{Polok13rss}. At the same time, in our latest work \cite{Ila15icra}, we showed how the uncertainty of the estimate can be calculated incrementally in a very efficient manner.

This paper comes to integrate all the above mentioned characteristics into a complete SLAM algorithm which not only maintains a scalable representation of the state but also efficiently contributes to the data association process without a significant computational overhead. Information theory measures play an important role in the proposed technique, allowing for principled methods to select only informative links and non-redundant poses. The proposed system automatically limits the growth of the map representation when continuously operating in the same environment. The results significantly outperform the state of the art in processing speed while maintaining an accurate estimation. The proposed method for distance-based loop closure detection has the benefit of being applicable to modalities of sensors other than image, being fast at the same time.

In addition to that, we also discuss several methods for efficiently recovering an estimate of the full state from the compact representation. While in here they are applied incrementally, it would also be possible to apply them in batch solving in order to convert the problem to a much smaller one by condensing all the vertices of order two. The full solution can then be efficiently reconstructed in linear time.


While the information theoretic measures provide a solid foundation for selecting informative links and non-redundant poses, there is also a question of robustness to outliers. Intuitively, outlier measurements are likely to have high mutual information, but at the same time they are undesirable. In the continuation of this work, we will aim to verify this claim and integrate robustness in the process of maintaining a compact and scalable representation of the SLAM problem. Furthermore, the resulting algorithm can easily be adapted to landmark SLAM and even to structure from motion allowing 3D mapping of large scale environments.



\balance

\subsection*{Acknowledgements} \label{Sec:Acknowledgements}
We are extremely grateful to the ARC Centre of Excellence
for Robotic Vision, project number CE140100016 for funding Dr. Viorela Ila carrying out this research.\\
The authors from Brno University of Technology received funding from the European Union, $7^{th}$ Framework Programme grant 316564-IMPART
and the IT4-Innovations Centre of Excellence project (CZ.1.05/1.1.00/02.0070), funded by the European Regional Development Fund and the national budget of the Czech Republic via the Research and Development for Innovations Operational Programme, as well as Czech Ministry of Education, Youth and Sports via the project Large Research, Development and Innovations Infrastructures (LM2011033). 
\bibliographystyle{apalike}
\bibliography{refs}

\end{document}